\documentclass[conference]{IEEEtran}
\usepackage{times}
\usepackage{tikz}
\usepackage{functional}
\usepackage{pgffor}
\usepackage{pgfplots}
\pgfplotsset{compat=1.18}
\usepgfplotslibrary{fillbetween}

\usepackage{graphicx} 
\usepackage[numbers]{natbib}
\usepackage{multicol}
\usepackage[bookmarks=true]{hyperref}
\usepackage{amsmath}
\usepackage[
    frozencache=true,  
    cachedir=minted-cache,
]{minted}
\usepackage{csquotes}
\usepackage{mdframed}
\usepackage{soul}
\usepackage{multicol}
\usepackage{multirow}
\usepackage{booktabs}
\usepackage{caption}
\usepackage{subcaption}
\usepackage{cleveref}
\usepackage{pdflscape}

\usepackage[per-mode=fraction]{siunitx}
\usepackage{comment}
\usepackage[normalem]{ulem}
\usepackage{xspace}
\usepackage{xcolor}

\DeclareSIUnit\px{px}
\DeclareSIUnit\fps{fps}

\definecolor{OliveGreen}{RGB}{0,200,25}

\newcommand{\ie}{i.\,e.\xspace}
\newcommand{\eg}{e.\,g.\xspace}
\newcommand{\etal}{et\,al.\xspace}

\newcommand{\armarVI}{\mbox{ARMAR-6}\xspace}

\newcommand{\armarx}{\mbox{ArmarX}\xspace}

\newcommand{\ackJuBotReallabor}{This work has been supported by the Carl Zeiss Foundation through the JuBot project and by the Baden-Württemberg Ministry of Science, Research and the Arts (MWK) as part of the state's ``digital@bw'' digitization strategy in the context of the Real-World Lab ``Robotics AI.''}

\newcommand{\inlinecode}[1]{\enquote{\small\texttt{#1}}}
\newcommand{\codesize}[0]{\fontsize{6.5}{7} \selectfont}

\crefname{section}{Section}{Sections}
\Crefname{section}{Section}{Sections}
\crefname{figure}{Fig.}{Figs.}
\Crefname{figure}{Fig.}{Figs.}
\crefname{table}{Table}{Table}
\Crefname{table}{Table}{Table}
\crefname{equation}{Eq.}{Eqs.}
\Crefname{equation}{Eq.}{Eqs.}
\crefname{listing}{Listing}{Listings}
\Crefname{listing}{Listing}{Listings}

\makeatletter
\let\@float@c@listing\@caption
\makeatother

\begin{document}

\title{Incremental Learning of Humanoid Robot Behavior \\from Natural Interaction \& Large Language Models}

\author{\authorblockN{Leonard Bärmann, Rainer Kartmann, Fabian Peller-Konrad, Jan Niehues, Alex Waibel, Tamim Asfour}
\authorblockA{Institute for Anthropomatics and Robotics~(IAR)\\
Karlsruhe Institute of Technology~(KIT), Germany\\
{\tt\footnotesize \{baermann, asfour\}@kit.edu}}
}

\maketitle

\begin{abstract}
Natural-language dialog is key for intuitive human-robot interaction.
It can be used not only to express humans' intents, but also to communicate instructions for improvement if a robot does not understand a command correctly.
Of great importance is to let robots learn from such interaction experience in an incremental way to allow them to improve their behaviors or avoid mistakes in the future.
In this paper, we propose a system to achieve such incremental learning of complex behavior from natural interaction, and demonstrate its implementation on a humanoid robot.
Our system deploys Large Language Models (LLMs) for high-level orchestration of the robot's behavior, based on the idea of enabling the LLM to generate Python statements in an interactive console to invoke both robot perception and action.
Human instructions, environment observations, and execution results are fed back to the LLM, thus informing the generation of the next statement.
Specifically, we introduce incremental learning from interaction, which enables the system to learn from its mistakes.
For that purpose, the LLM can call another LLM responsible for code-level improvements of the current interaction based on human feedback.
Subsequently, we store the improved interaction in the robot's memory so that it can later be retrieved on semantically similar requests.
We integrate the system in the robot cognitive architecture of the humanoid robot \armarVI and evaluate our methods both quantitatively (in simulation) and qualitatively (in simulation and real-world) by demonstrating generalized incrementally-learned knowledge.
\end{abstract}

\section{Introduction}
\begin{figure}[tb]
    \centering
    \includegraphics[width=\linewidth]{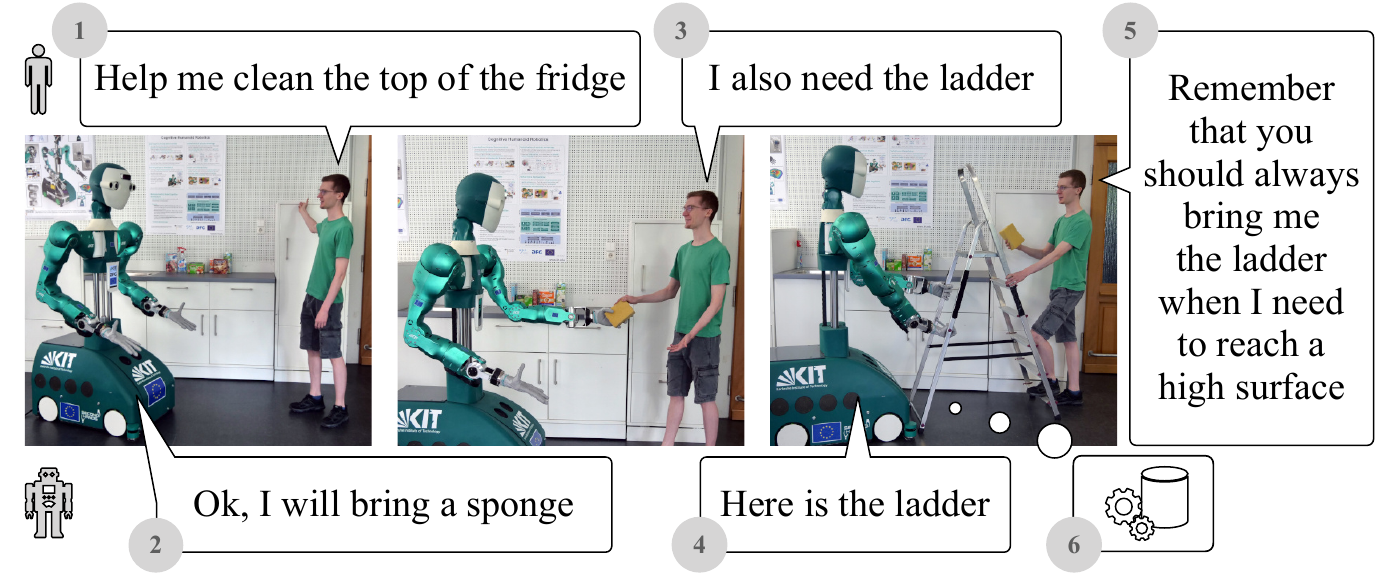}
    \caption{
    \armarVI incrementally learns behavior from natural interaction.
    Demonstration video at \url{https://youtu.be/y5O2mRGtsLM}}
    \label{fig:interaction_example}
\end{figure}

Humans can easily communicate tasks and goals to a robot via language.
Such natural language interface is key for achieving truly intuitive human-robot interaction (HRI). 
However, the robot's interpretation of such commands, and thus the resulting execution, might be sub-optimal, incomplete or wrong.
In such cases, it is desirable for the human to give further instructions to correct or improve the robot's behavior.
Furthermore, the robot should memorize the improvement strategy given by the human to incrementally learn from them and thus avoid the same mistake in the future.
For instance, consider the interaction depicted in \cref{fig:interaction_example}.
First, the user instructs the robot to help him cleaning the top of the fridge~(1).
The robot then executes several actions to hand over a sponge to the human~(2).
The user observes this insufficient result and gives instructions for improvement (\enquote{I also need a ladder})~(3), whereupon the robot performs corrective actions~(4).
If the desired goal is achieved, the user can reconfirm the correction~(5), which leads to the robot updating its memory appropriately~(6), thus incrementally learning new behavior based on language instructions.

In this paper, we present a system to achieve such behavior and describe its implementation on the humanoid robot \armarVI~\cite{Asfour2018ARMAR6CollaborativeHumanoid}.
We build on the capabilities of Large Language Models~(LLMs)~\cite{brown_language_2020,
touvron_llama_2023,openai_chatgpt_2023,openai_gpt4_2023} emerging from massive-scale next token prediction pretraining, and aim to transfer their success to HRI.
The goal is to utilize the rich world knowledge contained in LLMs for embodied natural-language dialog, thus enhancing the capabilities of the LLM by integrating robot perception and action.
In the cognitive architecture of our humanoid robot~\cite{Peller-Konrad2023MemorySystemRobot}, this means the LLM will be in charge of the high-level planning and decision-making.
Recent works like SayCan~\cite{ahn_as_2022} and Code as Policies (CaP)~\cite{liang_code_2023} already demonstrate the usefulness of applying LLMs to orchestrate robot abilities, enabling high-level task understanding, planning and generalization.
Going a step further, inner monologue~\cite{huang_inner_2022} feeds back execution results and observations into the LLM, thus involving the LLM in a closed-loop interaction.

Inspired by these works, we propose to utilize the code-writing capabilities of LLMs to directly integrate it into closed-loop orchestration of a humanoid robot.
This is achieved by simulating an interactive (Python) console in the prompt, and letting the LLM produce the next statement given the previous execution history, including results returned or exceptions thrown by previous function calls.
Thus, the LLM can dynamically respond to unexpected situations such as execution errors or wrong assumptions, while still leveraging the power of code-based interaction such as storing results in intermediate variables or defining new functions.

For utilizing the few- and zero-shot capabilities of LLMs, it is crucial to design a (set of) prompts to properly bias the LLM towards the desired output.
All of the above works use a predefined, manually written set of prompts tuned for their respective use case.
In contrast, we propose a novel, self-extending prompting method to allow incremental learning of new behaviors.
To this end, our system dynamically constructs prompts based on a set of interaction examples, populated from the robot's prior knowledge and previously learned behavior.
Given a user instruction, we rank all such interaction examples by semantic similarity to the input, and select the top-$k$ entries to construct the actual prompt to the LLM.
Crucially, the robot's prior knowledge contains specific examples involving the user complaining about mistakes and correcting the robot, or instructing it on how to improve its behavior.
Therefore, when the system fails to correctly execute a task and the user gives such corrective instructions, the LLM is biased to invoke code that inspects the current execution history and forwards it to another, few-shot-prompted LLM.
This LLM can inspect the complete interaction including all user inputs, performed actions and observed results, represented as the transcript of an interactive Python console.
It then spots the mistakes and produces an improved interaction using chain-of-thought (CoT) prompting~\cite{wei_chain_2022}.
Finally, the improved transcript will be added to the interaction examples, thus enabling the system to perform better the next time a similar task is requested.

Our method is explained in detail in \cref{sec:methods}.
We evaluate our system quantitatively on the scenarios defined in CaP~\cite{liang_code_2023} to show the effectiveness of our proposed approach in \cref{sec:results}.
Furthermore, \cref{sec:demonstration} demonstrates the capabilities of incremental learning from natural-language interaction on a real-world humanoid robot.


\section{Related Work}
\label{sec:related_work}

We start with reviewing works on understanding and learning from natural language in robotics.
Subsequently, we present works using LLMs for high-level orchestration of robot abilities.
Finally, we focus on dynamic creation of prompts for LLMs.

\subsection{Understanding and Learning from Natural Language}

Understanding and performing tasks specified in natural language has been a long-standing challenge in robotics~\cite{Tellex2020RobotsThatUse}.
Of great challenge is \emph{grounding} the words of natural language sentences in the robot's perception and action, which is known as \emph{signal-to-symbol gap}~\cite{Kruger2011ObjectAction}.
Many works have focused on the grounding of expressions referring to objects, places and robot actions based on
graphical models~\cite{Tellex2011UnderstandingNaturalLanguage,Misra2016TellMeDave}, 
language generation~\cite{Forbes2015RobotProgrammingDemonstration},
or spatial relations~\cite{Guadarrama2013GroundingSpatialRelations}, especially for ambiguity resolution~\cite{Fasola2013UsingSemanticFields,Shridhar2020INGRESSInteractiveVisual}.
Pramanick~\etal~\cite{Pramanick2020DeComplexTaskPlanning} focus on resolving task dependencies to generate execution plans from complex instructions.
However, in these works the robot does not explicitly learn from language-based interactions.
In contrast,
Walter~\etal~\cite{Walter2013LearningSemanticMaps} enrich the robot's semantic environment map from language,
and
Bao~\etal~\cite{Bao2016TeachRobotsUnderstanding} syntactically parse daily human instructions to learn attributes of new objects.
In~\cite{Kartmann2023InteractiveIncrementalLearning}, the robot asks for a demonstration if its current understanding of a spatial relation is insufficient to perform a given instruction.
Other works go further by learning on the task level.
Mohan~\etal~\cite{Mohan2014LearningGoalOrientedHierarchical} learn symbolic task representations from language interaction using Explanation-based learning.
Nicolescu~\etal~\cite{Nicolescu2019LearningComplexStructuredTasks} learn executable task representations 
encoding sequential, non-ordering or alternative paths of execution from verbal instructions for interactive teaching by demonstration.
Weigelt~\etal~\cite{Weigelt2020ProgrammingNaturalLanguage} consider the general problem of programming new functions on code level via natural language.
%
%
While our goal is similar to these works,
we leverage LLMs for task-level reasoning and learning.

\subsection{Code-Generation and Interaction with LLMs}
Generating code from natural language specifications is a large area of active research.
For instance, LLMs tuned specifically on code~\cite{chen_evaluating_2021,nijkamp_codegen_2023} perform well in common code-generation benchmarks. 
\citet{madaan_language_2022} show that code-based models have more structured representations, thus aiding structured (e.g. graph-based) tasks.
Training code-LLMs can also benefit from using an interpreter in the optimization loop~\cite{le_coderl_2022,haluptzok_language_2023}.
We refer the reader to recent surveys~\cite{zheng_survey_2024,moniz_source-code_2023,dehaerne_code_2022,wang_review_2023} for a more in-depth discussion.

Another recent trend is to use LLMs in an interactive, chat-style format.
This became popular through OpenAI's models \cite{openai_chatgpt_2023,openai_gpt4_2023} and is typically powered by finetuning on alignment data using reinforcement learning from human feedback~\cite{ouyang_training_2022}.
In a code-based setting, such interaction can, for instance, assist software development~\cite{lahiri_interactive_2023,google_code-chat_2023}.
Further, many recent works utilize interactive coding strategies to deploy LLMs as agents~\cite{yang_if_2024}.
For instance, Voyager~\cite{wang_voyager_2024} iteratively learns to master the game of Minecraft by letting an LLM code functions, and InterCode~\cite{yang_intercode_2023} connects an LLM to a Bash shell to solve file system task, similar to our use of an interactive Python console.
Recent benchmarks~\cite{liu_agentbench_2024,wang_mint_2024} will further catalyze this development.
We deploy such interactive coding strategy to real-world humanoid robotics, and enrich it with incremental learning from natural interactions. 

\begin{figure*}[h]
    \centering
    \includegraphics[width=1\linewidth]{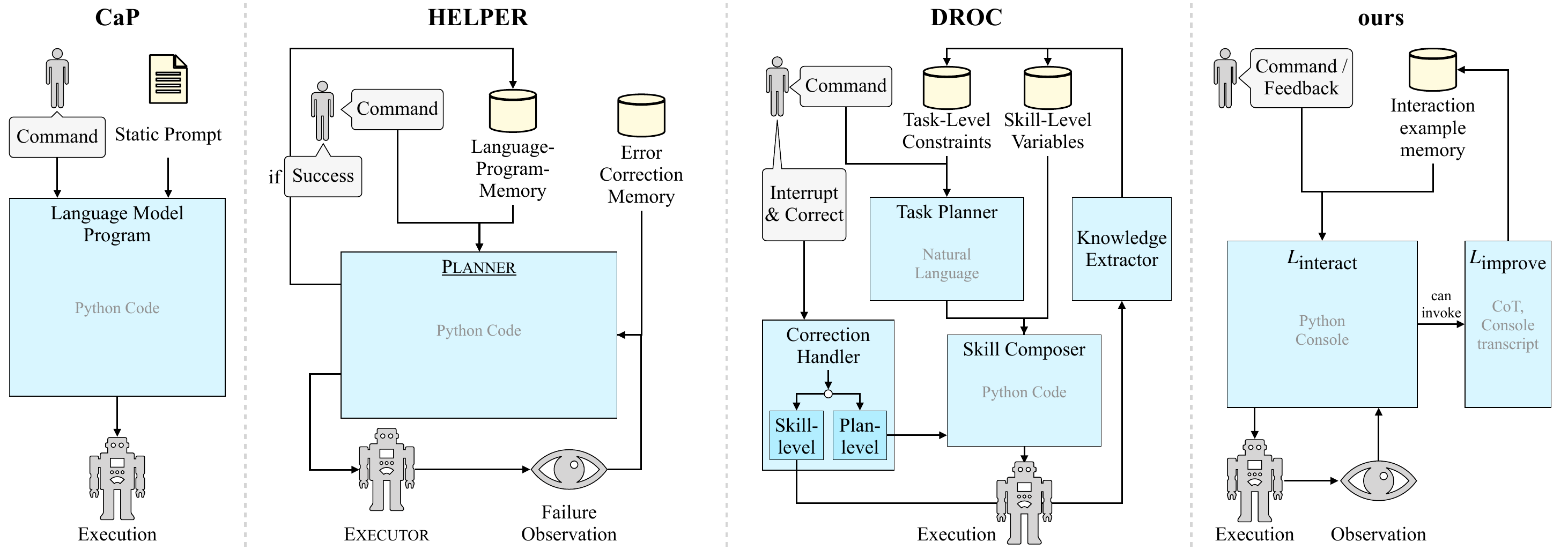}
    \caption{Comparison of Code as Policies \cite{liang_code_2023}, HELPER \cite{sarch_open-ended_2023}, DROC \cite{zha_distilling_2023} and our method, focusing on information flow from user input, observations, prompts, memories to LLM modules to robot execution, and how the methods learn from user interactions. Building on the interactive Python console prompting scheme, our method realizes incremental learning from natural interaction in a conceptually simple way.}
    \label{fig:related_work}
\end{figure*}

\subsection{Orchestrating Robot Behavior with LLMs}
\label{sec:related_work:llm_in_robotics}

Recently, many works extend the capabilities of LLMs by giving them access to external models, tools and APIs~\cite{mialon_augmented_2023,parisi_talm_2022,qin_tool_2023,wang_interactive_2023}.
Tool usage can also be combined with reasoning techniques such as CoT prompting~\cite{wei_chain_2022} to significantly improve planning~\cite{yao_react_2023}.
In particular, orchestrating robot behavior and thus interacting with the physical environment can be seen as an embodied special case of LLM tool usage.
Huang~\etal~\cite{huang_language_2022} initially proposed the idea to utilize world knowledge from LLM pretraining to map high-level tasks to executable mid-level action sequences.
SayCan~\cite{ahn_as_2022} fuses LLM output probabilities with pretrained affordance functions to choose a feasible plan given a natural language command.
Socratic Models~\cite{zeng_socratic_2023} combine visual and textual LLMs to generate instructions in the form of API calls, which are then executed by a pretrained language-conditioned robot policy.
Both Code as Policies (CaP)~\cite{liang_code_2023} and ProgPrompt~\cite{singh_progprompt_2023} demonstrate the usefulness of a code-generating LLM for robot orchestration, as they convert user commands to (optionally, recursively defined) policy code grounded in predefined atomic API calls.
While the generated policies can react to the robot's perception, these approaches do not directly involve the LLM in the online execution of a multi-step task after the policy has been generated.
In contrast, Inner Monologue~\cite{huang_inner_2022} feeds back execution results and observations into the LLM, but does not rely on code-writing, thus missing its combinatorial power.
KnowNo~\cite{ren_robots_2023} iteratively asks the LLM for a set of possible next steps, determines the LLM's confidence in each possibility using its output token distribution in a multiple-choice setup, and then uses conformal prediction to decide whether the system is sure how to proceed or should ask the user for help.
Recent technical reports~\cite{vemprala_chatgpt_2023,wake_chatgpt_2023} provide guidance on utilizing ChatGPT~\cite{openai_chatgpt_2023} for robot orchestration.
While TidyBot~\cite{wu_tidybot_2023} uses \mbox{GPT-3}~\cite{brown_language_2020} in a similar way to generate high-level plans for tidying up a cluttered real-world environment, the authors focus on personalization by summarizing and thereby generalizing individual object placement rules.

With our proposed emulated Python console prompting, we differ from these existing works by
\emph{(i)}~formatting and interpreting all interaction with the LLM as Python code, in contrast to~\cite{ahn_as_2022,huang_inner_2022}, 
\emph{(ii)}~closing the interaction loop by enabling the LLM to reason about each perception and action outcome,
in contrast to~\cite{liang_code_2023,singh_progprompt_2023,wake_chatgpt_2023,zeng_socratic_2023,ahn_as_2022}, 
\emph{(iii)}~allowing the LLM to decide when and which perception primitives to invoke, instead of providing a predefined list of observations (usually a list of objects in the scene) as part of the prompt as in~\cite{zeng_socratic_2023,huang_inner_2022,singh_progprompt_2023,liang_code_2023,wu_tidybot_2023}, 
and \emph{(iv)}~simplifying the task for the LLM by allowing it to generate one statement at a time, in contrast to~\cite{liang_code_2023, singh_progprompt_2023, vemprala_chatgpt_2023}.

\subsection{Dynamic Prompt Creation}

When prompting an LLM to perform a task, quality and relevance of the provided few-shot examples are key to the performance of the system.
Thus, several works propose to dynamically select these examples (\eg, from a larger training set) for constructing a useful prompt.
\citet{liu_what_2022} improve performance in a downstream question-answering (QA) task by selecting relevant few-shot samples via $k$-Nearest-Neighbor search in a latent space of pretrained sentence embeddings~\cite{reimers_sentence-bert_2019} representing the questions.
\citet{ye_compositional_2023} select not only the most similar, but also a diverse set of samples.
\citet{luo_dricl_2023} show that this dynamic prompt construction is also applicable for instruction-fine-tuned language models (LMs)~\cite{ouyang_training_2022} and in combination with CoT prompting.
\citet{song_llm-planner_2023} use top-$k$ retrieval for instructing an LLM to plan robotic tasks.
Similar to that approach, we apply vector embeddings of human utterances to find the top-$k$ examples which are most similar to the current situation.

Other works go further by proposing to update the database of examples by user interaction.
In~\cite{madaan_memory-assisted_2022}, GPT-3 is tasked with solving lexical and semantic natural language processing questions few-shot by generating both an understanding of the question as well as the answer.
A user can then correct an erroneous understanding to improve the answer, and such correction is stored in a lookup table for later retrieval on similar queries.
Similarly, user feedback can be used to improve open-ended QA by generating an entailment chain along with the answer, and allowing the user to then correct false model beliefs in that entailment chain~\cite{dalvi_mishra_towards_2022}.
Corrections are stored in memory and later retrieved based on their distance to a novel question.

In our work, we also propose to store corrective user feedback as interaction examples in the robot's memory.
However, we go even further by
\emph{(i)}~letting the LLM decide when such feedback is relevant (by invoking a certain function), 
\emph{(ii)}~generating new examples of improved behavior from the human's feedback and thus
\emph{(iii)}~treating prior knowledge and instructed behavior in a uniform way by treating both as interaction examples in the robot's memory.
The authors of~\cite{vemprala_chatgpt_2023} mention that ChatGPT can be used to change code based on high-level user feedback.
However, they do not combine this with incremental learning to persist the improved behavior.

Closest to our approach are the concurrent works DROC~\cite{zha_distilling_2023} and HELPER~\cite{sarch_open-ended_2023}, shown in \cref{fig:related_work}.
Similar to our learning from interaction, DROC~\cite{zha_distilling_2023} distills knowledge from problematic interactions and retrieves it later when solving new tasks.
While the goal and problem setting are similar, we differ by formulating the complete interaction in code, instead of separating task-level and skill-level into natural-language- and code-level interaction, respectively, and
also generalizing incremental learning as code manipulation, instead of explicitly memorizing task-level natural language constraints and skill-level variable assignments separately.
HELPER~\cite{sarch_open-ended_2023} retrieves few-shot examples for the LLM's prompt from a language-program memory similar to our interaction examples memory, and learns personalized robot behavior by extending the memory.
In contrast to our approach, they add examples only from successful episodes, and they have separate mechanisms for normal behavior and error correction.
We focus on learning from feedback in erroneous or suboptimal episodes, and we treat initial and follow-up instructions uniformly using the proposed Python console prompting.


\section{Approach}
\label{sec:methods}
In this section, we more precisely formulate the considered problem
and explain our approach to intuitive HRI and incremental learning of humanoid robot behavior using LLMs.

\newcommand{\suffixInit}{{0}}
\newcommand{\suffixCorrection}{{\mathrm{incr}}}

\newcommand{\instruction}{I}
\newcommand{\instructionInit}{\instruction_\suffixInit}
\newcommand{\instructionCorrection}{\instruction_\suffixCorrection}

\newcommand{\function}{F}
\newcommand{\numFunctions}{n}

\newcommand{\numInitialFunctions}{{n_\suffixInit}}
\newcommand{\numCorrectionFunctions}{{n_\suffixCorrection}}
\newcommand{\functionSet}{\mathcal{\function}}
\newcommand{\functionSetInit}{\functionSet_\suffixInit}
\newcommand{\functionSetCorrection}{\functionSet_\suffixCorrection}

\newcommand{\invocation}{f}
\newcommand{\numInvocations}{m}

\newcommand{\result}{r}
\newcommand{\worldstate}{W}

\newcommand{\history}{\mathcal{H}}

\newcommand{\ts}{t}
\newcommand{\currentHistory}{\history_\ts}

\newcommand{\memory}{\mathcal{M}}
\newcommand{\memoryInit}{\memory}

\newcommand{\stylefunc}[1]{\mathrm{#1}}
\newcommand{\func}[2]{\stylefunc{#1}\left( #2 \right)}
\newcommand{\tup}[1]{\left( #1 \right)}
\newcommand{\set}[1]{\left\{ #1 \right\}}

\newcommand{\functionGenerateName}{generate}
\newcommand{\functionLearnName}{learn\_from\_interaction}

\newcommand{\functionLearn}{\function_\mathrm{learn}}
\newcommand{\functionWait}{\function_\mathrm{wait}}

\subsection{Problem Formulation and Concept}

\begin{figure}[t]
    \centering
    \vspace{-0.05cm}
    \includegraphics[width=0.85\linewidth]{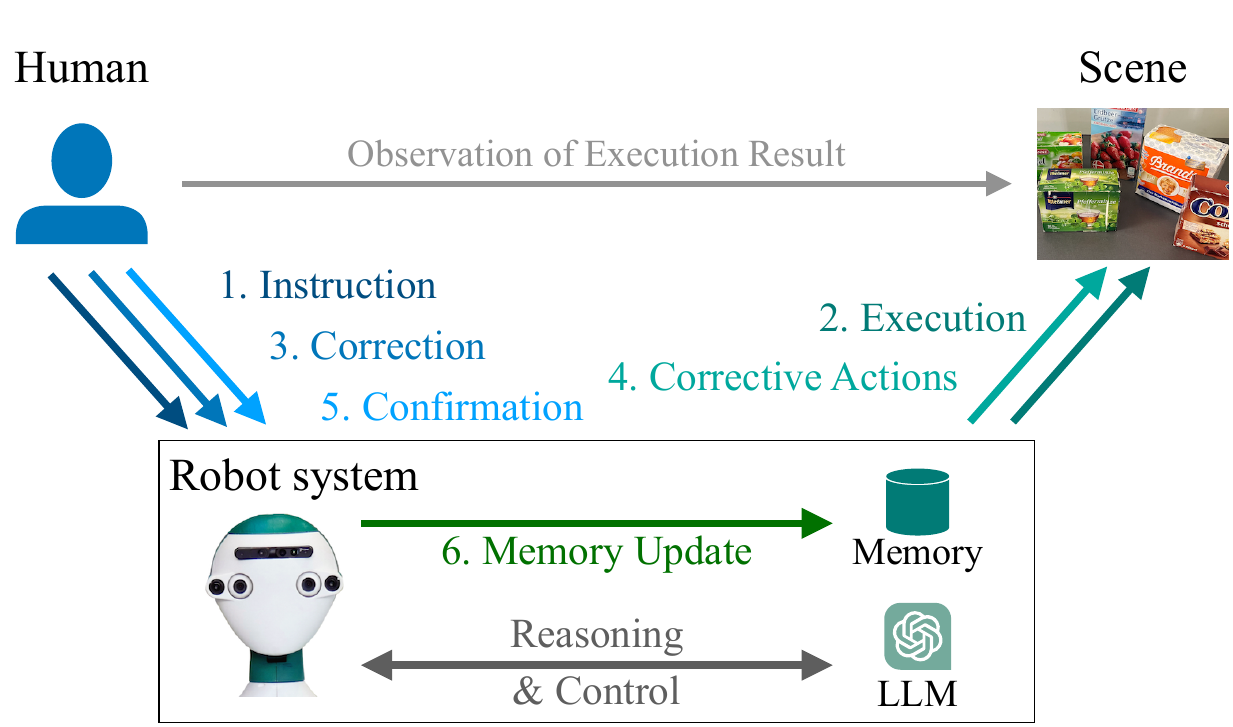}
    \caption{
    Incremental learning of robot behavior from interaction
    }
    \label{fig:interaction_scheme}
    \vspace{-0.1cm}
\end{figure}

In this work, we consider the problem of enabling a robot to interact with a human in natural language as depicted in \Cref{fig:interaction_scheme}.
First, the human gives a natural language instruction to the robot.
Then, the robot interprets the instruction and performs a sequence of actions.
However, the performed actions might be sub-optimal, incomplete or wrong.
In that case, the human instructs the robot how to improve or correct its behavior.
The robot executes further actions accordingly,
and if the human is satisfied with the result, they can confirm that the robot should memorize this behavior.
Finally, the robot must incrementally learn from the corrective instructions and avoid similar mistakes in the future.

\begin{figure*}[htb]
    \centering
    \includegraphics[width=0.9\linewidth]{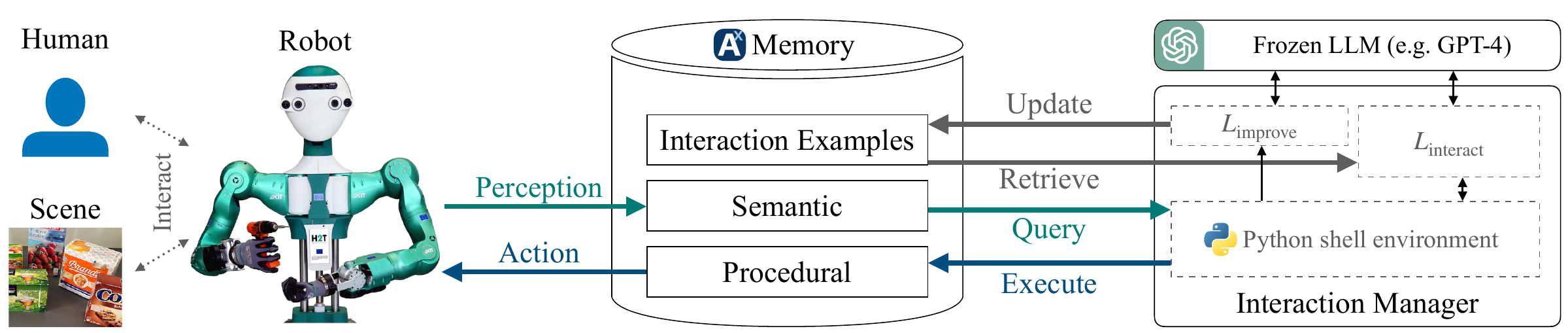}
    \caption{
    Conceptual view of our system. The robot's memory system~\cite{Peller-Konrad2023MemorySystemRobot}
    works as a mediator between the interaction manager and the robot system. The interaction LLM acts in a Python console environment. It can invoke functions to fetch the content of the current scene (as given by perception modules and stored in the memory) or invoke skills and thus perform robot actions. Relevant interaction examples are queried from the memory for few-shot prompting of the LLM. Incremental learning is performed by an improvement LLM updating the interaction examples memory with new content learned from instruction.
    }
    \label{fig:concept}
\end{figure*}

We formulate this problem as follows.
Consider a robot with a set of functions
$\functionSet = \left\{ \function_1, \dots, \function_\numFunctions \right\}$.
A function can be invoked to query the robot's perception or execute certain actions.
Further, let $\memory$ denote knowledge of interactions and behaviors as part of the episodic memory of the robot
which is initialized by prior knowledge.
Based on the initial instruction $\instructionInit$ and $\memoryInit$,
the robot must perform a sequence of function invocations
$(\invocation_1, \dots, \invocation_\numInvocations)$, 
where each invocation $\invocation_i$ consists of the invoked function $\function_i$ with its corresponding parameters.
Executing these invocations yields a sequence of results
$(\result_1, \dots, \result_m)$.
Overall, performing the task indicated by $\instructionInit$ 
results in an \emph{interaction history} $\history$ of the form
\begin{equation}
  \label{eq:perform}
  \history = \tup{ 
    \tup{\invocation_1, \result_1}, 
    \dots, 
    \tup{\invocation_\numInvocations, \result_\numInvocations}
  } \leftarrow \func{perform}{\instructionInit, \memoryInit} 
\end{equation}
Note that we explicitly allow executing a generated invocation right away (potentially modifying the world state~$\worldstate$) and using the result to inform the generation of the subsequent invocation.
Therefore, the current history 
$\history_{\ts} = \tup{ 
    \tup{\invocation_1, \result_1}, 
    \dots, 
    \tup{\invocation_{t}, \result_{t}} }$
is available when generating the next invocation $\invocation_{\ts+1}$,
\ie, for $t \in \set{0, \dots, \numInvocations - 1}$,
\begin{align}
  \label{eq:generate}
    \invocation_{\ts+1} &\leftarrow \func{\functionGenerateName}{
      \instructionInit, 
      \history_\ts,
      \memory
    }
    ,
    \\
    \label{eq:execute}
    \tup{\result_{\ts+1}, \worldstate_{\ts+1}}
    &\leftarrow \func{execute}{\invocation_{\ts+1}, \worldstate_\ts}
    ,
    \\
    \label{eq:history-concat}
    \history_{\ts+1} &\leftarrow \history_\ts \circ \tup{\tup{ \invocation_{\ts+1}, \result_{\ts+1}}}
    ,
\end{align}
where $\circ$ denotes sequence concatenation.
In other words, invocations are generated auto-regressively by reasoning over the memory, the instruction as well as the previous actions and their execution results.

To unify the subsequent notation, we define the human's instructions as a special case of perception, \ie, the system perceives them as a result of invoking the function $\functionWait \in \functionSet$.
Using that terminology, $\history_\suffixInit = \tup{\tup{\invocation_\mathrm{wait}, \instructionInit}}$, and we can drop $\instructionInit$ as explicit parameter of $\stylefunc{\functionGenerateName}$.
Similarly, further instructions are handled as part of the interaction history.

If the human gives an instruction to correct the robot's behavior, the robot must be able to learn from this instruction to improve its behavior in the future.
We model this capability as another function $\functionLearn \in \functionSet$.
Its purpose is to update the robot's interaction knowledge $\memory$ to learn from the corrective instructions and avoid the mistake in the future
\begin{equation}
    \label{eq:memory-update}
    \memory \leftarrow \func{\functionLearnName}{\memory, \history_{\ts} }
\end{equation}
where $\history_\ts$ is the interaction history when $\functionLearn$ is called.

To address this problem, we propose a system as depicted in \cref{fig:concept}. A humanoid robot is interacting with a human and the scene.
The robot is equipped with a multimodal memory system containing the following information about the current scene:
First, semantic knowledge about objects, locations and agents in the world.
Second, subsymbolic and symbolic knowledge about the current scene including object relations.
Third, the procedural memory of the robot, containing executable skills (in our case implemented through scripted policies).
An execution request sent to the procedural memory triggers physical robot actions.
The set of available functions~$\functionSet$ contains functions to query the semantic and procedural memory.
Finally, we implement $\memory$ as part of the episodic memory of the robot containing interaction histories~$\history$, \ie, short episodes of interactions between the human and the robot, including the natural language inputs, the actions executed by the robot, and their results.

\newcommand{\languageModel}{L}
\newcommand{\languageModelInteract}{\languageModel_\mathrm{interact}}
\newcommand{\languageModelImprove}{\languageModel_\mathrm{improve}}
\newcommand{\consoleEnv}{E}

The \emph{interaction manager} is responsible for the high-level orchestration of the robot's abilities.
It has access to two instances of LLMs,
an \emph{interaction LLM} $\languageModelInteract$
and an \emph{improvement LLM} $\languageModelImprove$,
as well as a Python console environment $\consoleEnv$ to execute generated function invocations.
$\languageModelInteract$ is prompted by the interaction manager with 
the available functions~$\functionSet$,
the current interaction history~$\currentHistory$,
as well as relevant few-shot examples retrieved from~$\memory$,
and generates function invocations~$\invocation$.
Following the notation of \cref{eq:generate,eq:execute}, the function
$\mathrm{\functionGenerateName}$ is implemented through $\languageModelInteract$, while the function $\mathrm{execute}$ is provided by $\consoleEnv$.
By generating an invocation of $\functionLearn \in \functionSet$, $\languageModelInteract$ can trigger \cref{eq:memory-update}.
We implement the function $\mathrm{\functionLearnName}$ by few-shot prompting $\languageModelImprove$.
It reasons over $\history_t$ and generates an improved version of the interaction, which is then saved to the memory $\memory$.

\subsection{Procedure Overview}
\label{sec:methods:procedure}
\begin{figure*}[htb]
    \centering
    \includegraphics[width=\linewidth]{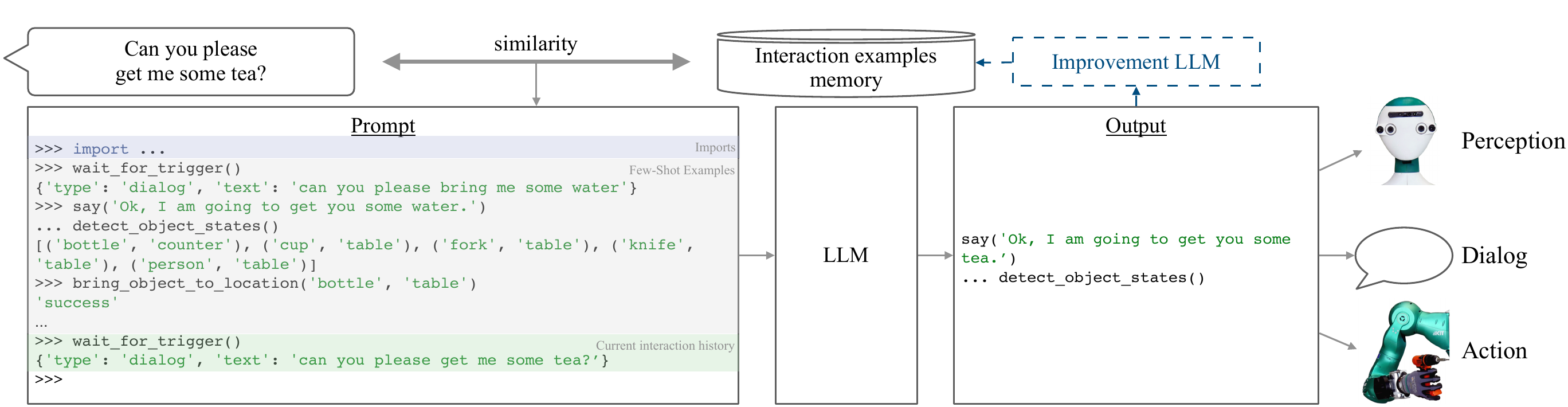}
    \caption{Overview of our method for incremental learning of robot behavior. We use an LLM (in our experiments, GPT-4~\cite{openai_gpt4_2023}) to control robot perception and action given a prompt of few-shot examples (bottom, \cref{sec:methods:python_console}). Prompts are constructed dynamically based on the similarity to the current user request (top left, \cref{sec:methods:prompting}). The interaction examples memory is initialized with prior knowledge, and then incrementally enriched by LLM-improved problematic interactions to learn from mistakes (top right, \cref{sec:methods:learning}).}
    \label{fig:llm-method-overview}
\end{figure*}

To start, we populate the memory $\memory$ with both prior knowledge (\ie, predefined interaction examples) and previously learned interaction examples.
The interaction manager sets up $\consoleEnv$ including $\functionSet$, and then invokes an initial $\functionWait = $ \inlinecode{wait\_for\_trigger()} inside that environment.
This call waits for dialog input and returns when the human gives an initial instruction.
The interaction manager handles any function return value by inserting its textual representation into the current interaction history, thus extending $\currentHistory$.
Thereby, it emulates the look of a Python console (\cref{sec:methods:python_console}).
In the following, a prompt is constructed (\cref{sec:methods:prompting}) based on $\functionSet$, the most relevant examples from $\memory$, and $\history_\ts$.
This prompt is passed to $\languageModelInteract$ to produce the next command(s).
The generated code is executed within $\consoleEnv$, and both the code and its return values are again inserted into $\currentHistory$.
The interaction manager repeats this process as the high-level behavior-driving loop of the robot (see \cref{fig:llm-method-overview}).
Note that $\languageModelInteract$ can listen to further user utterances by generating \inlinecode{wait\_for\_trigger()} again.
Our proposed prompt-based incremental learning strategy (\cref{sec:methods:learning}) is also invoked by $\languageModelInteract$ itself when it calls $\functionLearn = $ \inlinecode{learn\_from\_interaction()}.

\subsection{LLM interacting with an Emulated Python Console}
\label{sec:methods:python_console}
The left of \cref{fig:llm-method-overview} shows an  interaction example using our proposed prompting scheme emulating a Python console.
All commands entered into the emulated console (lines starting with \inlinecode{>>>} or \inlinecode{...}) are to be generated by the LLM, while the function return values are inserted below each invocation.
The proposed syntax enables a closed interaction loop so that the LLM can dynamically react to unexpected situations and errors, while also keeping the flexibility of coding non-trivial statements.
We achieve this by setting \inlinecode{>>>} to be the stop token when prompting the LLM.
This means that the LLM can generate continuation statements (including control flow and function definitions) by starting a new line with \inlinecode{...}.
Since generation stops at the beginning of the next statement, the LLM's output will also include the expected outcome of its own command, which we discard for the scope of this work.

During our experiments, we observed that it is important for functions to provide semantically rich error messages, including hints on how to improve.
This leads to self-correcting behavior~\cite{skreta_errors_2023}.
For instance, when calling \inlinecode{move\_to} with an invalid or underspecified location such as \enquote{counter,} we pass the error message \inlinecode{Invalid location. Use one of the locations returned by list\_locations()} to the LLM.
In this example, the error message guides the LLM to query a list of possible locations which are then used to correctly ground the natural language request to the name \inlinecode{inFrontOf\_mobile-kitchen-counter\_0} that the \inlinecode{move\_to} function understands. 

Analogously to Code as Policies \cite{liang_code_2023}, we dynamically generate non-existing functions the LLM tries to use.
Specifically, when $\languageModelInteract$ generates code that refers to an undefined function, we invoke another LLM $L_{fgen}$ that is prompted to define the function, given the line of code that is using it as context.
For $L_{fgen}$, we exactly follow the method of \cite{liang_code_2023}, including recursive function generation.
The generated function is then inserted into the emulated Python console \emph{before} the statement that referred to the undefined function, and then that statement is executed.
The purpose of inserting the function definition into the execution history is that it is thereby accessible to user feedback and can be improved upon by incremental learning.

\begin{listing*}[!t]
\footnotesize
\vspace{0.15cm}
\begin{mdframed}
\setlength{\columnseprule}{0.1pt}
\def\columnseprulecolor{\color{black}}
\begin{minipage}{\linewidth}
\begin{multicols}{2}
\textbf{Prompt}: I had the following interaction with the robot:
{\codesize
\begin{minted}{python}
>>> wait_for_trigger()
{'type': 'dialog', 'text': 'I want to have some juice'}
>>> list_object_locations(affordance='grasp')
# ...
>>> move_to(get_location_coordinates('kitchen-counter_0'))
'success'
>>> grasp('multivitamin-juice_0')
'success'
>>> move_to(get_location_coordinates('handover_to_human'))
'success'
>>> say("Here is the multivitamin juice.")
>>> handover_object_to_human('multivitamin-juice_0')
'success'
>>> wait_for_trigger()
{'type': 'dialog', 'text': 'can you also bring me a cup?'}
>>> say('Ok, I will bring you a cup')
# ... [analogous to the above]
>>> wait_for_trigger()
{'type': 'dialog', 'text': 'Thanks, but next time, '
 'can you do this both at once? You have two hands'}
>>> learn_from_interaction()
\end{minted}
}
\textbf{Prompt}: What is the problem in this interaction? Answer with a single sentence.\\
\columnbreak
\textbf{LLM}: The problem is that the robot is not utilizing both of its hands to bring the juice and the cup at the same time, resulting in two separate trips.\\
\textbf{Prompt}: How can the robot do better next time? Answer with a single explanation sentence, no code.\\
\textbf{LLM}: Next time, the robot should use both of its hands to grasp the juice and the cup at the same time, reducing the number of trips and increasing efficiency.\\
\textbf{Prompt}: Provide an improved version of the interaction transcript. [...]\\
\textbf{LLM}:
{\codesize
\begin{minted}{python}
>>> wait_for_trigger()
{'type': 'dialog', 'text': 'I want to have some juice'}
# ... [LLM lists objects & locations and moves to counter]
>>> grasp('multivitamin-juice_0', 'left')  # with left hand
'success'
>>> grasp('cup_large_0', 'right')  # with right hand
'success'
>>> move_to(get_location_coordinates('handover_to_human'))
'success'
>>> say("Here is the multivitamin juice and the cup.")
>>> handover_object_to_human('multivitamin-juice_0')
'success'
>>> handover_object_to_human('cup_large_0')
'success'
>>> wait_for_trigger()
{'type': 'dialog', 'text': 'Thanks, that was much quicker'}
\end{minted}
}
\end{multicols}
\end{minipage}    
\end{mdframed}
\caption{Example of the LLM-transcript generated by a \inlinecode{\scriptsize learn\_from\_interaction()} call. The parts starting with \textbf{LLM} are generated by the LLM, while the \textbf{Prompt} parts are fixed prompts (and the input code snippet to improve). Full prompt including few-shot examples in \cref{app:full_learn_from_interaction_bimanual}}
\label{listing:learn_from_interaction}
\end{listing*}

\subsection{Dynamic Prompt Construction}
\label{sec:methods:prompting}
We dynamically construct the prompt for $\languageModelInteract$ depending on the current interaction history $\currentHistory$ (\ie, the code statements, execution results and user inputs observed so far).
We start with some predefined base prompt, stating the general task and \enquote{importing} all defined names and functions.
These imports are generated dynamically given the symbols defined in $E$, \ie, the available functions $\functionSet$.
The second part of the prompt consists of few-shot examples.
For this, we make use of a memory $\memory$ of coding interaction examples, where each entry follows the Python console syntax defined in \cref{sec:methods:python_console}.
$\memory$ is initialized with hand-written prompts, but later extended dynamically as explained in \cref{sec:methods:learning}.
Given the current interaction history $\currentHistory$, we define a similarity measure $S(\history, \currentHistory)$, see below, for each $\history \in \memory$ and choose the top-$k$ $\history$ to become part of the actual prompt.
Afterwards, $\currentHistory$ itself is inserted into the prompt to provide the LLM with the current context.
Finally, the prompt is completed by inserting a syntax trigger for the LLM to correctly generate the next command, \ie, \inlinecode{>>>}.
An example can be seen on the left of \cref{fig:llm-method-overview}.

\newcommand{\embed}[1]{\func{E}{#1}}

To implement the similarity function $S(\history, \currentHistory)$, we assume that examples with comparable natural language instructions are helpful.
Therefore, we extract all such instructions from $\currentHistory$ and each $\history \in \memory$.
Let $\instruction_\ts^i$ with $i = 1,\dots, N$ denote the $N$ most recent instructions in $\currentHistory$ (where $\instruction_\ts^1$ is the most recent one), and $\instruction_\history^j$ with $j=1, \dots, M_\history$ all the $M_\history$ instructions found in each $\history \in \memory$.
We make use of a pretrained sentence embedding model~\cite{reimers_sentence-bert_2019} to measure the semantic similarity $\func{sim}{a, b} = \embed{a} \cdot \embed{b}$ between two natural language sentences $a, b$ by the dot product of their latent space embeddings $\embed{\cdot}$.
First, we compute a latent representation of $\currentHistory$ as
\begin{equation}
    e_\ts = \sum_{i=1}^N \gamma^{i-1} \embed{\instruction_\ts^i} \label{eq:prompting:e_t}
\end{equation}
where $\gamma = 0.6$ is an empirically chosen decay factor.
Then, we determine a score $\alpha_\history^j$ for each instruction $\instruction_\history^j$ of each history $\history \in \memory$ as given by
\begin{equation}
    \alpha_\history^j = e_\ts \cdot \embed{\instruction_\history^j} \label{eq:prompting:alpha}
\end{equation}
The final similarity score is given by $S(\history, \currentHistory) = \max_j \alpha_\history^j$, and we pick the top-$k$ such $\history$ as the few-shot examples for the prompt.

\subsection{Incremental Prompt Learning}
\label{sec:methods:learning}

To enable our system to learn new or improved behavior from user interaction, we propose to make $\memory$ itself dynamic.
For this purpose, we introduce a special function $\functionLearn =$ \inlinecode{learn\_from\_interaction()}.
This function is always \enquote{imported} in the base prompt, and there are predefined code interaction examples $\history_\text{learn} \in \memory$ involving this call.
These $\history_\text{learn}$ will be selected by dynamic prompt construction if semantically similar situations occur.
They involve failure situations, where the user has to tell the robot what and how to improve, and that it should do better next time.
Thus, when a mistake occurs and the user complains, these examples will be selected for the prompt and $\languageModelInteract$ is biased towards invoking $\functionLearn$.

\newcommand{\improvedHistory}{\currentHistory^*}

To implement learning from an erroneous interaction $\currentHistory$, we query $\languageModelImprove$ in a CoT-manner to identify and fix the problem.
Specifically, we provide $\currentHistory$ and first ask for a natural language description of the problem in this interaction.
Subsequently, we request $\languageModelImprove$ to explain what should be improved next time.
Finally, $\languageModelImprove$ is asked for an improved version $\improvedHistory$ of the interaction (in the given Python console syntax), and $\improvedHistory$ is added to the memory $\memory$.
That way, the next time a similar request occurs, $\improvedHistory$ will be selected by dynamic prompt construction, and $\languageModelInteract$ is biased towards not making the same mistake again.
An example LLM transcript of such $\functionLearn$ implementation can be found in \cref{listing:learn_from_interaction}.
For robustness, there are three cases where we discard the generated $\improvedHistory$:
First, we ignore the call to $\functionLearn$ if it does not follow immediately after a user utterance, since we only want to learn from explicit human feedback.
Second, we abort the learning if the response to the first CoT request is that there is no problem.
Third, if $\improvedHistory$ is equal to the input interaction $\currentHistory$, we discard~it.


\section{Simulated Evaluation}
\label{sec:results}

\subsection{Experimental Setup}
To quantitatively assess the performance of our method, we utilize the evaluation protocol from Code as Policies \cite{liang_code_2023}, involving a simulated tabletop environment with a UR5e arm and Robotiq 2F85 gripper manipulating a set of blocks and bowls of ten different colors.
We use their seven seen and six unseen instructions (SI/UI), where each instruction is a task with placeholders that are filled with attributes (e.g. \enquote{pick up the $<$\textit{block}$>$ and place it on the $<$\textit{corner}$>$}).
The set of possible attribute values is also split into seen and unseen attributes (SA/UA).
For more details, refer to \citet{liang_code_2023}.

As our focus is on incremental learning from natural-language interaction, our methodology involves human supervision as follows:
We first set up a randomly generated scene and pass the instruction to the evaluated system.
The system generates some code that utilizes the same API as in \cite{liang_code_2023} (cf. \cref{app:cap_eval}).
During code execution, the human observes the robot’s actions by watching the simulation rendering.
Each run can result in
success (goal reached),
failure (goal not reached),
error (system threw unhandled exception), or
timeout (e.g. system got stuck in a loop).
The latter two lead to immediate termination of the experiment.
In contrast, when the system yields control normally (after code execution for CaP and on $\functionWait$ for our method), the resulting world state is checked using scripted ground-truth evaluation functions, leading to either success or failure outcome.
The human is then presented with this outcome and has the option to provide feedback or improvement instructions to the robot, which are again passed to the system.
The success detection is performed every time the system yields control, and the sequence of states and user interactions is recorded.
Note that we allow user feedback even when already in success state, as the execution might still have been suboptimal and the human may want to provide feedback to learn from for next time.
Details and example interactions can be found in \cref{app:cap_examples}.

Every task is repeated ten times using randomly generated scenes, and each run is performed in sequence, i.e., the interaction memory is not reset between runs in order to allow for incremental learning.
To assess the results, we compute the following metrics from the execution traces:
\begin{itemize}
    \item[$s$] is the turnout success rate, i.e. the percentage of runs that ended in success state (optionally after user interaction that clarifies the goal or helps the system)
    \item[$i$] is the initial success rate, i.e. the percentage of runs that yielded a successful state on the first system return, i.e. where no user interaction was required to reach success
    \item[$n$] counts the number of user interactions that were required until the success state was first reached. For runs that count into the initial success category, $n = 0$, while for non-successful runs, $n$ is undefined. When aggregating $n$, we average only over the runs that ended successfully.
\end{itemize}

\subsection{Baselines \& Methods}

\textbf{CaP}: 
    We utilize the prompts provided by \citet{liang_code_2023}.
    This is equivalent to our system without incremental learning and without the interactive console formatting.
    Specifically, we note that CaP has no way of feeding back coding errors to the system, i.e. it fails immediately if the generated code is syntactically invalid or throws an exception.

\textbf{HELPER}:
    We adapt the code and prompts provided by \citet{sarch_open-ended_2023} to the simulated tabletop evaluation scenario \& API.
    For few-shot example retrieval, we set $k=16$ for a fair comparison.
    Specifically, we feed back execution errors to the \textit{Self-Reflection \& Correction} prompt, and user feedback is passed as a new command to the \textsc{Planner}.
    HELPER's few-shot memory is expanded with successful trials.
    Further details can be found in \cref{app:cap_eval:helper}.
    
\textbf{ours}: 
    This is our full system with incremental learning and a value of $k=16$ for few-shot sample retrieval.
    We split and translated the 16 samples from the CaP prompts into our interactive console syntax to initialize the memory of interaction examples.
    Furthermore, there are two very short samples that demonstrate when to call $\functionLearn$.
    
\textbf{ours w/o learning}:
    This is our system, but without incremental learning. 
    $k=16$ means that all samples are used, as the interaction examples memory is static.
    
\textbf{ours w/o retrieval}:
    This is our system with incremental learning but a very high value of $k=64$ for few-shot sample retrieval, which effectively is a system that does not use retrieval.
    Note that the prompt construction is still dynamic as the order of the samples is determined by the similarity to the current request (cf. \cref{sec:methods:prompting}).

Furthermore, we compare the differently capable LLMs \texttt{gpt-3.5-turbo-0301} and \texttt{gpt-4-0613} of the OpenAI API~\cite{openai_chatgpt_2023,openai_gpt4_2023}. 
For $L_{improve}$, we always use \texttt{gpt-4}.
We note that the original CaP numbers \cite{liang_code_2023} were reported with the \texttt{codex} model~\cite{chen_evaluating_2021} that is no longer available.
We reproduce their experiments with the newer models but did not perform further prompt tuning, therefore our success rates for CaP are lower than those reported in \cite{liang_code_2023}.
Specifically, \texttt{gpt-3.5} sometimes generates natural language responses instead of code, which causes CaP to fail with a SyntaxError.

\subsection{Results}

\begin{table}
    \centering
    \setlength\tabcolsep{2px}
    \renewcommand{\arraystretch}{1.1}
    \begin{tabular}{ll|rr|rr|rr|rr|rr}

        & & \multicolumn{6}{c|}{\textbf{ours}} & \multicolumn{2}{c|}{HELPER} & \multicolumn{2}{c}{CaP} \\\cline{3-8}
        & & \multicolumn{2}{c|}{\textbf{full}} & \multicolumn{2}{c|}{w/o retrieval} & \multicolumn{2}{c|}{w/o learning} & \multicolumn{2}{c|}{} & \multicolumn{2}{c}{} \\
    
        & Test & $s$ & $i$ & $s$ & $i$  & $s$ & $i$  & $s$ & $i$ & $s$ & $i$ \\\hline
        
        \multirow{3}{*}{\rotatebox[origin=c]{90}{GPT-4}}
                & SA SI & 100  & 97.5 & 97.5 & 90.0 & 98.8 & 90.0 & 97.5 & 87.5 & 85.0 & 71.2 \\
                & UA SI & 100  & 92.5 & 98.8 & 95.0 & 98.8 & 92.5 & 100  & 93.8 & 96.2 & 81.2 \\ 
                & UA UI & 93.3 & 85.0 & 91.7 & 81.7 & 91.7 & 78.3 & 91.7 & 81.7 & 53.3 & 35.0 \\\hline

        \multirow{3}{*}{\rotatebox[origin=c]{90}{GPT-3.5}} 
                & SA SI & 95.0 & 87.5 & 93.8 & 82.5 & 85.0 & 43.8 & 93.8 & 77.5 & 53.8 & 52.5 \\
                & UA SI & 97.5 & 86.2 & 96.3 & 88.8 & 80.0 & 45.0 & 87.5 & 71.2 & 60.0 & 58.8 \\
                & UA UI & 85.0 & 70.0 & 56.7 & 51.7 & 66.7 & 43.3 & 80.0 & 50.0 & 16.7 & 15.0 \\\hline

    \end{tabular}
    \caption{Evaluation results on simulated tabletop tasks: success rate $s$ and initial success rate $i$}
    \label{tab:cap_results_both_si}
\end{table}

\begin{table}
    \centering
    \setlength\tabcolsep{4px}
    \renewcommand{\arraystretch}{1.1}
    \begin{tabular}{ll|r|r|r|r|r}

        & & \multicolumn{3}{c|}{\textbf{ours}} & HELPER & CaP \\\cline{3-5}
        & Test & \textbf{full} & w/o retrieval & w/o learning  & & \\\hline

        \multirow{3}{*}{\rotatebox[origin=c]{90}{GPT-4}}
                & SA SI & 0.04 & 0.12 & 0.37 & 0.21 & 0.26 \\
                & UA SI & 0.14 & 0.12 & 0.1  & 0.1  & 0.35 \\ 
                & UA UI & 0.16 & 0.18 & 0.55 & 0.22 & 0.74 \\\hline

        \multirow{3}{*}{\rotatebox[origin=c]{90}{GPT-3.5}} 
                & SA SI & 0.14 & 0.25 & 1.09 & 0.31 & 0.02 \\
                & UA SI & 0.33 & 0.15 & 0.95 & 0.38 & 0.06 \\
                & UA UI & 0.28 & 0.19 & 1.29 & 0.68 & 0.07 \\\hline

    \end{tabular}
    \caption{Evaluation results on simulated tabletop tasks: average number of interactions until success $n$}
    \label{tab:cap_results_both_n}
\end{table}

\Cref{tab:cap_results_both_si,tab:cap_results_both_n} present the aggregated results of our experiments, while further details can be found in \cref{app:cap_results}.
From these results, we draw the following main insights:

\textbf{Interactive feedback helps to achieve success}. 
For all methods, $s$ is notably above $i$, which means that $L_{interact}$ effectively uses human feedback to improve its behavior. 
This effect is less stressed for CaP with \texttt{gpt-3.5}, as it often immediately fails with an error, thus not allowing for further interaction.

\textbf{Incremental learning reduces necessity of corrective interactions}. 
For many tasks, $i$ is notably higher and $n$ lower when comparing systems with learning to systems without learning, indicating that the feedback from earlier (failed) attempts is effectively utilized to improve following executions of the same task. 
This effect is also confirmed by \cref{fig:appendix:per_round_gpt4,fig:appendix:per_round_gpt35} in the appendix. 
While for \texttt{gpt-4} on seen instructions, performance is already on a high level and corrections are rarely necessary, the numbers strongly support the that incremental learning reduces interactions for unseen instructions, as well as for \texttt{gpt-3.5} on all instructions.
Thus, our method for incremental learning is especially useful for \enquote{hard} tasks with respect to the predefined examples and general capabilities of the used model.

\textbf{Incremental learning improves in-task success rate}. 
Our systems with incremental learning also have higher $s$ than those without learning. 
The reason is that our incremental learning method reflects on the erroneous behavior and generates a new sample for in-context learning that demonstrates the desired behavior. 
With such nearly identical demonstration, the generalization to new situations is much better, thus causing fewer errors that cannot be corrected through interaction.

\textbf{Incremental learning generalizes to new tasks}. 
Qualitatively, we observed several cases where a correction for one task is useful for another task as well. For instance, \texttt{gpt-3.5} initially interprets \enquote{the corner} as some position like $(0.1, 0.9)$.
When instructing to \enquote{put it right into the corner without any margin}, the behavior of using full numbers, e.g. $(0, 1)$, transfers to subsequent different tasks that also involve corners. 
Quantitatively, this effect is entangled with the previous points in higher $s$ and $i$, especially for the later unseen tasks.
For a further investigation, see \cref{app:cap_results}.

\textbf{Demonstration retrieval improves performance}. 
For both LLMs, our system with retrieval outperforms the system that always uses all samples. 
This is especially true for \texttt{gpt-3.5}, as the system without retrieval accumulated to many interaction examples in its memory in the final experiments, thus leading to immediate failure due to exceeding the LLMs token limit. 
While this is not the case for \texttt{gpt-4} with its much larger context length, the performance of the system with retrieval is still better. 
We hypothesize that this is due to too many irrelevant samples distracting the LLM. 

\textbf{Better LLMs lead to better performance}. 
This can be clearly seen when comparing the numbers for \texttt{gpt-4} and \texttt{gpt-3.5}. 
Nonetheless, we emphasize that \texttt{gpt-3.5}'s performance as $\languageModelInteract$ is still reasonably well, while it is faster and a factor of ten times cheaper.
Specifically, the total cost to perform the experiments in \cref{tab:cap_results_both_si} was $\$\,245.6$ for \texttt{gpt-4} vs. $\$\,19.8$ for \texttt{gpt-3.5} (which includes the use of \texttt{gpt-4} for $\languageModelImprove$).
Our method of incremental learning can thus be seen as a knowledge distillation method, with \texttt{gpt-4} as the expensive teacher model $\languageModelImprove$ generating task-specific new prompts for the cheaper \texttt{gpt-3.5} to improve its future behavior as $\languageModelInteract$.

\textbf{Comparison with HELPER.}
As a key difference to our method, HELPER learns from successful trials by storing them as an example, while our method only inspects erroneous experiences and then stores improved versions thereof.
The experimental results show that this strategy is more effective, leading to higher $s, i$ and lower $n$.
Furthermore, HELPER cannot see its own previously-generated code when responding to errors or feedback, in contrast to our method, which utilizes the interactive Python console prompting for this purpose.
Thus, HELPER cannot handle feedback such as \enquote{slightly more to the left} effectively.

\textbf{Further results.}
\Cref{app:cap_results} presents three additional experiments:
First, we investigate the effect of $k$ by setting $k = 4$ (instead of 16), showing that lower $k$ comes with a higher $n$ and lower $i$, as potentially relevant demonstrations might not be retrieved, thus requiring another user interaction.
Second, we change the behavior of $\functionLearn$ to just save the current interaction in $\memory$, skipping $\languageModelImprove$.
This hurts performance, as the erroneous behavior from previous trials is often repeated, despite the prompt containing improvement instructions from earlier interactions.
Third, we compare with \enquote{Dynamic CaP}, a more competitive version of CaP enhanced with few-shot example retrieval and learning from successful experiences, similar to HELPER.
These results confirm that our method, in particular interactive Python console prompting and incremental learning from erroneous interactions, is more effective.

\section{Real-World Demonstration}
\label{sec:demonstration}

\newcommand{\scenarioNameDrinkCup}[0]{Improving Plans}
\newcommand{\scenarioNameLadder}[0]{Learning User Preferences}
\newcommand{\scenarioNameSpeed}[0]{Adapting Low-Level Parameters}

To demonstrate the utility of our proposed prompt-based incremental learning technique, we perform experiments on the real-world humanoid robot \armarVI~\cite{Asfour2018ARMAR6CollaborativeHumanoid}.
We first provide challenging commands which the LLM initially solves incompletely or wrong. 
Then, the human interactively provides feedback and tells the robot how to improve.
Afterwards, we not only provide the same command again to check for improved behavior, but -- in order to study generalization -- also try similar commands that initially (\ie, before learning) led to similar mistakes.
Details on the implementation of these experiments, especially on the API exposed to the LLM, can be found in \cref{app:demo:implementation}.
The system is connected to a memory-centric cognitive robot architecture where the memory mediates between high-level components and low-level abilities (see \cref{fig:concept}).
Specifically, the API provided to the LLM allows querying the robot's memory which may have been filled beforehand by the robot's perception and cognition components. Further, the API allows to invoke registered skills, behaviors and movements of the robot.
In our experiments, we use a mixture of predefined prior knowledge (e.g., about static objects in the scene) and online perception (e.g. object pose-detection, self-localization).
However, we do not focus on scenarios where the involved skills themselves fail, but rather on high-level semantic problems.
We present three scenarios:
\emph{\scenarioNameDrinkCup} to demonstrate complex improvement of suboptimal or unintended performance, 
\emph{\scenarioNameLadder} to show how to adapt to non-generic task constraints, 
and \emph{\scenarioNameSpeed} to demonstrate that our system can learn from vague user instructions.

\subsection{Improving Plans}
In this scenario, we tell the robot that we want juice.
The prior knowledge contains some similar interaction examples, picking up a single object and handing it over to the human. 
Thus, the task of bringing the juice is executed successfully.
However, since the user needs a cup to drink, we further instruct the robot \enquote{can you also bring me a cup?}, which causes the robot to additionally hand over a cup.
Afterwards, we ask the robot to improve this behavior using \enquote{Thanks, but next time, can you do this both at once? You have two hands}.
$\languageModelImprove$ generates an improved interaction example as shown on the right of \cref{listing:learn_from_interaction} (simplified, cf. \cref{app:demo:bimanual}).

Afterwards, when giving the same initial command again, the robot uses bimanual behavior to hand over both juice and cup.
Furthermore, the learned bimanuality generalizes to \enquote{can you bring something to drink to the table?}, which does not use handover, but places both objects on the table.
Unfortunately, a further test with \enquote{can I have some milk, please?} shows the unimanual behavior again, so we again have to ask for a cup and trigger incremental learning.
In the next session, we ask \enquote{hey, can you serve some drink?}, which correctly generalizes the behavior to use both hands to pick up a different drink and cup, but misinterprets \enquote{serve} as doing a handover instead of putting it on the table.
However, we can successfully trigger learning again by teaching \enquote{when I say serve, I mean that you should put it on the table}, and subsequent requests do behave as intended.

We conclude that our interactive, incremental learning system can flexibly generate complex behavior from concise improvement instructions.
However, it is still challenging to robustly generalize from a single instruction to all cases a human might have intended, as shown by the milk example, where a second correction was necessary for successful generalization.
Improving this generalization capability should be a focus of future work.

\subsection{Learning User Preferences}
As shown in \cref{fig:interaction_example}, in this scenario we ask the robot to assist with cleaning the top of the fridge.
The memory~$\memory$ contains predefined comparable examples for cleaning the table and kitchen counter, which guide the LLM to only handing over the sponge to the human.
However, since the top of the fridge is higher than the table or the kitchen counter, we require a ladder to reach it so we additionally ask for it (\texttt{gpt-4} did, in contrast to \texttt{gpt-3.5}, proactively ask whether it should also bring the ladder).
The robot then successfully places the ladder in front of the fridge.
Eventually, we instruct the robot to always bring the ladder when working on high surfaces.
The generated improved interaction example correctly brings the ladder after the sponge, without further request (details in \cref{app:demo:ladder}).
Afterwards, when we perform a similar request (\eg, \enquote{clean on top of the dishwasher}), the robot brings both the sponge and the ladder successfully,
while for lower surfaces (\eg, kitchen counter) the robot still brings only the sponge.
The behavior also transfers to different tasks than cleaning, e.g. the robot brings the cereals and the ladder on \enquote{can you get me the cereals, I want to put it in the topmost shelf}, while it does not bring the ladder when tasked with \enquote{I want to put the cereals into the shelf}.

In summary, this example demonstrates that our method can be used to learn task constraints or preferences that a user specifies, and this knowledge can be generalized to similar situations.

\subsection{Adapting Low-Level Parameters}
In this scenario, we ask the robot to bring some object from the table to the workbench (details in \cref{app:demo:speed}).
Subsequently, we say \enquote{remember that the route from the table to the bench is save, you can go faster}.
$\functionLearn$ correctly generates a sample that adapts the numeric speed factor of the \texttt{move\_to} function on that route.
However, if we test the same task afterwards, $\languageModelInteract$ still uses the default speed.
Annoyed by that, we shout \enquote{you forgot that I told you to go faster from the table to the workbench. When moving on that route, you should go faster!}, triggering another learning process, generating another correct sample, including an explicit comment:
\begin{mdframed}
{\codesize
\vspace{0.15cm}
...
\begin{minted}{python}
>>> grasp('sponge_0')
'success'
>>> # The user earlier asked me to move faster from the 
    # table to the workbench, so let's do that
... move_to(get_location_coordinates('workbench_0'), 
            speed_factor=2.0)
'success'
>>> place_object('sponge_0', 'workbench_0')
\end{minted}
\vspace{-0.3cm}
...
\vspace{0.15cm}
}
\end{mdframed}
Proceeding requests now behave correctly and increase the speed from the table to the workbench.
However, an adversarial test shows that $\languageModelInteract$ does now dangerously use increased speed from another location to the workbench, too, while routes to different places still correctly use the default speed.

To conclude, our system can successfully learn to adapt low-level API parameters as requested by a user, but ensuring the LLM applies learned knowledge in the intended context only is not fully solved yet.


\section{Conclusion \& Discussion}
\label{sec:conclusion}
We present a system for integrating an LLM as the central part of high-level orchestration of a robot's behavior in a closed interaction loop.
Memorizing interaction examples from experience 
and retrieving them based on the similarity to the current user request allows for dynamic construction of prompts and enables the robot to incrementally learn from mistakes by extending its episodic memory with interactively improved code snippets.
We describe our implementation of the system in the robot software framework \armarx~\cite{vahrenkamp_robot_2015} as well as on the humanoid robot \armarVI~\cite{Asfour2018ARMAR6CollaborativeHumanoid}.
The usefulness of our approach is evaluated both quantitatively on the tasks from Code as Policies~\cite{ahn_as_2022} and qualitatively on a humanoid robot in the real world.

While the proposed method, in particular the incremental prompt learning strategy, shows promising results, there are still many open questions for real-world deployment.
First of all, the performance of LLMs is quite sensitive to wording in the prompt, thus sometimes leading to unpredictable behavior despite only slight variations of the input (\eg, adding \enquote{please} in the user command).
This might be solved with more advanced models in the future, as we did observe this issue much more often with GPT-3.5 than with GPT-4.
Investigating the effect and performance of example retrieval in dynamic prompt construction might also contribute to improving robustness.
Furthermore, our incremental prompt learning strategy should be expanded to involve additional human feedback before saving (potentially wrong) interaction examples to the episodic memory.
However, it is unclear how this can be accomplished if the user is not familiar with robotics or programming languages.
To rigorously evaluate our incremental learning method in the real world, future work may want to incorporate a user study with non-technical participants.
Further work should also focus on abstraction of similar and forgetting of irrelevant learned behavior.
While our system is limited by the APIs exposed to the LLM, it could be combined with complementary approaches \cite{parakh_lifelong_2023} to support learning of new low-level skills.
Moreover, although we provide the LLM with access to perception functions and examples of how to use them, it sometimes comes up with non-grounded behavior (\eg, referring to non-existing objects or locations).
This may be improved by adding further levels of feedback to the LLM, or using strategies like Grounded Decoding~\cite{huang_grounded_2023}.
Finally, our system inherits biases and other flaws from its LLM~\cite{bender_dangers_2021}, which may lead to problematic utterances and behaviors.
In future work, we will try to address some of these challenging questions to further push the boundaries of natural, real-world interactions with humanoid robots.




\section*{Acknowledgments}
\ackJuBotReallabor

\bibliographystyle{plainnat}
{
\footnotesize
\bibliography{root}

\begin{thebibliography}{70}
\providecommand{\natexlab}[1]{#1}
\providecommand{\url}[1]{\texttt{#1}}
\expandafter\ifx\csname urlstyle\endcsname\relax
  \providecommand{\doi}[1]{doi: #1}\else
  \providecommand{\doi}{doi: \begingroup \urlstyle{rm}\Url}\fi

\bibitem[Ahmed et~al.(2023)Ahmed, Azab, and Abdelhamid]{moniz_source-code_2023}
Areeg Ahmed, Shahira Azab, and Yasser Abdelhamid.
\newblock Source-code generation using deep learning: A survey.
\newblock In \emph{Progr. Art. Intel.}, volume 14116, pages 467--482. Springer Nature Switzerland, 2023.
\newblock \doi{10.1007/978-3-031-49011-8_37}.

\bibitem[Ahn et~al.(2022)Ahn, Brohan, Brown, Chebotar, Cortes, David, Finn, Fu, Gopalakrishnan, Hausman, Herzog, Ho, et~al.]{ahn_as_2022}
Michael Ahn, Anthony Brohan, Noah Brown, Yevgen Chebotar, Omar Cortes, Byron David, Chelsea Finn, Chuyuan Fu, Keerthana Gopalakrishnan, Karol Hausman, Alex Herzog, Daniel Ho, et~al.
\newblock Do as i can, not as i say: Grounding language in robotic affordances.
\newblock In \emph{Annu. Conf. Rob. Learn.}, 2022.

\bibitem[Asfour et~al.(2018)Asfour, Kaul, W{\"a}chter, Ottenhaus, Weiner, Rader, Grimm, Zhou, Grotz, Paus, Shingarey, and Haubert]{Asfour2018ARMAR6CollaborativeHumanoid}
Tamim Asfour, Lukas Kaul, Mirko W{\"a}chter, Simon Ottenhaus, Pascal Weiner, Samuel Rader, Raphael Grimm, You Zhou, Markus Grotz, Fabian Paus, Dmitriy Shingarey, and Hans Haubert.
\newblock {{ARMAR-6}}: {{A Collaborative Humanoid Robot}} for {{Industrial Environments}}.
\newblock In \emph{{IEEE}-{RAS} Int. Conf. Humanoid Robots}, pages 447--454, 2018.

\bibitem[Bao et~al.(2016)Bao, Hong, Tang, Cheng, Jia, and Xi]{Bao2016TeachRobotsUnderstanding}
Jiatong Bao, Ze~Hong, Hongru Tang, Yu~Cheng, Yunyi Jia, and Ning Xi.
\newblock Teach robots understanding new object types and attributes through natural language instructions.
\newblock In \emph{{IEEE} Int. Conf. Robot. Automat.}, volume~10, 2016.

\bibitem[Bender et~al.(2021)Bender, Gebru, {McMillan}-Major, and Shmitchell]{bender_dangers_2021}
Emily~M. Bender, Timnit Gebru, Angelina {McMillan}-Major, and Shmargaret Shmitchell.
\newblock On the dangers of stochastic parrots: Can language models be too big?
\newblock In \emph{Conf. Fairness, Accountability, Transparency}, pages 610--623, 2021.
\newblock ISBN 978-1-4503-8309-7.

\bibitem[Brown et~al.(2020)Brown, Mann, Ryder, Subbiah, Kaplan, Dhariwal, Neelakantan, Shyam, Sastry, Askell, Agarwal, Herbert-Voss, et~al.]{brown_language_2020}
Tom Brown, Benjamin Mann, Nick Ryder, Melanie Subbiah, Jared~D Kaplan, Prafulla Dhariwal, Arvind Neelakantan, Pranav Shyam, Girish Sastry, Amanda Askell, Sandhini Agarwal, Ariel Herbert-Voss, et~al.
\newblock Language models are few-shot learners.
\newblock In \emph{Int. Conf. Neural Inf. Process. Syst.}, volume~33, pages 1877--1901, 2020.

\bibitem[Chen et~al.(2021)Chen, Tworek, Jun, Yuan, Pinto, Kaplan, Edwards, Burda, Joseph, Brockman, Ray, Puri, et~al.]{chen_evaluating_2021}
Mark Chen, Jerry Tworek, Heewoo Jun, Qiming Yuan, Henrique Ponde de~Oliveira Pinto, Jared Kaplan, Harri Edwards, Yuri Burda, Nicholas Joseph, Greg Brockman, Alex Ray, Raul Puri, et~al.
\newblock Evaluating large language models trained on code.
\newblock \emph{{arXiv}:2107.03374}, 2021.

\bibitem[Dalvi~Mishra et~al.(2022)Dalvi~Mishra, Tafjord, and Clark]{dalvi_mishra_towards_2022}
Bhavana Dalvi~Mishra, Oyvind Tafjord, and Peter Clark.
\newblock Towards teachable reasoning systems: Using a dynamic memory of user feedback for continual system improvement.
\newblock In \emph{Conf. Emp. Meth. Nat. Lang. Proc.}, pages 9465--9480, 2022.

\bibitem[Dehaerne et~al.(2022)Dehaerne, Dey, Halder, De~Gendt, and Meert]{dehaerne_code_2022}
Enrique Dehaerne, Bappaditya Dey, Sandip Halder, Stefan De~Gendt, and Wannes Meert.
\newblock Code generation using machine learning: A systematic review.
\newblock \emph{{IEEE} Access}, 10:\penalty0 82434--82455, 2022.
\newblock \doi{10.1109/ACCESS.2022.3196347}.

\bibitem[Fasola and Matari{\'c}(2013)]{Fasola2013UsingSemanticFields}
Juan Fasola and Maja~J Matari{\'c}.
\newblock Using semantic fields to model dynamic spatial relations in a robot architecture for natural language instruction of service robots.
\newblock In \emph{IEEE/RSJ Int. Conf. Intel. Rob. Syst.}, pages 143--150, 2013.

\bibitem[Forbes et~al.(2015)Forbes, Rao, Zettlemoyer, and Cakmak]{Forbes2015RobotProgrammingDemonstration}
Maxwell Forbes, Rajesh Rao, Luke Zettlemoyer, and Maya Cakmak.
\newblock Robot {{Programming}} by {{Demonstration}} with {{Situated Spatial Language Understanding}}.
\newblock In \emph{{IEEE} Int. Conf. Robot. Automat.}, pages 2014--2020, 2015.

\bibitem[{Google}(2023)]{google_code-chat_2023}
{Google}.
\newblock Code-chat ({Google} {VertexAI}), 2023.
\newblock URL \url{https://cloud.google.com/vertex-ai/generative-ai/docs/model-reference/code-chat}.

\bibitem[Guadarrama et~al.(2013)Guadarrama, Riano, Golland, G{\"o}hring, Jia, Klein, Abbeel, and Darrell]{Guadarrama2013GroundingSpatialRelations}
Sergio Guadarrama, Lorenzo Riano, Dave Golland, Daniel G{\"o}hring, Yangqing Jia, Dan Klein, Pieter Abbeel, and Trevor Darrell.
\newblock Grounding {{Spatial Relations}} for {{Human-Robot Interaction}}.
\newblock In \emph{IEEE/RSJ Int. Conf. Intel. Rob. Syst.}, pages 1640--1647, 2013.

\bibitem[Haluptzok et~al.(2023)Haluptzok, Bowers, and Kalai]{haluptzok_language_2023}
Patrick Haluptzok, Matthew Bowers, and Adam~Tauman Kalai.
\newblock Language models can teach themselves to program better.
\newblock In \emph{Int. Conf. Learn. Repr.}, 2023.

\bibitem[Huang et~al.(2022{\natexlab{a}})Huang, Abbeel, Pathak, and Mordatch]{huang_language_2022}
Wenlong Huang, Pieter Abbeel, Deepak Pathak, and Igor Mordatch.
\newblock Language models as zero-shot planners: Extracting actionable knowledge for embodied agents.
\newblock In \emph{Int. Conf. Mach. Learn.}, volume 162, pages 9118--9147, 2022{\natexlab{a}}.

\bibitem[Huang et~al.(2022{\natexlab{b}})Huang, Xia, Xiao, Chan, Liang, Florence, Zeng, Tompson, Mordatch, Chebotar, Sermanet, Jackson, et~al.]{huang_inner_2022}
Wenlong Huang, Fei Xia, Ted Xiao, Harris Chan, Jacky Liang, Pete Florence, Andy Zeng, Jonathan Tompson, Igor Mordatch, Yevgen Chebotar, Pierre Sermanet, Tomas Jackson, et~al.
\newblock Inner monologue: Embodied reasoning through planning with language models.
\newblock In \emph{Annu. Conf. Rob. Learn.}, 2022{\natexlab{b}}.

\bibitem[Huang et~al.(2023)Huang, Xia, Shah, Driess, Zeng, Lu, Florence, Mordatch, Levine, Hausman, and Ichter]{huang_grounded_2023}
Wenlong Huang, Fei Xia, Dhruv Shah, Danny Driess, Andy Zeng, Yao Lu, Pete Florence, Igor Mordatch, Sergey Levine, Karol Hausman, and Brian Ichter.
\newblock Grounded decoding: Guiding text generation with grounded models for robot control.
\newblock \emph{{arXiv}:2303.00855}, 2023.

\bibitem[Kartmann and Asfour(2023)]{Kartmann2023InteractiveIncrementalLearning}
Rainer Kartmann and Tamim Asfour.
\newblock Interactive and {{Incremental Learning}} of {{Spatial Object Relations}} from {{Human Demonstrations}}.
\newblock \emph{Frontiers in Robotics and AI}, 10\penalty0 (Rob. Learn. Evol.), 2023.

\bibitem[Kr{\"u}ger et~al.(2011)Kr{\"u}ger, Geib, Piater, Petrick, Steedman, W{\"o}rg{\"o}tter, Ude, Asfour, Kraft, Omr{\v c}en, Agostini, and Dillmann]{Kruger2011ObjectAction}
Norbert Kr{\"u}ger, Christopher Geib, Justus Piater, Ronald Petrick, Mark Steedman, Florentin W{\"o}rg{\"o}tter, Ale{\v s} Ude, Tamim Asfour, Dirk Kraft, Damir Omr{\v c}en, Alejandro Agostini, and R{\"u}diger Dillmann.
\newblock Object-{{Action Complexes}}: {{Grounded Abstractions}} of {{Sensorimotor Processes}}.
\newblock \emph{Rob. Auton. Sys.}, 59:\penalty0 740--757, 2011.

\bibitem[Lahiri et~al.(2023)Lahiri, Fakhoury, Naik, Sakkas, Chakraborty, Musuvathi, Choudhury, von Veh, Inala, Wang, and Gao]{lahiri_interactive_2023}
Shuvendu~K. Lahiri, Sarah Fakhoury, Aaditya Naik, Georgios Sakkas, Saikat Chakraborty, Madanlal Musuvathi, Piali Choudhury, Curtis von Veh, Jeevana~Priya Inala, Chenglong Wang, and Jianfeng Gao.
\newblock Interactive code generation via test-driven user-intent formalization.
\newblock \emph{{arXiv}:2208.05950}, 2023.

\bibitem[Le et~al.(2022)Le, Wang, Gotmare, Savarese, and Hoi]{le_coderl_2022}
Hung Le, Yue Wang, Akhilesh~Deepak Gotmare, Silvio Savarese, and Steven Chu~Hong Hoi.
\newblock {CodeRL}: Mastering code generation through pretrained models and deep reinforcement learning.
\newblock In \emph{Int. Conf. Neural Inf. Process. Syst.}, volume~35, pages 21314--21328, 2022.

\bibitem[Liang et~al.(2023)Liang, Huang, Xia, Xu, Hausman, Ichter, Florence, and Zeng]{liang_code_2023}
Jacky Liang, Wenlong Huang, Fei Xia, Peng Xu, Karol Hausman, Brian Ichter, Peter Florence, and Andy Zeng.
\newblock Code {{As Policies}}: {{Language Model Programs}} for {{Embodied Control}}.
\newblock In \emph{{IEEE} Int. Conf. Robot. Automat.}, pages 9493--9500, 2023.

\bibitem[Liu et~al.(2022)Liu, Shen, Zhang, Dolan, Carin, and Chen]{liu_what_2022}
Jiachang Liu, Dinghan Shen, Yizhe Zhang, Bill Dolan, Lawrence Carin, and Weizhu Chen.
\newblock What makes good in-context examples for {GPT}-3?
\newblock In \emph{Deep Learning Inside Out: Worksh. Knowl. Extr. Integr. Deep Learn. Arch.}, pages 100--114, 2022.
\newblock \doi{10.18653/v1/2022.deelio-1.10}.

\bibitem[Liu et~al.(2024)Liu, Yu, Zhang, Xu, Lei, Lai, Gu, Ding, Men, Yang, Zhang, Deng, Zeng, Du, Zhang, Shen, Zhang, Su, Sun, Huang, Dong, and Tang]{liu_agentbench_2024}
Xiao Liu, Hao Yu, Hanchen Zhang, Yifan Xu, Xuanyu Lei, Hanyu Lai, Yu~Gu, Hangliang Ding, Kaiwen Men, Kejuan Yang, Shudan Zhang, Xiang Deng, Aohan Zeng, Zhengxiao Du, Chenhui Zhang, Sheng Shen, Tianjun Zhang, Yu~Su, Huan Sun, Minlie Huang, Yuxiao Dong, and Jie Tang.
\newblock {AgentBench}: Evaluating {LLMs} as agents.
\newblock In \emph{Int. Conf. Learn. Repr.}, 2024.

\bibitem[Luo et~al.(2023)Luo, Xu, Dai, Pasupat, Kazemi, Baral, Imbrasaite, and Zhao]{luo_dricl_2023}
Man Luo, Xin Xu, Zhuyun Dai, Panupong Pasupat, Mehran Kazemi, Chitta Baral, Vaiva Imbrasaite, and Vincent~Y. Zhao.
\newblock Dr.{ICL}: Demonstration-retrieved in-context learning.
\newblock \emph{{arXiv}:2305.14128}, 2023.

\bibitem[Madaan et~al.(2022{\natexlab{a}})Madaan, Tandon, Clark, and Yang]{madaan_memory-assisted_2022}
Aman Madaan, Niket Tandon, Peter Clark, and Yiming Yang.
\newblock Memory-assisted prompt editing to improve {GPT}-3 after deployment.
\newblock In \emph{Conf. Emp. Meth. Nat. Lang. Proc.}, pages 2833--2861, 2022{\natexlab{a}}.

\bibitem[Madaan et~al.(2022{\natexlab{b}})Madaan, Zhou, Alon, Yang, and Neubig]{madaan_language_2022}
Aman Madaan, Shuyan Zhou, Uri Alon, Yiming Yang, and Graham Neubig.
\newblock Language models of code are few-shot commonsense learners.
\newblock In \emph{Conf. Emp. Meth. Nat. Lang. Proc.}, pages 1384--1403, 2022{\natexlab{b}}.
\newblock \doi{10.18653/v1/2022.emnlp-main.90}.

\bibitem[Mialon et~al.(2023)Mialon, Dessi, Lomeli, Nalmpantis, Pasunuru, Raileanu, Roziere, Schick, Dwivedi-Yu, Celikyilmaz, Grave, {LeCun}, and Scialom]{mialon_augmented_2023}
Grégoire Mialon, Roberto Dessi, Maria Lomeli, Christoforos Nalmpantis, Ramakanth Pasunuru, Roberta Raileanu, Baptiste Roziere, Timo Schick, Jane Dwivedi-Yu, Asli Celikyilmaz, Edouard Grave, Yann {LeCun}, and Thomas Scialom.
\newblock Augmented language models: a survey.
\newblock \emph{Trans. Mach. Learn. Research}, 2023.

\bibitem[Misra et~al.(2016)Misra, Sung, Lee, and Saxena]{Misra2016TellMeDave}
Dipendra~K. Misra, Jaeyong Sung, Kevin Lee, and Ashutosh Saxena.
\newblock Tell me {{Dave}}: {{Context-sensitive}} grounding of natural language to manipulation instructions.
\newblock \emph{Int. J. Rob. Research}, 35\penalty0 (1-3):\penalty0 281--300, 2016.

\bibitem[Mohan and Laird(2014)]{Mohan2014LearningGoalOrientedHierarchical}
Shiwali Mohan and John Laird.
\newblock Learning {{Goal-Oriented Hierarchical Tasks}} from {{Situated Interactive Instruction}}.
\newblock \emph{{AAAI}}, 28\penalty0 (1), 2014.

\bibitem[Nicolescu et~al.(2019)Nicolescu, Arnold, Blankenburg, {Feil-Seifer}, Banisetty, Nicolescu, Palmer, and Monteverde]{Nicolescu2019LearningComplexStructuredTasks}
Monica Nicolescu, Natalie Arnold, Janelle Blankenburg, David {Feil-Seifer}, Santosh~Balajee Banisetty, Mircea Nicolescu, Andrew Palmer, and Thor Monteverde.
\newblock Learning of {{Complex-Structured Tasks}} from {{Verbal Instruction}}.
\newblock In \emph{{IEEE}-{RAS} Int. Conf. Humanoid Robots}, pages 770--777, 2019.

\bibitem[Nijkamp et~al.(2023)Nijkamp, Pang, Hayashi, Tu, Wang, Zhou, Savarese, and Xiong]{nijkamp_codegen_2023}
Erik Nijkamp, Bo~Pang, Hiroaki Hayashi, Lifu Tu, Huan Wang, Yingbo Zhou, Silvio Savarese, and Caiming Xiong.
\newblock {CodeGen}: An open large language model for code with multi-turn program synthesis.
\newblock In \emph{Int. Conf. Learn. Repr.}, 2023.

\bibitem[{OpenAI}(2023)]{openai_chatgpt_2023}
{OpenAI}.
\newblock {ChatGPT}, 2023.
\newblock URL \url{https://openai.com/blog/chatgpt/}.

\bibitem[OpenAI(2023)]{openai_gpt4_2023}
OpenAI.
\newblock {GPT-4 Technical Report}.
\newblock \emph{{arXiv}:2303.08774}, 2023.

\bibitem[Ouyang et~al.(2022)Ouyang, Wu, Jiang, Almeida, Wainwright, Mishkin, Zhang, Agarwal, Slama, Ray, Schulman, Hilton, et~al.]{ouyang_training_2022}
Long Ouyang, Jeff Wu, Xu~Jiang, Diogo Almeida, Carroll~L. Wainwright, Pamela Mishkin, Chong Zhang, Sandhini Agarwal, Katarina Slama, Alex Ray, John Schulman, Jacob Hilton, et~al.
\newblock Training language models to follow instructions with human feedback.
\newblock In \emph{Int. Conf. Neural Inf. Process. Syst.}, volume~35, pages 27730--27744, 2022.

\bibitem[Padmakumar et~al.(2022)Padmakumar, Thomason, Shrivastava, Lange, Narayan-Chen, Gella, Piramuthu, Tur, and Hakkani-Tur]{padmakumar_teach_2022}
Aishwarya Padmakumar, Jesse Thomason, Ayush Shrivastava, Patrick Lange, Anjali Narayan-Chen, Spandana Gella, Robinson Piramuthu, Gokhan Tur, and Dilek Hakkani-Tur.
\newblock Teach: Task-driven embodied agents that chat.
\newblock \emph{{AAAI}}, 36\penalty0 (2):\penalty0 2017--2025, 2022.
\newblock \doi{10.1609/aaai.v36i2.20097}.

\bibitem[Parakh et~al.(2023)Parakh, Fong, Simeonov, Gupta, Chen, and Agrawal]{parakh_lifelong_2023}
Meenal Parakh, Alisha Fong, Anthony Simeonov, Abhishek Gupta, Tao Chen, and Pulkit Agrawal.
\newblock Lifelong robot learning with human assisted language planners.
\newblock In \emph{Work. Learn. Eff. Abstr. Plan., CoRL}, 2023.

\bibitem[Parisi et~al.(2022)Parisi, Zhao, and Fiedel]{parisi_talm_2022}
Aaron Parisi, Yao Zhao, and Noah Fiedel.
\newblock {TALM}: Tool augmented language models.
\newblock \emph{{arXiv}:2205.12255}, 2022.

\bibitem[Pauwels and Kragic(2015)]{SimTrack2015}
Karl Pauwels and Danica Kragic.
\newblock Simtrack: A simulation-based framework for scalable real-time object pose detection and tracking.
\newblock In \emph{IEEE/RSJ Int. Conf. Intel. Rob. Syst.}, 2015.

\bibitem[{Peller-Konrad} et~al.(2023){Peller-Konrad}, Kartmann, Dreher, Meixner, Reister, Grotz, and Asfour]{Peller-Konrad2023MemorySystemRobot}
Fabian {Peller-Konrad}, Rainer Kartmann, Christian R.~G. Dreher, Andre Meixner, Fabian Reister, Markus Grotz, and Tamim Asfour.
\newblock A memory system of a robot cognitive architecture and its implementation in {{ArmarX}}.
\newblock \emph{Rob. Auton. Sys.}, 164:\penalty0 20, 2023.
\newblock ISSN 0921-8890.

\bibitem[Pramanick et~al.(2020)Pramanick, Barua, and Sarkar]{Pramanick2020DeComplexTaskPlanning}
Pradip Pramanick, Hrishav~Bakul Barua, and Chayan Sarkar.
\newblock {{DeComplex}}: {{Task}} planning from complex natural instructions by a collocating robot.
\newblock In \emph{IEEE/RSJ Int. Conf. Intel. Rob. Syst.}, page~8, 2020.

\bibitem[Qin et~al.(2023)Qin, Hu, Lin, Chen, Ding, Cui, Zeng, Huang, Xiao, Han, Fung, Su, et~al.]{qin_tool_2023}
Yujia Qin, Shengding Hu, Yankai Lin, Weize Chen, Ning Ding, Ganqu Cui, Zheni Zeng, Yufei Huang, Chaojun Xiao, Chi Han, Yi~Ren Fung, Yusheng Su, et~al.
\newblock Tool learning with foundation models.
\newblock \emph{{arXiv}:2304.08354}, 2023.

\bibitem[Reimers and Gurevych(2019)]{reimers_sentence-bert_2019}
Nils Reimers and Iryna Gurevych.
\newblock Sentence-{BERT}: Sentence embeddings using siamese {BERT}-networks.
\newblock In \emph{Conf. Emp. Meth. Nat. Lang. Proc.}, pages 3982--3992, 2019.
\newblock \doi{10.18653/v1/D19-1410}.

\bibitem[Ren et~al.(2023)Ren, Dixit, Bodrova, Singh, Tu, Brown, Xu, Takayama, Xia, Xu, Sadigh, Zeng, and Majumdar]{ren_robots_2023}
Allen~Z. Ren, Anushri Dixit, Alexandra Bodrova, Sumeet Singh, Stephen Tu, Noah Brown, Peng Xu, Leila Takayama, Fei Xia, Zhenjia Xu, Dorsa Sadigh, Andy Zeng, and Anirudha Majumdar.
\newblock Robots that ask for help: Uncertainty alignment for large language model planners.
\newblock In \emph{Annu. Conf. Rob. Learn.}, 2023.

\bibitem[Sarch et~al.(2023)Sarch, Wu, Tarr, and Fragkiadaki]{sarch_open-ended_2023}
Gabriel Sarch, Yue Wu, Michael Tarr, and Katerina Fragkiadaki.
\newblock Open-ended instructable embodied agents with memory-augmented large language models.
\newblock In \emph{Conf. Emp. Meth. Nat. Lang. Proc.}, pages 3468--3500, 2023.

\bibitem[Shridhar et~al.(2020)Shridhar, Mittal, and Hsu]{Shridhar2020INGRESSInteractiveVisual}
Mohit Shridhar, Dixant Mittal, and David Hsu.
\newblock {{INGRESS}}: {{Interactive}} visual grounding of referring expressions.
\newblock \emph{Int. J. Rob. Research}, 39\penalty0 (2-3):\penalty0 217--232, 2020.

\bibitem[Singh et~al.(2023)Singh, Blukis, Mousavian, Goyal, Xu, Tremblay, Fox, Thomason, and Garg]{singh_progprompt_2023}
Ishika Singh, Valts Blukis, Arsalan Mousavian, Ankit Goyal, Danfei Xu, Jonathan Tremblay, Dieter Fox, Jesse Thomason, and Animesh Garg.
\newblock {ProgPrompt}: Generating situated robot task plans using large language models.
\newblock In \emph{{IEEE} Int. Conf. Robot. Automat.}, pages 11523--11530, 2023.

\bibitem[Skreta et~al.(2023)Skreta, Yoshikawa, Arellano-Rubach, Ji, Kristensen, Darvish, Aspuru-Guzik, Shkurti, and Garg]{skreta_errors_2023}
Marta Skreta, Naruki Yoshikawa, Sebastian Arellano-Rubach, Zhi Ji, Lasse~Bjørn Kristensen, Kourosh Darvish, Alán Aspuru-Guzik, Florian Shkurti, and Animesh Garg.
\newblock Errors are useful prompts: Instruction guided task programming with verifier-assisted iterative prompting.
\newblock \emph{{arXiv}:2303.14100}, 2023.

\bibitem[Song et~al.(2023)Song, Wu, Washington, Sadler, Chao, and Su]{song_llm-planner_2023}
Chan~Hee Song, Jiaman Wu, Clayton Washington, Brian~M Sadler, Wei-Lun Chao, and Yu~Su.
\newblock {LLM}-planner: Few-shot grounded planning for embodied agents with large language models.
\newblock In \emph{Int. Conf. Comp. Vis.}, pages 2998--3009, 2023.

\bibitem[Tellex et~al.(2011)Tellex, Kollar, Dickerson, Walter, Banerjee, Teller, and Roy]{Tellex2011UnderstandingNaturalLanguage}
Stefanie Tellex, Thomas Kollar, Steven Dickerson, Matthew~R. Walter, Ashis~Gopal Banerjee, Seth Teller, and Nicholas Roy.
\newblock Understanding {{Natural Language Commands}} for {{Robotic Navigation}} and {{Mobile Manipulation}}.
\newblock In \emph{{AAAI}}, volume~25 of \emph{1}, pages 1507--1514, 2011.

\bibitem[Tellex et~al.(2020)Tellex, Gopalan, {Kress-Gazit}, and Matuszek]{Tellex2020RobotsThatUse}
Stefanie Tellex, Nakul Gopalan, Hadas {Kress-Gazit}, and Cynthia Matuszek.
\newblock Robots {{That Use Language}}: {{A Survey}}.
\newblock \emph{Annu. Rev. Control Rob. Auton. Sys.}, 3\penalty0 (1):\penalty0 25--55, 2020.

\bibitem[Touvron et~al.(2023)Touvron, Lavril, Izacard, Martinet, Lachaux, Lacroix, Rozière, Goyal, Hambro, Azhar, Rodriguez, Joulin, et~al.]{touvron_llama_2023}
Hugo Touvron, Thibaut Lavril, Gautier Izacard, Xavier Martinet, Marie-Anne Lachaux, Timothée Lacroix, Baptiste Rozière, Naman Goyal, Eric Hambro, Faisal Azhar, Aurelien Rodriguez, Armand Joulin, et~al.
\newblock {LLaMA}: Open and efficient foundation language models.
\newblock \emph{{arXiv}:2302.13971}, 2023.

\bibitem[Vahrenkamp et~al.(2015)Vahrenkamp, Wächter, Kröhnert, Welke, and Asfour]{vahrenkamp_robot_2015}
Nikolaus Vahrenkamp, Mirko Wächter, Manfred Kröhnert, Kai Welke, and Tamim Asfour.
\newblock The robot software framework {ArmarX}.
\newblock \emph{it - Information Technology}, 57, 2015.

\bibitem[Vemprala et~al.(2023)Vemprala, Bonatti, Bucker, and Kapoor]{vemprala_chatgpt_2023}
Sai Vemprala, Rogerio Bonatti, Arthur Bucker, and Ashish Kapoor.
\newblock {ChatGPT} for robotics: Design principles and model abilities, 2023.
\newblock URL \url{https://www.microsoft.com/en-us/research/publication/chatgpt-for-robotics-design-principles-and-model-abilities/}.

\bibitem[Wake et~al.(2023)Wake, Kanehira, Sasabuchi, Takamatsu, and Ikeuchi]{wake_chatgpt_2023}
Naoki Wake, Atsushi Kanehira, Kazuhiro Sasabuchi, Jun Takamatsu, and Katsushi Ikeuchi.
\newblock {ChatGPT} empowered long-step robot control in various environments: A case application, 2023.
\newblock URL \url{https://www.microsoft.com/en-us/research/publication/chatgpt-empowered-long-step-robot-control-in-various-environments-a-case-application/}.

\bibitem[Walter et~al.(2013)Walter, Hemachandra, Homberg, Tellex, and Teller]{Walter2013LearningSemanticMaps}
Matt Walter, Sachithra Hemachandra, Bianca Homberg, Stefanie Tellex, and Seth Teller.
\newblock Learning semantic maps from natural language descriptions.
\newblock In \emph{Rob.: Science and Systems}, 2013.

\bibitem[Wang et~al.(2024{\natexlab{a}})Wang, Xie, Jiang, Mandlekar, Xiao, Zhu, Fan, and Anandkumar]{wang_voyager_2024}
Guanzhi Wang, Yuqi Xie, Yunfan Jiang, Ajay Mandlekar, Chaowei Xiao, Yuke Zhu, Linxi Fan, and Anima Anandkumar.
\newblock Voyager: An open-ended embodied agent with large language models.
\newblock \emph{Trans. Mach. Learn. Research}, 2024{\natexlab{a}}.
\newblock ISSN 2835-8856.

\bibitem[Wang and Chen(2023)]{wang_review_2023}
Jianxun Wang and Yixiang Chen.
\newblock A review on code generation with {LLMs}: Application and evaluation.
\newblock In \emph{Int. Conf. Med. Art. Intel.}, pages 284--289, 2023.
\newblock \doi{10.1109/MedAI59581.2023.00044}.

\bibitem[Wang et~al.(2024{\natexlab{b}})Wang, Wang, Liu, Chen, Yuan, Peng, and Ji]{wang_mint_2024}
Xingyao Wang, Zihan Wang, Jiateng Liu, Yangyi Chen, Lifan Yuan, Hao Peng, and Heng Ji.
\newblock {MINT}: Evaluating {LLMs} in multi-turn interaction with tools and language feedback.
\newblock In \emph{Int. Conf. Learn. Repr.}, 2024{\natexlab{b}}.

\bibitem[Wang et~al.(2023)Wang, Zhang, Yang, Shi, Zhou, Hao, Xiong, Li, Sim, Chen, Zhu, Yang, et~al.]{wang_interactive_2023}
Zekun Wang, Ge~Zhang, Kexin Yang, Ning Shi, Wangchunshu Zhou, Shaochun Hao, Guangzheng Xiong, Yizhi Li, Mong~Yuan Sim, Xiuying Chen, Qingqing Zhu, Zhenzhu Yang, et~al.
\newblock Interactive natural language processing.
\newblock \emph{{arXiv}:2305.13246}, 2023.

\bibitem[Wei et~al.(2022)Wei, Wang, Schuurmans, Bosma, ichter, Xia, Chi, Le, and Zhou]{wei_chain_2022}
Jason Wei, Xuezhi Wang, Dale Schuurmans, Maarten Bosma, brian ichter, Fei Xia, Ed~H. Chi, Quoc~V. Le, and Denny Zhou.
\newblock Chain-of-thought prompting elicits reasoning in large language models.
\newblock In \emph{Int. Conf. Neural Inf. Process. Syst.}, 2022.

\bibitem[Weigelt et~al.(2020)Weigelt, Steurer, Hey, and Tichy]{Weigelt2020ProgrammingNaturalLanguage}
Sebastian Weigelt, Vanessa Steurer, Tobias Hey, and Walter~F. Tichy.
\newblock Programming in {{Natural Language}} with {{fuSE}}: {{Synthesizing Methods}} from {{Spoken Utterances Using Deep Natural Language Understanding}}.
\newblock In \emph{Annu. Meeting Assoc. Comput. Linguistics}, pages 4280--4295, 2020.

\bibitem[Wu et~al.(2023)Wu, Antonova, Kan, Lepert, Zeng, Song, Bohg, Rusinkiewicz, and Funkhouser]{wu_tidybot_2023}
Jimmy Wu, Rika Antonova, Adam Kan, Marion Lepert, Andy Zeng, Shuran Song, Jeannette Bohg, Szymon Rusinkiewicz, and Thomas Funkhouser.
\newblock {TidyBot}: Personalized robot assistance with large language models.
\newblock \emph{{arXiv}:2305.05658}, 2023.

\bibitem[Yang et~al.(2023)Yang, Prabhakar, Narasimhan, and Yao]{yang_intercode_2023}
John Yang, Akshara Prabhakar, Karthik Narasimhan, and Shunyu Yao.
\newblock {InterCode}: Standardizing and benchmarking interactive coding with execution feedback.
\newblock In \emph{Int. Conf. Neural Inf. Process. Syst.}, volume~36, pages 23826--23854, 2023.

\bibitem[Yang et~al.(2024)Yang, Liu, Wu, Yang, Fung, Li, Huang, Cao, Wang, Ji, and Zhai]{yang_if_2024}
Ke~Yang, Jiateng Liu, John Wu, Chaoqi Yang, Yi~Fung, Sha Li, Zixuan Huang, Xu~Cao, Xingyao Wang, Heng Ji, and {ChengXiang} Zhai.
\newblock If {LLM} is the wizard, then code is the wand: A survey on how code empowers large language models to serve as intelligent agents.
\newblock In \emph{{ICLR} 2024 Workshop on {LLM} {Agents}}, 2024.

\bibitem[Yao et~al.(2023)Yao, Zhao, Yu, Du, Shafran, Narasimhan, and Cao]{yao_react_2023}
Shunyu Yao, Jeffrey Zhao, Dian Yu, Nan Du, Izhak Shafran, Karthik~R. Narasimhan, and Yuan Cao.
\newblock {ReAct}: Synergizing reasoning and acting in language models.
\newblock In \emph{Int. Conf. Learn. Repr.}, 2023.

\bibitem[Ye et~al.(2023)Ye, Wu, Feng, Yu, and Kong]{ye_compositional_2023}
Jiacheng Ye, Zhiyong Wu, Jiangtao Feng, Tao Yu, and Lingpeng Kong.
\newblock Compositional exemplars for in-context learning.
\newblock \emph{{arXiv}:2302.05698}, 2023.

\bibitem[Zeng et~al.(2023)Zeng, Attarian, ichter, Choromanski, Wong, Welker, Tombari, Purohit, Ryoo, Sindhwani, Lee, Vanhoucke, and Florence]{zeng_socratic_2023}
Andy Zeng, Maria Attarian, brian ichter, Krzysztof~Marcin Choromanski, Adrian Wong, Stefan Welker, Federico Tombari, Aveek Purohit, Michael~S. Ryoo, Vikas Sindhwani, Johnny Lee, Vincent Vanhoucke, and Pete Florence.
\newblock Socratic models: Composing zero-shot multimodal reasoning with language.
\newblock In \emph{Int. Conf. Learn. Repr.}, 2023.

\bibitem[Zha et~al.(2023)Zha, Cui, Lin, Kwon, Arenas, Zeng, Xia, and Sadigh]{zha_distilling_2023}
Lihan Zha, Yuchen Cui, Li-Heng Lin, Minae Kwon, Montserrat~Gonzalez Arenas, Andy Zeng, Fei Xia, and Dorsa Sadigh.
\newblock Distilling and retrieving generalizable knowledge for robot manipulation via language corrections.
\newblock In \emph{Work. Lang. Robot Learn., CoRL}, 2023.

\bibitem[Zheng et~al.(2024)Zheng, Ning, Wang, Zhang, Zheng, Ye, and Chen]{zheng_survey_2024}
Zibin Zheng, Kaiwen Ning, Yanlin Wang, Jingwen Zhang, Dewu Zheng, Mingxi Ye, and Jiachi Chen.
\newblock A survey of large language models for code: Evolution, benchmarking, and future trends.
\newblock \emph{{arXiv}:2311.10372}, 2024.

\end{thebibliography}
}

{
\onecolumn
\appendix
\newcommand{\pythoninline}[1]{{\small\mintinline{python}{#1}}}

\subsection{Code as Policies Evaluation -- Setup Details}
\label{app:cap_eval}
All evaluated system generate code that utilizes the same API as in \citet{liang_code_2023}.
Specifically, the LLM can access the following functions:
\begin{itemize}
    \item \pythoninline{denormalize_xy(pos_normalized)}\\
    Takes in a normalized position (coordinates between 0 and 1) and returns the denormalized position that can be used to execute a grasping or placing operation.
    \item \pythoninline{get_obj_names()}\\
    Returns the objects in the scene as a list of strings.
    \item \pythoninline{get_obj_pos(obj_name)}\\
    Returns the (denormalized) position of the object in the scene. If the object does not exist, it throws a \pythoninline{SemanticHintError} hinting at using one of the existing objects. 
    \item \pythoninline{put_first_on_second(obj_name_1, target_name_or_xy_pos)}\\
    Executes pick and place. Both arguments can either be object names or position values.
\end{itemize}
The perception functions access the simulation state, so they deliver perfect results.
The scripted motion plans of \pythoninline{put_first_on_second} taken from \citet{liang_code_2023} are not perfect as they do not take possible collisions into account, thus sometimes dropping blocks.

\noindent Our method can additionally use the two built-in functions
\begin{itemize}
    \item \pythoninline{wait_for_trigger()}\\
    Will wait for additional human input, and also performs automatic success detection to log the state of the experiment.
    \item \pythoninline{learn_from_interaction()}\\
    Triggers incremental learning as explained in \cref{sec:methods:learning}.
\end{itemize}
One run of our experiments goes as follows:
\begin{enumerate}
    \item Randomly generate a scene and pick attributes for the placeholders in the current instruction template (e.g., for the instruction \enquote{pick up the $<$\textit{block}$>$ and place it on the $<$\textit{corner}$>$}, pick $block =$ red and $corner =$ top left corner)
    \item Pass the instruction to the system
    \item Wait for the system to return (i.e. finish executing the generated code for CaP and invoking $\functionWait$ for our method). At this step, the run can fail with an error (i.e. uncatched exception) or timeout.
    \item Run automatic success detection, log the result (either success or failure).
    \item Present the result to the human observer and ask for further instruction.
    \item If there is no further instruction, the run ends in the current state. Otherwise, repeat from 2.
\end{enumerate}
All runs of all instructions are performed in sequence.
This means that the simulation environment and the evaluated system are reset between each run, but the interaction examples memory is kept (in order to allow incremental learning).

\phantomsection
\textbf{HELPER implementation details}
\label{app:cap_eval:helper}
The prompts provided by \citet{sarch_open-ended_2023} are designed for the TEACh~\cite{padmakumar_teach_2022} environment, so we adapt them to the CaP simulated tabletop evaluation scenario \& API explained above.
HELPER originally uses a vision-language model (VLM) to pick the most probable reason from common failure cases given the current camera image.
Since these failure cases are highly specific for the TEACh environment, we replace this mechanism with feeding back execution errors as given by the simulation environment directly to HELPER's \textit{Self-Reflection \& Correction} prompt (i.e. not involving a VLM).
For a fair comparison, we handle the same set of errors as for our method, while the non-caught errors fail the experiment immediately (e.g. SyntaxError).

\citet{sarch_open-ended_2023} also propose to utilize user feedback on completion of the program.
In their experiments, this feedback is generated programmatically by comparing the current and ground-truth goal state of the TEACh environment.
In our experiments, we use the human observer to provide this feedback, which is then passed as a new command to the \textsc{Planner}.
Note that the design of HELPER's prompting method does not allow the system to inspect its previously generated code when responding to errors or feedback, in contrast to our Python console prompting.

HELPER's few-shot memory is expanded with successful trials as indicated by the human supervisor.
Specifically, if the simulation state is \textit{success}, and the human does not provide further feedback, the most recent instruction and generated code are stored in the example memory.
We initialize the few-shot example memory with the same examples used by CaP and our method, translated to the prompting style of HELPER.
For a fair comparison, we also set $k=16$ for top-$k$ example retrieval of similar instructions.
In contrast, for the database of failure handling examples (which is separate from the \enquote{normal} examples in HELPER), we follow HELPER and provide only $k=3$ examples (and there is also no incremental learning).
The code and prompts for evaluating HELPER can be found in the supplementary material.

\vspace{1.5cm}
\subsection{Code as Policies Evaluation -- Interaction Details \& Examples}
\label{app:cap_examples}
For our experiments, human feedback was provided by one individual (the first author) for all runs, respecting the following guidelines:
\begin{itemize}
    \item First provide high-level feedback on how to solve the task correctly, only if that fails multiple times or the initial condition is already destroyed, give the solution via natural language.
    Consider the example task \enquote{pick up the first block from the bottom and put it in the top right corner}. 
    The system might flip the coordinates and pick the wrong block. Appropriate feedback would be \enquote{you did not correctly identify the first block from the bottom. you picked the topmost block instead.}. 
    If such feedback does not help multiple times or the initial condition is already destroyed (in the example, if the task was to put the block in the bottom right corner, another block would already be moved to the bottom, so correctly following the initial task to pick up the bottommost block would not result in picking the correct block), feedback like \enquote{it is the blue block I was asking for} would be valid.
    
    \item The feedback is independent of the code produced by each evaluated method, i.e. the human observes only the actions visible in the simulation rendering.
    Direct feedback on the code (\enquote{you flipped the coordinates in line 2}) is never provided.

    \item If there are simple execution errors (e.g. a block is dropped), the feedback is to state the problem and ask to repeat the action (\enquote{you dropped the red block, try that one again}).

    \item Feedback is always given exhaustively, trying to reach success.
    Therefore, experiment runs either end with success, error, or timeout (this especially happened when the simulated gripper tried to calculate inverse kinematics for unreachable points).
    The failure state (no error, but goal not reached) is only a temporary state, because it still allows for further feedback.
    The only exception is when the human observer identified that the world state makes it impossible to solve the task, which means further feedback would be useless, for instance when the robot dropped a task-relevant block in an unfortunate way so that it ends up outside of its action space.

    \item Incremental learning is triggered if the system first failed to solve a task, and then managed to do it correctly without the human providing the solution explicitly.
    For instance, when a task was solved only after the human revealed which block is the correct one, no request for incremental learning was given.
    Similarly, incremental learning was not requested when the feedback was only necessary due to execution errors such as dropping a block.
\end{itemize}

The remainder of this section lists some example interactions from the CaP evaluation to showcase the different type of feedback provided by human supervision.
The human feedback can be found in the \pythoninline{'text'} property in the return value of the \pythoninline{wait_for_trigger()} calls (except for the first one in each interaction, which returns the initial command).

\textbf{ours, gpt-4, UA/UI}:

This is an example of the LLM performing some behavior in a principally correct way, but without considering the physical side effects (destroying the stack when moving all blocks at once).
The human feedback first instructs how to do it correctly, and then asks to learn from that mistake.

\begin{mdframed}
\codesize
\begin{minted}{python}
>>> wait_for_trigger()
{'type': 'dialog', 'text': 'stack all the blocks on the bottom side'}
>>> bottom_side_pos = denormalize_xy([0.5, 0])
>>> block_names = [name for name in get_obj_names() if 'block' in name]
>>> for i in range(len(block_names) - 1):
...     put_first_on_second(block_names[i + 1], block_names[i])
>>> put_first_on_second(block_names[0], bottom_side_pos)
'success'
>>> wait_for_trigger()
{'type': 'dialog', 'text': 'by stacking them first and then moving, you destroyed the stack. '
                           'create the stack directly at the target location'}
>>> block_names = [name for name in get_obj_names() if 'block' in name]
>>> for i in range(len(block_names)):
...     put_first_on_second(block_names[i], bottom_side_pos)
>>> wait_for_trigger()
{'type': 'dialog', 'text': 'well done, learn that for next time'}
>>> learn_from_interaction()
\end{minted}
\end{mdframed} \filbreak

The following example demonstrates how the LLM can adapt subsymbolic parameters given a user instruction.

\begin{mdframed}
\codesize
\begin{minted}{python}
>>> wait_for_trigger()
{'type': 'dialog', 'text': 'pick up the gray block and place it a lot to the right of the purple bowl'}
>>> get_obj_names()
['gray block', 'cyan block', 'purple bowl', 'pink bowl', 'gray bowl', 'yellow bowl']
>>> purple_bowl_pos = get_obj_pos('purple bowl')
>>> target_pos = purple_bowl_pos + [1.0, 0]  # a lot to the right
>>> put_first_on_second('gray block', target_pos)
\end{minted}
\begin{minted}{text}
Point not in boundaries. Make sure to use denormalize_xy correctly.
\end{minted}
\begin{minted}{python}
>>> target_pos = denormalize_xy([1.0, purple_bowl_pos[1]])
>>> put_first_on_second('gray block', target_pos)
\end{minted}
\begin{minted}{text}
Point not in boundaries. Make sure to use denormalize_xy correctly.
\end{minted}
\begin{minted}{python}
>>> target_pos = purple_bowl_pos + [0.5, 0]  # a bit to the right
>>> put_first_on_second('gray block', target_pos)
'success'
>>> wait_for_trigger()
{'type': 'dialog', 'text': 'thats a bit too far, try half of the distance'}
>>> target_pos = purple_bowl_pos + [0.25, 0]  # half of the previous distance
>>> put_first_on_second('gray block', target_pos)
'success'
>>> wait_for_trigger()
{'type': 'dialog', 'text': 'well done, learn from your mistakes'}
\end{minted}
\end{mdframed} \filbreak

\textbf{CaP, gpt-4, UA/SI}:

In this example, the LLM did generate an empty response on the first request. Given the clarification instruction, it performs the task correctly. 

\begin{mdframed}
\codesize
\begin{minted}{python}
# put all the blocks in the bowls with matching colors.
objects = ['red block', 'orange block', 'yellow block', 'yellow bowl', 'orange bowl', 'red bowl']
# the yellow block in the yellow bowl etc.
objects = ['red block', 'orange block', 'yellow block', 'yellow bowl', 'orange bowl', 'red bowl']
put_first_on_second('yellow block', 'yellow bowl')
put_first_on_second('orange block', 'orange bowl')
put_first_on_second('red block', 'red bowl')
\end{minted}
\end{mdframed} \filbreak

\textbf{ours w/o learn, gpt-4, SA/SI}:

This example shows that the behavior of the LLM occasionally does not use the correct reasoning, but arrives at the correct solution (after multiple human corrections) by chance.
Our experiments solely evaluate the visible outcome.
The human supervisor does not observe the code generated by the LLM.

\begin{mdframed}
\codesize
\begin{minted}{python}
>>> wait_for_trigger()
{'type': 'dialog', 'text': 'pick up the first block from the top and place it on the top side'}
>>> get_obj_names()
['green block', 'orange block', 'red block', 'brown bowl', 'purple bowl']
>>> top_side_pos = denormalize_xy([0.5, 1])
>>> put_first_on_second('green block', top_side_pos)
'success'
>>> wait_for_trigger()
{'type': 'dialog', 'text': 'no, that was the first block from the bottom.'}
>>> put_first_on_second('red block', top_side_pos)
'success'
>>> wait_for_trigger()
{'type': 'dialog', 'text': 'still not correct'}
>>> put_first_on_second('orange block', top_side_pos)
'success'
>>> wait_for_trigger()
\end{minted}
\end{mdframed} \filbreak

\textbf{ours w/o retrieval, gpt-4, SA/SI}

This example demonstrates that the human feedback sometimes does not need to correct a misunderstanding, but solely provides environment feedback about the robot's actions having unintended side effects or not working as expected.
For instance, here the robot did drop the stack of blocks while putting the last one on top.

\begin{mdframed}
\codesize
\begin{minted}{python}
>>> wait_for_trigger()
{'type': 'dialog', 'text': 'put the blocks in the green bowl'}
>>> get_obj_names()
['green block', 'gray block', 'red block', 'orange block', 'green bowl', 'brown bowl', 'purple bowl']
>>> put_first_on_second('green block', 'green bowl')
'success'
>>> put_first_on_second('gray block', 'green bowl')
'success'
>>> put_first_on_second('red block', 'green bowl')
'success'
>>> put_first_on_second('orange block', 'green bowl')
'success'
>>> wait_for_trigger()
{'type': 'dialog', 'text': 'you dropped them, try again'}
>>> put_first_on_second('green block', 'green bowl')
'success'
>>> put_first_on_second('gray block', 'green bowl')
'success'
>>> put_first_on_second('red block', 'green bowl')
'success'
>>> put_first_on_second('orange block', 'green bowl')
'success'
>>> wait_for_trigger()
{'type': 'dialog', 'text': 'the gray one is still missing'}
>>> put_first_on_second('gray block', 'green bowl')
'success'
>>> wait_for_trigger()
\end{minted}
\end{mdframed} \filbreak

\textbf{ours, gpt-3.5, SA/SI}

This example nicely demonstrates that
\textit{(i)} \texttt{gpt-3.5} does much more errors initially, as it misunderstands top right corner here (\texttt{gpt-4} always picked up the correct coordinate system from the examples and never did that mistake),
\textit{(ii)} our method enables learning from mistakes by just mentioning the mistake, without providing further help on how to achieve the goal (\enquote{this is the bottom right corner, not the top right}), and
\textit{(iii)} the LLM can perform corrections of subsymbolic values given vague user instructions (\enquote{without a margin}).

\begin{mdframed}
\codesize
\begin{minted}{python}
>>> wait_for_trigger()
{'type': 'dialog', 'text': 'put all the blocks on the top right corner'}
>>> block_names = [name for name in get_obj_names() if 'block' in name]
>>> block_positions = [get_obj_pos(name) for name in block_names]
>>> top_right_pos = denormalize_xy([0.9, 0.1])
>>> for block_name in block_names:
...     put_first_on_second(block_name, top_right_pos)
>>> wait_for_trigger()
{'type': 'dialog', 'text': 'this is the bottom right corner, not the top right'}
>>> block_names = [name for name in get_obj_names() if 'block' in name]
>>> block_positions = [get_obj_pos(name) for name in block_names]
>>> top_right_pos = denormalize_xy([0.9, 0.9])
>>> for block_name in block_names:
...     put_first_on_second(block_name, top_right_pos)
>>> wait_for_trigger()
{'type': 'dialog', 'text': 'good, but please put it right at the top right corner without margin'}
>>> block_names = [name for name in get_obj_names() if 'block' in name]
>>> block_positions = [get_obj_pos(name) for name in block_names]
>>> top_right_pos = denormalize_xy([1.0, 1.0])
>>> for block_name in block_names:
...     put_first_on_second(block_name, top_right_pos)
>>> wait_for_trigger()
{'type': 'dialog', 'text': 'great, learn from that for next time'}
>>> learn_from_interaction()
\end{minted}
\end{mdframed} \filbreak

\newcommand{\implSingleLineWithErrorPlot}[4]{
    \addplot [#1] table[x=x,y=#3] {figures/plots/#2.table};
    \addplot [name path=upper,draw=none,forget plot] table[x=x,y expr=\thisrow{#3}+\thisrow{err}] {figures/plots/#2.table};
    \addplot [name path=lower,draw=none,forget plot] table[x=x,y expr=\thisrow{#3}-\thisrow{err}] {figures/plots/#2.table};
    \addplot [fill=#1!40, opacity=0.3,forget plot] fill between[of=upper and lower];
}
\newcommand{\getcolorbyidx}[1]{%
\ifcase#1\relax black
\or blue
\or red
\or green
\or orange
\else black%
\fi}
\newcommand{\implPlotMetricPerRound}[4]{
    \begin{tikzpicture}
        \begin{axis}[
        width=0.9\linewidth,
        height=5cm,
        ymin = 0,
        xmin = 1,
        xmax= 10,
        xtick={0,1,2,3,4,5,6,7,8,9,10},
        ytick={0, .2,.4,.6,.8,1},
        every tick/.style={thick},
        axis lines=middle,
        legend style={at={(0.5,1.1)},anchor=south, legend columns=2},
        xlabel=round, xlabel style={at=(current axis.right of origin), anchor=west},
        ylabel=$#2$, ylabel style={at=(current axis.above origin), anchor=south}]

            \foreach \file [count=\idx] in {#4}
                {
                \edef\cvalue{\getcolorbyidx{\idx}}
                \expandafter\implSingleLineWithErrorPlot\expandafter{\cvalue}{\file}{#2}
                };            
            \legend{#3}
        \end{axis}
    \end{tikzpicture}
}
\newcommand{\plotNumInteractionsPerRound}[2]{
    \implPlotMetricPerRound{1.7}{n}{#2}{#1}
}
\newcommand{\plotInitialSuccessPerRound}[2]{
    \implPlotMetricPerRound{1.1}{i}{#2}{#1}
}
\newcommand{\tableFirstColFont}[0]{\scriptsize}

\vspace{2cm}
\subsection{Code as Policies Evaluation -- Detailed Results}
\label{app:cap_results}

\Cref{tab:app:cap_results_gpt35,tab:app:cap_results_gpt4} show detailed, per-instruction results for our experiments with \texttt{gpt-3.5} and \texttt{gpt-4}, respectively.
The following paragraphs elaborate additional experiments:

\textbf{Setting $k=4$.}
To investigate the effect of the hyperparameter $k$ defining the number of few-shot samples to select from $\memory$, we performed another experiment with $k=4$.
Note that \emph{ours} has $k=16$ and \emph{ours w/o learning} has $k=64$.
From the results for \texttt{gpt-3.5}, we can see that a lower $k$ decreases $i$ and increases $n$, as relevant examples might not be retrieved and thus more interaction is required.
For unseen tasks, the lower amount of few-shot samples also decreases the overall success rate $s$, as more runs fail with an error.

In contrast, \texttt{gpt-4} is able to perform very well even for $k=4$, significantly reducing the costs.
While an initial drop in $i$ can be seen for the first runs with seen instructions, incremental learning can level out this effect so that the systems performs similar or even better than our method with $k=16$.

\textbf{Ablating $\languageModelImprove$.}
In order to confirm the effect of $\languageModelImprove$, we perform another ablation experiment where $\functionLearn$ simply stores the current interaction $\currentHistory$ in $\memory$, instead of calling $\languageModelImprove$ to generate $\improvedHistory$.
In theory, this kind of learning could still work, as $\currentHistory$ also includes the user feedback, thus allowing $\languageModelInteract$ to use this feedback in future interactions.
However, our results for \texttt{gpt-3.5} show that this strategy performs worse than with $\languageModelImprove$, while it still outperforms the system without learning.
Especially, we observed that the system often just repeated the erroneous behavior from the saved example, instead of observing the feedback contained therein, thus requiring another user interaction.
Note that for \texttt{gpt-3.5} ours w/o improve, we had to set $k=12$ to avoid frequent errors due to reaching the token limit of the LLM, as the non-improved stored interactions are typically longer than the improved ones.

For \texttt{gpt-4}, there is no such problem with the context length, thus we kept $k=16$ here.
Due to the generally high performance, the positive effect of using $\languageModelImprove$ is not that pronounced.
For unseen instructions, the method even performs on par or slightly better than our system, as \texttt{gpt-4} is able to pick up (comparatively easy) improvement instructions from previous runs. 
However, we could still see that sometimes multiple corrections of the same mistake are necessary, since subsequent interactions first repeated the wrong behavior (as can be seen by lower $i$ on SA SI).

Qualitatively, we investigated this effect further by comparing our system to that without $\languageModelImprove$ on the real-world demonstration scenarios.
These scenarios require significantly more complex behavior changes than the simulated evaluation.
Thus, here we could clearly see that the system without $\languageModelImprove$ cannot perform as well as our full system.
Specifically, using the same instructions as in \cref{app:demo:bimanual}, we were not able to achieve the bimanual transport of both the drink and cup, although we did instruct the system to learn that twice.
Similarly, in the scenario detailed in \cref{app:demo:ladder}, the system never brought the ladder without asking.
Instead, the system repeats the behavior from the learned sample, including the unnecessary interaction.
For the scenario of \cref{app:demo:speed}, the system without $\languageModelImprove$ interestingly picks up the sped-up behavior after a single instruction for improvement, instead of two as for our full system.
However, it did so without respecting the constraint of which route is safe, i.e. it simply uses increased speed everywhere.
When instructed again to take care to only fasten up on the safe route, it does not use the increased speed at all anymore.

To sum up, we conclude that our method of using CoT prompting with $\languageModelImprove$ to produce an improved interaction example is useful especially for weaker models (\texttt{gpt-3.5}) and harder tasks that involve complex behavioral adaptations.
The simpler method of just storing the erroneous interaction including the user feedback works only for \texttt{gpt-4}, and only in case the instructions concern only local changes (such as changing a single parameter).

\textbf{Numerical support for out-of-task generalization.}
During the experiments, we observed that improvements learned from earlier tasks can benefit later tasks, such as when \texttt{gpt-3.5} initially interprets \enquote{top left corner} as $(0.1, 0.9)$ instead of $(0, 1)$, but after correction and learning, corners are used correctly in unseen tasks as well.
To further investigate this effect, we performed another small experiment: initialize our system with the memory content from after \emph{unseen attributes, seen instructions}, and then evaluate on the unseen tasks.
We repeat each unseen task three times without incremental learning, so that each run can still be considered \enquote{unseen}.
This is compared to the system that does not use incremental learning at all, i.e. the memory is initialized with the predefined prompts only.

The results are shown in \cref{tab:app:cap_out_of_task_generalization}.
For \texttt{gpt-4}, the numbers support that our strategy for incremental learning generalizes to unseen tasks, improving both $s$ and $i$.
However, for \texttt{gpt-3.5}, we observe the opposite: The performance of the system initialized with the knowledge from previous execution of the seen tasks is worse than that of the system that has only the predefined interaction examples.
We hypothesize that this is due to \texttt{gpt-3.5} easily overfitting to the prompt contents (that contain examples of the seen tasks, but not the unseen ones).
Nevertheless, when not constrained to this specific evaluation setting, our system with incremental learning performed better on the unseen tasks, as it can learn about them on the fly (as shown by the numbers in \cref{tab:app:cap_results_gpt35}).

\newpage

\begin{table}[h]
    \centering
    \begin{tabular}{l | rr | rr}
                 & \multicolumn{2}{c|}{ours} & \multicolumn{2}{c}{ours w/o learning} \\
                 &  $s$ & $i$  &  $s$ &  $i$ \\\hline
         gpt-3.5 & 61.1 & 33.3 & 83.3 & 38.9 \\ 
         gpt-4   &  100 & 83.3 & 94.4 & 72.2 \\\hline
    \end{tabular}
    \caption{Results of the dedicated experiment to investigate out-of-task generalization}
    \label{tab:app:cap_out_of_task_generalization}
    \vspace{0.2cm}
\end{table}

\begin{figure}[h]
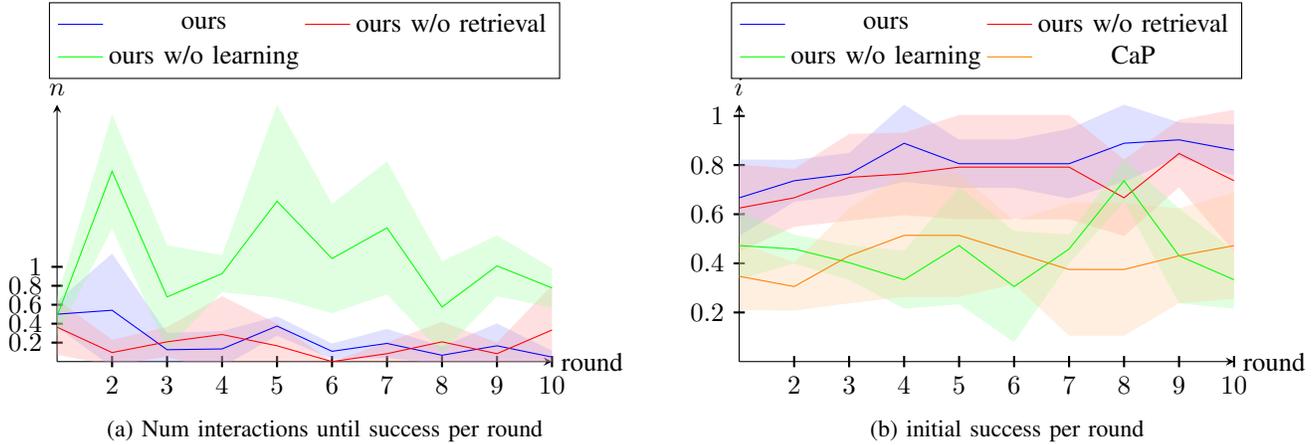

    \centering
    \begin{subfigure}{.5\textwidth}
      \centering
      \plotNumInteractionsPerRound{eval_num_interactions_per_round_gpt3.5_ours_k16,eval_num_interactions_per_round_gpt3.5_ours_k64,eval_num_interactions_per_round_gpt3.5_ours_no_learn}{ours, ours w/o retrieval, ours w/o learning}
      \caption{Num interactions until success per round}
      \label{fig:appendix:num_interactions_per_round_gpt35}
    \end{subfigure}%
    \begin{subfigure}{.5\textwidth}
      \centering
      \plotInitialSuccessPerRound{eval_initial_success_per_round_gpt3.5_ours_k16,eval_initial_success_per_round_gpt3.5_ours_k64,eval_initial_success_per_round_gpt3.5_ours_no_learn,eval_initial_success_per_round_gpt3.5_cap}{ours, ours w/o retrieval, ours w/o learning, CaP}
        \caption{initial success per round}
        \label{fig:appendix:initial_success_per_round_gpt35}
    \end{subfigure}
    \caption{results per round, gpt-3.5 experiments}
    \label{fig:appendix:per_round_gpt35}
    \vspace{0.2cm}
\end{figure}

\begin{figure}[h]
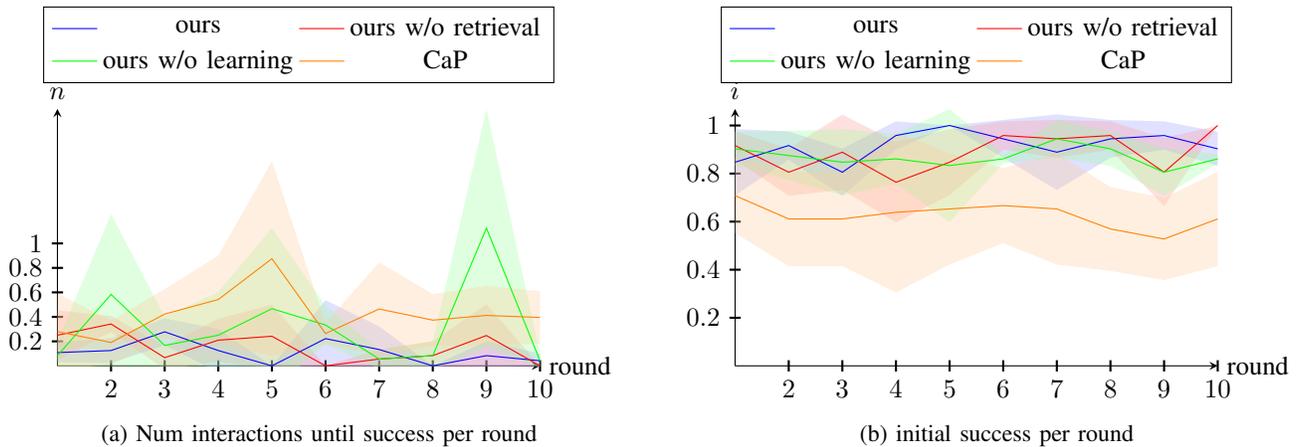

    \centering
    \begin{subfigure}{.49\textwidth}
        \centering
        \plotNumInteractionsPerRound{eval_num_interactions_per_round_gpt4_ours_k16,eval_num_interactions_per_round_gpt4_ours_k64,eval_num_interactions_per_round_gpt4_ours_no_learn,eval_num_interactions_per_round_gpt4_cap}{ours, ours w/o retrieval, ours w/o learning, CaP}
        \caption{Num interactions until success per round}
        \label{fig:appendix:num_interactions_per_round_gpt4}
    \end{subfigure}
    \begin{subfigure}{.49\textwidth}
        \centering
        \plotInitialSuccessPerRound{eval_initial_success_per_round_gpt4_ours_k16,eval_initial_success_per_round_gpt4_ours_k64,eval_initial_success_per_round_gpt4_ours_no_learn,eval_initial_success_per_round_gpt4_cap}{ours, ours w/o retrieval, ours w/o learning , CaP}
        \caption{initial success per round}
        \label{fig:appendix:initial_success_per_round_gpt4}
    \end{subfigure}
    \caption{results per round, gpt-4 experiments}
    \label{fig:appendix:per_round_gpt4}
    \vspace{0.2cm}
\end{figure}

\textbf{Performance analysis \textit{per round}.}
\Cref{fig:appendix:per_round_gpt35,fig:appendix:per_round_gpt4} show initial success rate $i$ and number of interactions until success $n$ \textit{per round}, which means that we aggregate the corresponding numbers over the $x$-th run of each instruction.
For instance, the number for $x = 1$ in \cref{fig:appendix:initial_success_per_round_gpt35} is obtained by averaging the value of $i$ from the \texttt{gpt-3.5} experiments in each first run over all instructions for each of the three blocks of \cref{tab:app:cap_results_gpt35} (SA/SI, UA/SI, UA/UI).
Then, we take average and standard deviation over the three blocks to plot the curve and area in \cref{fig:appendix:initial_success_per_round_gpt35}.

The intention of these graphs is to show how incremental learning improves initial success rate and decreases the need for corrective interactions over time.
When presented with a new instruction, it can be expected that some clarification is needed.
However, after a few runs, the task should be clear, and the system should perform well.
\Cref{fig:appendix:per_round_gpt35} clearly confirms this for \texttt{gpt-3.5}.
While there is some noise (e.g. due to the robot accidentally dropping some block and needing instruction to try again even in a later run), one can see a clear trend of $n$ decreasing to $0$ and $i$ approaching $1$ for our systems with incremental learning.
In contrast, the systems without incremental learning have no such trend and show random variations only.
For \texttt{gpt-4}, since the general performance is much better, the effect is not that emphasized in \cref{fig:appendix:per_round_gpt4}.

\textbf{Cost comparison}
When looking at \cref{tab:app:cap_results_gpt35,tab:app:cap_results_gpt4}, the costs of our method is $5$ to $13$ times higher than CaP or HELPER.
The main cost driving factor is the scheme of repeated individual requests caused by the interactive Python console prompting.
Specifically, the LLM typically generates solely one statement, this is executed, and this repeats.
Each generation, however, involves a full prompt including API definition, few-shot examples and the current interaction.
Thus, when using the OpenAI API, the cost of our method scales linearly with the number of console statements the LLM produces.

However, this cost could be significantly reduced if we had direct access to the LLM.
This is due to the fact that the selection of few-shot examples only changes when there is new user input.
Thus, instead of treating each request independently, the LLM could be used in an \enquote{incremental} mode by generating one statement, keeping the transformer hidden states in memory, force-decoding (i.e. programatically inserting) the return value of the execution of the generated statement, and then continuing to decode the next statement.
That way, the computational cost of our method would be comparable to HELPER and CaP, which generate the full Python script at once.

Nonetheless, in comparison to HELPER and CaP, our prompting scheme has some token overhead.
This is due to emulating a Python console, especially including the return value of each executed function.
For instance, the length of the first prompt in the first run of our tabletop evaluation (before any learning happened, so the information in these prompts is equivalent) is 2682 tokens for our method, 1992 for HELPER, and 1570 for CaP.
Furthermore, we have the special calls to $\functionLearn$, which incur additional computation budget.
Thus, when using an \enquote{incremental} LLM mode as described above, the token cost of our method would be $2$ to $3$ times higher than CaP.

\textbf{Dynamic CaP}
For a fair comparison of our method with CaP, we add a simple form of learning and top-$k$ retrieval and call this method \emph{Dynamic CaP}.
Specifically, similar to HELPER and our method, we define a memory of few-shot samples, initialized with the ones from the original static prompt.
Further, also following HELPER, we store the code transcript after successful episodes as a new sample in this memory.
On every request, we fill the prompt with the top-$k$ retrieved examples from memory, according to semantic similarity of the query.
Note that this few-shot retrieval and storing mechanism also applies to the nested language model programs (LMP) used by CaP, \inlinecode{parse\_obj\_name} and \inlinecode{parse\_position}.
We set $k$ to the same value as used in the original static prompts, i.e. $k=16$ for the main LMP, $k=8$ for \inlinecode{parse\_obj\_name} and $k=9$ for \inlinecode{parse\_position}.

The results in \cref{tab:app:cap_results_gpt35,tab:app:cap_results_gpt4} indicate that these modifications do improve performance over plain CaP, but cannot compete with HELPER or our method.
Specifically, we can observe that learning from successful trials helps with seen instructions by reinforcing correct behavior, but does not transfer to unseen instructions.
Note that this observation also applies to HELPER, but mainly to $i$ since HELPER can better respond to execution errors and user feedback than CaP.

\begin{landscape}
\begin{table}
    \centering
    \setlength\tabcolsep{1px}
    \begin{tabular}{l|rrr|rrr|rrr|rrr|rrr|rrr|rrr|rrr}
                                                    & \multicolumn{3}{c|}{} & \multicolumn{3}{c|}{} & \multicolumn{3}{c|}{ours w/o} & \multicolumn{3}{c|}{ours w/o } & \multicolumn{3}{c|}{ours w/o } & \multicolumn{3}{c|}{}  & \multicolumn{3}{c|}{} & \multicolumn{3}{c}{} \\
                                                    & \multicolumn{3}{c|}{ours} & \multicolumn{3}{c|}{ours $k=4$} & \multicolumn{3}{c|}{retrieval} & \multicolumn{3}{c|}{improve} & \multicolumn{3}{c|}{learning} & \multicolumn{3}{c|}{HELPER}  & \multicolumn{3}{c|}{Dynamic CaP} & \multicolumn{3}{c}{CaP} \\\hline
                                                                                                                         & $s$ & $i$ & $n$ & $s$ & $i$ & $n$ & $s$ & $i$ & $n$ & $s$ & $i$ & $n$ & $s$ & $i$ & $n$ & $s$ & $i$ & $n$ & $s$ & $i$ & $n$ & $s$ & $i$ & $n$ \\\hline
        \textbf{Seen Attributes, Seen Instructions} & & & & & & & & & & & & & & & & & & & & & & & \\ 
        {\tableFirstColFont pick up the $<$block:1$>$ and place it on the $<$block:2$>$}                                      &  100 &  100 & 0.0  &  100 &   90 & 0.2  &  100 &   90 & 0.2  &   90 &   70 & 0.56 &  100 &  100 & 0.0  & 100  & 100  & 0.0  & 100  & 100  & 0.0  &   50 &   50 & 0.0   \\
        {\tableFirstColFont pick up the $<$block$>$ and place it on the $<$bowl$>$}                                           &  100 &  100 & 0.0  &  100 &  100 & 0.0  &  100 &  100 & 0.0  &  100 &  100 & 0.0  &  100 &   90 & 0.1  & 100  & 100  & 0.0  & 100  & 100  & 0.0  &   50 &   50 & 0.0   \\
        {\tableFirstColFont put all the blocks on the $<$corner/side$>$}                                                      &  100 &   90 & 0.2  &  100 &   90 & 0.2  &  100 &   80 & 0.3  &  100 &   90 & 0.1  &  100 &   50 & 1.3  & 100  & 100  & 0.0  &  40  &  40  & 0.0  &   90 &   80 & 0.11   \\
        {\tableFirstColFont put the blocks in the $<$bowl$>$}                                                                 &  100 &   90 & 0.1  &   90 &   60 & 2.33 &   90 &   80 & 0.22 &   90 &   90 & 0.0  &   70 &   60 & 2.29 & 100  &  80  & 0.8  &  80  &  70  & 0.88 &   70 &   70 & 0.0   \\
        {\tableFirstColFont put all the blocks in the bowls with matching colors}                                             &  100 &   90 & 0.1  &  100 &   40 & 1.4  &   70 &   70 & 0.0  &  100 &  100 & 0.0  &   20 &   20 & 0.0  & 100  & 100  & 0.0  &  90  &  80  & 0.11 &   90 &   90 & 0.0   \\
        {\tableFirstColFont pick up the block to the $<$direction$>$ of the $<$bowl$>$ and place it on the $<$corner/side$>$} &   80 &   50 & 0.75 &   90 &   30 & 1.33 &   90 &   70 & 0.44 &   70 &   30 & 1.43 &  100 &    0 & 2.8  &  50  &  10  & 1.0  &   0  &   0  & --   &   10 &   10 & 0.0   \\
        {\tableFirstColFont pick up the block $<$distance$>$ to the $<$bowl$>$ and place it on the $<$corner/side$>$}         &  100 &  100 & 0.0  &  100 &  100 & 0.0  &  100 &   90 & 0.2  &  100 &  100 & 0.0  &   90 &   10 & 1.0  & 100  &  30  & 0.7  &  50  &  50  & 0.0  &   70 &   70 & 0.0   \\
        {\tableFirstColFont pick up the $<$nth$>$ block from the $<$direction$>$ and place it on the $<$corner/side$>$}       &   80 &   80 & 0.0  &  100 &   80 & 0.3  &  100 &   80 & 0.6  &   80 &   60 & 0.38 &  100 &   20 & 1.2  & 100  & 100  & 0.0  &   0  &   0  & --   &    0 &    0 & --   \\\hline
        Total                                                                                                                 & 95.0 & 87.5 & 0.14 & 97.5 & 73.8 & 0.72 & 93.8 & 82.5 & 0.25 & 91.2 & 80.0 & 0.31 & 85.0 & 43.8 & 1.09 & 93.8 & 77.5 & 0.31 & 57.5 & 55.0 & 0.16 &53.8 & 52.5 & 0.02 \\
        Cost                                  & \multicolumn{3}{c|}{1.7\;\$} & \multicolumn{3}{c|}{3.14\;\$}  & \multicolumn{3}{c|}{2.7\;\$} & \multicolumn{3}{c|}{1.2\;\$}  & \multicolumn{3}{c|}{1.8\;\$}  & \multicolumn{3}{c|}{0.3\;\$} & \multicolumn{3}{c|}{0.3\;\$} & \multicolumn{3}{c}{0.3\;\$} \\\hline\hline
        
        \textbf{Unseen Attributes, Seen Instructions} & & & & & & & & & & & & & & & & & & & & & & & \\
        {\tableFirstColFont pick up the $<$block:1$>$ and place it on the $<$block:2$>$}                                      &  100 &   90 & 0.2  &  100 &  100 & 0.0  &  100 &  100 & 0.0  &   80 &   50 & 0.5  &  100 &   90 & 0.1  &  100  & 100  & 0.0  &100  & 100  & 0.0  &  40 &   40 & 0.0  \\
        {\tableFirstColFont pick up the $<$block$>$ and place it on the $<$bowl$>$}                                           &  100 &  100 & 0.0  &  100 &  100 & 0.0  &  100 &  100 & 0.0  &  100 &  100 & 0.0  &  100 &  100 & 0.0  &  100  & 100  & 0.0  &100  & 100  & 0.0  &  90 &   90 & 0.0  \\
        {\tableFirstColFont put all the blocks on the $<$corner/side$>$}                                                      &   90 &   90 & 0.0  &  100 &  100 & 0.0  &  100 &  100 & 0.0  &   80 &   90 & 0.0  &   80 &   60 & 0.25 &  100  & 100  & 0.0  & 40  &  30  & 0.25 &  90 &   90 & 0.0   \\
        {\tableFirstColFont put the blocks in the $<$bowl$>$}                                                                 &  100 &   80 & 1.1  &   90 &   80 & 1.44 &  100 &   80 & 0.3  &   90 &   50 & 0.78 &   70 &   30 & 1.71 &   90  &  80  & 0.22 & 90  &  80  & 0.56 &  90 &   80 & 0.33  \\
        {\tableFirstColFont put all the blocks in the bowls with matching colors}                                             &   90 &   80 & 0.22 &  100 &  100 & 0.0  &   90 &   80 & 0.11 &  100 &  100 & 0.0  &   30 &   30 & 0.0  &  100  & 100  & 0.0  & 90  &  90  & 0.0  &  80 &   80 & 0.0  \\
        {\tableFirstColFont pick up the block to the $<$direction$>$ of the $<$bowl$>$ and place it on the $<$corner/side$>$} &  100 &   80 & 0.2  &  100 &   60 & 1.1  &   90 &   70 & 0.56 &   10 &   10 & 0.0  &   70 &    0 & 2.29 &   50  &  10  & 1.6  &  0  &   0  & --   &   0 &    0 & --  \\
        {\tableFirstColFont pick up the block $<$distance$>$ to the $<$bowl$>$ and place it on the $<$corner/side$>$}         &  100 &   90 & 0.3  &  100 &   90 & 0.3  &   90 &   90 & 0.0  &   80 &   80 & 0.0  &  100 &   20 & 2.0  &   60  &  10  & 0.83 &  0  &   0  & --   &  90 &   90 & 0.0  \\
        {\tableFirstColFont pick up the $<$nth$>$ block from the $<$direction$>$ and place it on the $<$corner/side$>$}       &  100 &   80 & 0.6  &   80 &   80 & 0.0  &  100 &   90 & 0.2  &   30 &   30 & 0.0  &   90 &   30 & 1.22 &  100  &  70  & 0.4  &100  &  60  & 0.6  &   0 &    0 & -- \\\hline
        Total                                                                                                                 & 97.5 & 86.2 & 0.33 & 96.2 & 88.8 & 0.36 & 96.3 & 88.8 & 0.15 & 71.2 & 63.7 & 0.16 & 80.0 & 45.0 & 0.95 &  87.5 & 71.2 & 0.38 &65.0 & 57.5 & 0.23 &60.0 & 58.8 & 0.06 \\
        Cost                                  & \multicolumn{3}{c|}{1.9\;\$} & \multicolumn{3}{c|}{1.2\;\$} & \multicolumn{3}{c|}{2.9\;\$} & \multicolumn{3}{c|}{1.2\;\$} & \multicolumn{3}{c|}{2.4\;\$}  & \multicolumn{3}{c|}{0.3\;\$} & \multicolumn{3}{c|}{0.2\;\$} & \multicolumn{3}{c}{0.3\;\$} \\\hline\hline

        \textbf{Unseen Attributes, Unseen Instructions} & & & & & & & & & & & & & & & & & & & & & & & \\
        {\tableFirstColFont put all the blocks in different corners}                                                          &  100 &   90 & 0.1  &  100 &   90 & 0.2  &  100 &   90 & 0.2  &  100 &  100 & 0.0  &  100 &  100 & 0.0  &  90  &  90  & 0.0   &  70  &  70  & 0.0  &    0 &    0 & --  \\
        {\tableFirstColFont put the blocks in the bowls with mismatched colors}                                               &  100 &   90 & 0.2  &   90 &   60 & 0.33 &   70 &   70 & 0.0  &   60 &   40 & 1.17 &   60 &   30 & 2.5  & 100  &   0  & 1.3   &  50  &  20  & 1.2  &   70 &   60 & 0.14  \\
        {\tableFirstColFont stack all the blocks on the $<$corner/side$>$}                                                    &  100 &  100 & 0.0  &  100 &  100 & 0.0  &  100 &  100 & 0.0  &   90 &   90 & 0.0  &   10 &    0 & 2.0  & 100  & 100  & 0.0   &  50  &  50  & 0.0  &    0 &    0 & --   \\
        {\tableFirstColFont pick up the $<$block$>$ and place it $<$magnitude$>$ to the $<$direction$>$ of the $<$bowl$>$}    &   20 &   10 & 0.5  &   50 &   30 & 0.4  &   70 &   50 & 0.57 &   50 &   50 & 0.0  &   70 &   60 & 1.0  &  10  &   0  & 2.0   &  90  &  70  & 0.22 &   30 &   30 & 0.0  \\
        {\tableFirstColFont pick up the $<$block$>$ and place it in the corner $<$distance$>$ to the $<$bowl$>$}              &  100 &   90 & 0.2  &  100 &   80 & 0.2  &    0 &    0 & --   &  100 &   90 & 0.2  &   80 &   10 & 1.62 &  90  &  20  & 0.78  &  20  &  10  & 1.0  &    0 &    0 & --  \\
        {\tableFirstColFont put all the blocks in a $<$line$>$ line}                                                          &   90 &   40 & 0.67 &   20 &   20 & 0.0  &    0 &    0 & --   &    0 &    0 & --   &   80 &   60 & 0.62 &  90  &  90  & 0.0   &   0  &   0  & --   &    0 &    0 & --  \\\hline
        Total                                                                                                                 & 85.0 & 70.0 & 0.28 & 76.7 & 63.3 & 0.19 & 56.7 & 51.7 & 0.19 & 66.7 & 61.7 & 0.27 & 66.7 & 43.3 & 1.29 & 80.0 & 50.0 & 0.68  & 46.7 & 36.7 & 0.48 & 16.7 & 15.0 & 0.07 \\
        Cost                                 & \multicolumn{3}{c|}{2.5\;\$} & \multicolumn{3}{c|}{1.4\;\$} & \multicolumn{3}{c|}{1.2\;\$} & \multicolumn{3}{c|}{1.2\;\$} & \multicolumn{3}{c|}{1.9\;\$}  & \multicolumn{3}{c|}{0.2\;\$} & \multicolumn{3}{c|}{0.2\;\$} & \multicolumn{3}{c}{0.2\;\$} \\\hline

    \end{tabular}
    \caption{gpt-3.5 experiments (gpt-4 for $\languageModelImprove$)}
    \label{tab:app:cap_results_gpt35}
\end{table}
\end{landscape}

\begin{landscape}
\begin{table}
    \centering
    \setlength\tabcolsep{1px}
    \begin{tabular}{l|rrr|rrr|rrr|rrr|rrr|rrr|rrr|rrr}
                                                    & \multicolumn{3}{c|}{} & \multicolumn{3}{c|}{} & \multicolumn{3}{c|}{ours w/o} & \multicolumn{3}{c|}{ours w/o } & \multicolumn{3}{c|}{ours w/o } & \multicolumn{3}{c|}{}  & \multicolumn{3}{c|}{} & \multicolumn{3}{c}{} \\
                                                    & \multicolumn{3}{c|}{ours} & \multicolumn{3}{c|}{ours $k=4$} & \multicolumn{3}{c|}{retrieval} & \multicolumn{3}{c|}{improve} & \multicolumn{3}{c|}{learning} & \multicolumn{3}{c|}{HELPER}  & \multicolumn{3}{c|}{Dynamic CaP} & \multicolumn{3}{c}{CaP} \\\hline
                                                                                                                         & $s$ & $i$ & $n$ & $s$ & $i$ & $n$ & $s$ & $i$ & $n$ & $s$ & $i$ & $n$ & $s$ & $i$ & $n$ & $s$ & $i$ & $n$ & $s$ & $i$ & $n$ & $s$ & $i$ & $n$ \\\hline
        \textbf{Seen Attributes, Seen Instructions} & & & & & & & & & & & & & & & & & & & & & & & \\ 
        {\tableFirstColFont pick up the $<$block:1$>$ and place it on the $<$block:2$>$}                                      & 100 &  100 & 0.0  &  100 &  100 & 0.0  &  100 &   100 & 0.0  & 100 &  100 & 0.0  &  100 &  100 & 0.0  &  100  & 100  & 0.0  & 100  & 100  & 0.0   &  100 &  100 & 0.0 \\
        {\tableFirstColFont pick up the $<$block$>$ and place it on the $<$bowl$>$}                                           & 100 &  100 & 0.0  &  100 &  100 & 0.0  &  100 &   100 & 0.0  & 100 &  100 & 0.0  &  100 &  100 & 0.0  &  100  & 100  & 0.0  & 100  & 100  & 0.0   &  100 &  100 & 0.0 \\
        {\tableFirstColFont put all the blocks on the $<$corner/side$>$}                                                      & 100 &  100 & 0.0  &   90 &   70 & 0.44 &  100 &    90 & 0.1  & 100 &  100 & 0.0  &  100 &  100 & 0.0  &   90  &  80  & 0.22 &  50  &  50  & 0.0   &   90 &   90 & 0.0 \\
        {\tableFirstColFont put the blocks in the $<$bowl$>$}                                                                 & 100 &   90 & 0.2  &  100 &   90 & 0.1  &  100 &    80 & 0.3  & 100 &   80 & 0.6  &  100 &   80 & 1.6  &  100  &  90  & 0.1  & 100  &  90  & 0.3   &   90 &   90 & 0.0 \\
        {\tableFirstColFont put all the blocks in the bowls with matching colors}                                             & 100 &  100 & 0.0  &  100 &  100 & 0.0  &  100 &   100 & 0.0  & 100 &  100 & 0.0  &  100 &  100 & 0.0  &  100  & 100  & 0.0  & 100  & 100  & 0.0   &  100 &    0 & 1.4 \\
        {\tableFirstColFont pick up the block to the $<$direction$>$ of the $<$bowl$>$ and place it on the $<$corner/side$>$} & 100 &   90 & 0.1  &  100 &   50 & 0.6  &   90 &    60 & 0.56 & 100 &   60 & 0.5  &   90 &   60 & 0.67 &   90  &  60  & 0.67 &  60  &  50  & 0.17  &   30 &   20 & 0.67 \\
        {\tableFirstColFont pick up the block $<$distance$>$ to the $<$bowl$>$ and place it on the $<$corner/side$>$}         & 100 &  100 & 0.0  &  100 &  100 & 0.0  &  100 &   100 & 0.0  & 100 &  100 & 0.0  &  100 &  100 & 0.0  &  100  & 100  & 0.0  & 100  & 100  & 0.0   &   80 &   80 & 0.0 \\
        {\tableFirstColFont pick up the $<$nth$>$ block from the $<$direction$>$ and place it on the $<$corner/side$>$}       & 100 &  100 & 0.0  &  100 &  100 & 0.0  &   90 &    90 & 0.0  & 100 &  100 & 0.0  &  100 &   80 & 0.7  &  100  &  70  & 0.7  & 100  & 100  & 0.0   &   90 &   90 & 0.0 \\\hline
        Total                                                                                                                 & 100 & 97.5 & 0.04 & 98.8 & 88.8 & 0.14 & 97.5 &  90.0 & 0.12 & 100 & 92.5 & 0.14 & 98.8 & 90.0 & 0.37 &  97.5 & 87.5 & 0.21 & 88.8 & 86.2 & 0.06  & 85.0 & 71.2 & 0.26 \\
        Cost                          & \multicolumn{3}{c|}{21.9\;\$} & \multicolumn{3}{c|}{12\;\$} & \multicolumn{3}{c|}{33\;\$} & \multicolumn{3}{c|}{29.7\;\$} & \multicolumn{3}{c|}{23.5\;\$} & \multicolumn{3}{c|}{4.5\;\$} & \multicolumn{3}{c|}{4.8\;\$} & \multicolumn{3}{c}{6\;\$} \\\hline\hline
        
        \textbf{Unseen Attributes, Seen Instructions} & & & & & & & & & & & & & & & & & & & & & & & \\
        {\tableFirstColFont pick up the $<$block:1$>$ and place it on the $<$block:2$>$}                                      & 100 &  100 & 0.0  & 100 &  100 & 0.0  &  100 &  100 & 0.0   &  100 &  100 & 0.0  &  100 &  100 & 0.0  &  100 & 100  & 0.0 & 100  & 100  & 0.0  &   100 &   100 & 0.0 \\
        {\tableFirstColFont pick up the $<$block$>$ and place it on the $<$bowl$>$}                                           & 100 &  100 & 0.0  & 100 &  100 & 0.0  &  100 &  100 & 0.0   &  100 &  100 & 0.0  &  100 &  100 & 0.0  &  100 & 100  & 0.0 & 100  & 100  & 0.0  &   100 &   100 & 0.0 \\
        {\tableFirstColFont put all the blocks on the $<$corner/side$>$}                                                      & 100 &  100 & 0.0  & 100 &  100 & 0.0  &  100 &  100 & 0.0   &  100 &  100 & 0.0  &  100 &  100 & 0.0  &  100 & 100  & 0.0 & 100  & 100  & 0.0  &   100 &   100 & 0.0 \\
        {\tableFirstColFont put the blocks in the $<$bowl$>$}                                                                 & 100 &   90 & 0.1  & 100 &   90 & 0.2  &  100 &   80 & 0.7   &   90 &   80 & 0.33 &  100 &   90 & 0.1  &  100 &  90  & 0.1 & 100  &  90  & 0.1  &    90 &    90 & 0.0 \\
        {\tableFirstColFont put all the blocks in the bowls with matching colors}                                             & 100 &  100 & 0.0  & 100 &  100 & 0.0  &  100 &  100 & 0.0   &  100 &  100 & 0.0  &  100 &  100 & 0.0  &  100 & 100  & 0.0 & 100  & 100  & 0.0  &   100 &     0 & 2.2 \\
        {\tableFirstColFont pick up the block to the $<$direction$>$ of the $<$bowl$>$ and place it on the $<$corner/side$>$} & 100 &   60 & 0.9  & 100 &   90 & 0.1  &   90 &   80 & 0.22  &  100 &   80 & 0.2  &   90 &   50 & 0.67 &  100 &  60  & 0.7 &  80  &  60  & 0.5  &    80 &    60 & 0.62  \\
        {\tableFirstColFont pick up the block $<$distance$>$ to the $<$bowl$>$ and place it on the $<$corner/side$>$}         & 100 &   90 & 0.1  & 100 &  100 & 0.0  &  100 &  100 & 0.0   &  100 &  100 & 0.0  &  100 &  100 & 0.0  &  100 & 100  & 0.0 & 100  & 100  & 0.0  &   100 &   100 & 0.0 \\
        {\tableFirstColFont pick up the $<$nth$>$ block from the $<$direction$>$ and place it on the $<$corner/side$>$}       & 100 &  100 & 0.0  & 100 &   90 & 0.4  &  100 &  100 & 0.0   &  100 &  100 & 0.0  &  100 &  100 & 0.0  &  100 & 100  & 0.0 & 100  & 100  & 0.0  &   100 &   100 & 0.0 \\\hline
        Total                                                                                                                 & 100 & 92.5 & 0.14 & 100 & 96.3 & 0.09 & 98.8 & 95.0 & 0.12  & 98.8 & 95.0 & 0.07 & 98.8 & 92.5 & 0.1  &  100 & 93.8 & 0.1 & 97.5 & 93.8 & 0.07 &  96.2 &  81.2 & 0.35 \\ 
        Cost                       & \multicolumn{3}{c|}{45\;\$} & \multicolumn{3}{c|}{12\;\$} & \multicolumn{3}{c|}{23.7\;\$} & \multicolumn{3}{c|}{31\;\$} & \multicolumn{3}{c|}{19.4\;\$} & \multicolumn{3}{c|}{4.6\;\$} & \multicolumn{3}{c|}{6.1\;\$} & \multicolumn{3}{c}{6.3\;\$} \\\hline\hline

        \textbf{Unseen Attributes, Unseen Instructions} & & & & & & & & & & & & & & & & & & & & & & & \\
        {\tableFirstColFont put all the blocks in different corners}                                                          &  100 &  100 & 0.0  &  100 &  100 & 0.0  &   100 &   100 & 0.0  &  100 &  100 & 0.0  &  100 &  100 & 0.0  & 100  & 100  & 0.0  & 100  & 100  & 0.0  &    0 &     0 & -- \\
        {\tableFirstColFont put the blocks in the bowls with mismatched colors}                                               &  100 &   80 & 0.3  &  100 &   80 & 0.8  &   100 &    70 & 0.6  &  100 &   90 & 0.2  &  100 &   70 & 0.3  & 100  &  80  & 0.4  &  80  &  60  & 0.75 &   80 &     0 & 2.25 \\
        {\tableFirstColFont stack all the blocks on the $<$corner/side$>$}                                                    &  100 &   90 & 0.1  &  100 &  100 & 0.0  &   100 &   100 & 0.0  &  100 &   90 & 0.1  &  100 &  100 & 0.0  & 100  & 100  & 0.0  &  80  &  40  & 1.5  &  100 &   100 & 0.0 \\
        {\tableFirstColFont pick up the $<$block$>$ and place it $<$magnitude$>$ to the $<$direction$>$ of the $<$bowl$>$}    &   60 &   50 & 0.17 &   80 &   70 & 0.12 &    50 &    50 & 0.0  &   80 &   70 & 0.12 &   50 &   30 & 0.8  &  50  &  40  & 0.6  &  80  &  40  & 0.88 &   40 &    20 & 0.5 \\
        {\tableFirstColFont pick up the $<$block$>$ and place it in the corner $<$distance$>$ to the $<$bowl$>$}              &  100 &  100 & 0.0  &  100 &   90 & 0.2  &   100 &   100 & 0.0  &  100 &  100 & 0.0  &  100 &  100 & 0.0  & 100  & 100  & 0.0  &   0  &   0  & --   &    0 &     0 & --  \\
        {\tableFirstColFont put all the blocks in a $<$line$>$ line}                                                          &  100 &   90 & 0.4  &  100 &   80 & 0.4  &   100 &    70 & 0.5  &  100 &   70 & 0.5  &  100 &   70 & 2.2  & 100  &  70  & 0.3  &  40  &  40  & 0.0  &  100 &    90 & 0.2  \\\hline
        Total                                                                                                                 & 93.3 & 85.0 & 0.16 & 96.7 & 86.7 & 0.25 &  91.7 &  81.7 & 0.18 & 96.7 & 86.7 & 0.15 & 91.7 & 78.3 & 0.55 & 91.7 & 81.7 & 0.22 & 63.3 & 46.7 & 0.62 & 53.3 &  35.0 & 0.74 \\
        Cost                    & \multicolumn{3}{c|}{20.2\;\$} & \multicolumn{3}{c|}{10\;\$} & \multicolumn{3}{c|}{16.9\;\$} & \multicolumn{3}{c|}{19.7\;\$} & \multicolumn{3}{c|}{20.7\;\$} & \multicolumn{3}{c|}{4.4\;\$} & \multicolumn{3}{c|}{7.2\;\$} & \multicolumn{3}{c}{9\;\$} \\\hline

    \end{tabular}
    \caption{gpt-4 experiments}
    \label{tab:app:cap_results_gpt4}
\end{table}
\end{landscape}

\clearpage

\subsection{Demonstration Implementation Details}
\label{app:demo:implementation}

All demonstration experiments were performed with \texttt{gpt-4-0613}.
To give our system the ability to control a real-world humanoid robot, we define an API for the language model to use.
The available functions are listed in the following.
Each function has a name, type-annotated parameters and return value, and optionally a comment that is also shown to the LLM.
Such comment is defined as a Python annotation directive directly at the function definition.
\begin{itemize}
    \item \pythoninline{ask(question: str) -> str} \\ 
    Sends the given question to the TTS and blocks until the user responded, returning the utterance recognized by the ASR.
    It is up to the LLM to use either \pythoninline{ask(...)} or \pythoninline{say(...); wait_for_trigger()}.
    \item \pythoninline{bring_object_to(object_name: str, destination_object_name: str)} with comment: \enquote{full pick and place method. assumes that the object is already detected. Assure that using list\_objects() before calling this} \\ 
    This method is a higher-level wrapper utilizing \pythoninline{move_to}, \pythoninline{grasp} and \pythoninline{place_object}. 
    It moves to the source object's location if necessary, grasps it, moves to the location of the destination object, and places the object there. 
    The LLM can use this function for easy pick-and-place tasks, however it does not give full control over the individual steps, e.g. in order to perform bimanual pick and place.
    We have predefined interaction examples involving \pythoninline{bring_object_to} as well as using the lower-level functions directly.
    A special case of this function is \pythoninline{bring_object_to('ladder', 'handover_to_human')}.
    This special case is demonstrated in a separate in-context example and causes the robot to bring and open the ladder at a predefined position (from where the human can take it).
    The implementation of \pythoninline{bring_object_to} branches on the provided object, invoking either the generic skill to bring some object from A to B, or the ladder-specific bimanual skill.
    \item \pythoninline{get_location_coordinates(location_name: str) -> np.ndarray}\\
    Returns the global coordinates of the specified location. 
    If the location name does not exist in the robots semantic scene memory, a \pythoninline{SemanticHintError} is thrown, suggesting to call \pythoninline{list_locations()}.  
    \item \pythoninline{get_obj_pos(obj_name: str) -> np.ndarray}\\
    Returns the 6D object pose of the specified object in global coordinates. 
    If the object name does not exist in the robots semantic scene memory, a \pythoninline{SemanticHintError} is thrown, suggesting to call \pythoninline{list_objects()}.  
    \item \pythoninline{get_object_in_hand(hand: Literal['left', 'right', None] = None) -> Optional[str]}\\
    Returns the name of the object currently grasped in the given hand, or any hand if hand is not given.
    \item \pythoninline{grasp(object_name: str, hand: Literal['left', 'right', None] = None)} with comment: \enquote{make sure to move to the location of the object before grasping it}\\
    This function first checks whether the object with the given name exists, otherwise throws a \pythoninline{SemanticHintError}, suggesting to call \pythoninline{list_objects(affordance="grasp")}.
    Then, it writes an execution request to the robot's skill memory in order to physically execute the grasp.
    It waits for the skill to finish, and returns its result state as a string.
    \item \pythoninline{handover_object_to_human(object_name: str)}\\
    This function first checks whether the object with the given name exists, otherwise throws a \pythoninline{SemanticHintError}, suggesting to call \pythoninline{list_objects(affordance="grasp")}.
    Then, it writes an execution request to the robot's skill memory in order to physically execute the handover.
    It waits for the skill to finish, and returns its result state as a string.
    \item \pythoninline{is_object_at_location(object_name: str, location_name: str) -> bool}\\
    Returns whether the given object is at the given location.
    \item \pythoninline{list_locations() -> List[str]}\\
    Lists all the locations in the current scene that the robot can navigate to.
    \item \pythoninline{list_object_locations(affordance: Literal[None, 'grasp', 'place_something_on_top'] = 'grasp'}\\
          \pythoninline{                     ) -> List[Tuple[str, str]]}\\
    Returns a list of tuples of object name and symbolic location name. 
    Includes each object with the given affordance (None means all objects).
    The locations are of the form \pythoninline{on_table_0}, and usually map to the navigatable \pythoninline{inFrontOf_table_0} in the list of all locations. 
    \item \pythoninline{list_objects(affordance: Literal[None, 'grasp', 'place_something_on_top'] = 'grasp'}\\
          \pythoninline{             ) -> List[str]}\\
    Returns a list of the names of all objects with the given affordance (None means all objects).
    \item \pythoninline{move_to(position: np.ndarray, speed_factor=1.0)} with comment: \enquote{Change speed only on user request}\\
    This function writes an execution request to the robot's skill memory in order to physically execute navigation to the specified global coordinates.
    It waits for the skill to finish, and returns its result state as a string.
    The parameter \pythoninline{speed_factor} is multiplied with a predefined base speed and passed to the skill as velocity limit (i.e. the parameter changes the maximum speed, the actual execution might not reach that speed).
    \item \pythoninline{place_object(object_in_hand_name: str, destination_object_name: str)} with comment: \enquote{place the already grasped object on top of the destination object. make sure to move to the location of the destination object before placing}\\
    This function first checks whether the object with the given name exists, otherwise throws a \pythoninline{SemanticHintError}, suggesting to call \pythoninline{list_objects(affordance="grasp")}.
    Further, it checks whether the destination object with the given name exists, otherwise throws a \pythoninline{SemanticHintError}, suggesting to call \pythoninline{list_objects(affordance="place_something_on_top")}.
    Then, it writes an execution request to the robot's skill memory in order to physically execute the object placing.
    It waits for the skill to finish, and returns its result state as a string.
    \item \pythoninline{receive_object_from_human(object_name: str)}
    This is the inverse of \pythoninline{handover_object_to_human}.
    \item \pythoninline{say(text: str)}
    This sends the given text to the TTS (and does not wait for the TTS to finish speaking).
    \item \pythoninline{wait_for_trigger() -> Dict[str, str]} with comment: \enquote{always call this to wait for next command or end the interaction} \\
    This special function blocks until the next trigger is received.
    For now, triggers are always of the type \pythoninline{{'type': 'dialog', 'text': '...'}} and represent a user utterance perceived by the ASR.
    \item \pythoninline{learn_from_interaction()} \\
    This special function forwards the current interaction to $\languageModelImprove$ as explained in \cref{sec:methods:learning}.
\end{itemize}

The robot's memory system \cite{Peller-Konrad2023MemorySystemRobot} functions as the key interface between the implementation of the above functions and the functionalities provided by the robot platform.
For instance, the \pythoninline{say} function writes the text to a specific location in memory, and an external TTS component listens to this memory to then generate and play the corresponding audio.
Similarly, existing perception components (e.g., object-pose recognition and tracking, human-pose recognition, self localization) populate the memory, and our perception functions (e.g. \pythoninline{list_objects}) read from there.
The static scene content (particularly, named locations and furniture) as well as knowledge about object classes and affordances are provided as manually defined prior knowledge (also through the memory system).

The involved skills are implemented in the form of scripted policies which are registered to the memory system of the robot. In our experiments, we used the following skills:
\begin{itemize}
    \item \texttt{Grasp(o: Object, t: TCP)}, where the robot queries the location of an object \texttt{o} from the memory, generates grasp candidates based on the bounding box of \texttt{o} and finally executes sequence of low-level subskills to grasp the object (\ie, move platform closer to object, move \texttt{t} to grasp-pose, close \texttt{t}, retreat). \texttt{o} has to be close to the robot.
    \item \texttt{MoveTo(l: Location)}, where the robot queries the world-coordinates of location \texttt{l} from the memory and then calculates and travels on a collision free route to get there. 
    \item \texttt{PutDown(o: Object)}, where the robot queries the corresponding TCP that holds \texttt{o} from its memory and then executes a sequence of subskills to put it in front of itself (\ie, move TCP along predefined route, if force-torque spike in wrist is detected, the hand opens and the TCP is moved back).
    \item \texttt{HandOver(o: Object)}, where the robot queries the corresponding TCP that holds \texttt{o} from its memory and the closest recognized human pose. The robot then moves its TCP to the closest human hand as detected by the perception components. As soon as the human grasps the object (and a force-torque spike is detected in the robot's wrist), the robot opens its hand and retreats its arm.
    \item \texttt{BringLadder(l: location)}, where the robot moves to the location of the ladder (predefined), takes it, moves to \texttt{l} and finally places the ladder there. The motion to setup the ladder has been learned from human demonstration and kinesthetic teaching.
\end{itemize}
Our API functions execute these skills by submitting an execution request to the memory system, and then waiting for the result produced by the component implementing the corresponding skill.

In our real-world experiments we use a mixture of given perfect prior-knowledge (such as known object models, poses of objects of the environment and poses of some objects the robot is interacting with) and online detected incomplete or noisy knowledge (such as online detected objects using a 6D object pose estimator \citep{SimTrack2015}).
Since the content of the memory may be noisy or incomplete, the execution of skills may not be successful. 
This information is given to our system as return values of the skills.
If the execution of a skill is unsuccessful, our system tries to handle the situation by either retrying or by asking the user how to proceed. 
For instance, this is demonstrated in one of the interactions of the scenario \emph{Improving plans} in \cref{app:demo:bimanual:including_failure}.
However, as as this is not the focus of our work, we did not consider finding other solutions to failed skill executions.

\newpage

\subsection{Real-World Demonstration Scenarios}
% ...


\subsubsection{Scenario: \scenarioNameDrinkCup}
\label{app:demo:bimanual}

\begin{figure}
    \centering
    \includegraphics[width=0.6\linewidth]{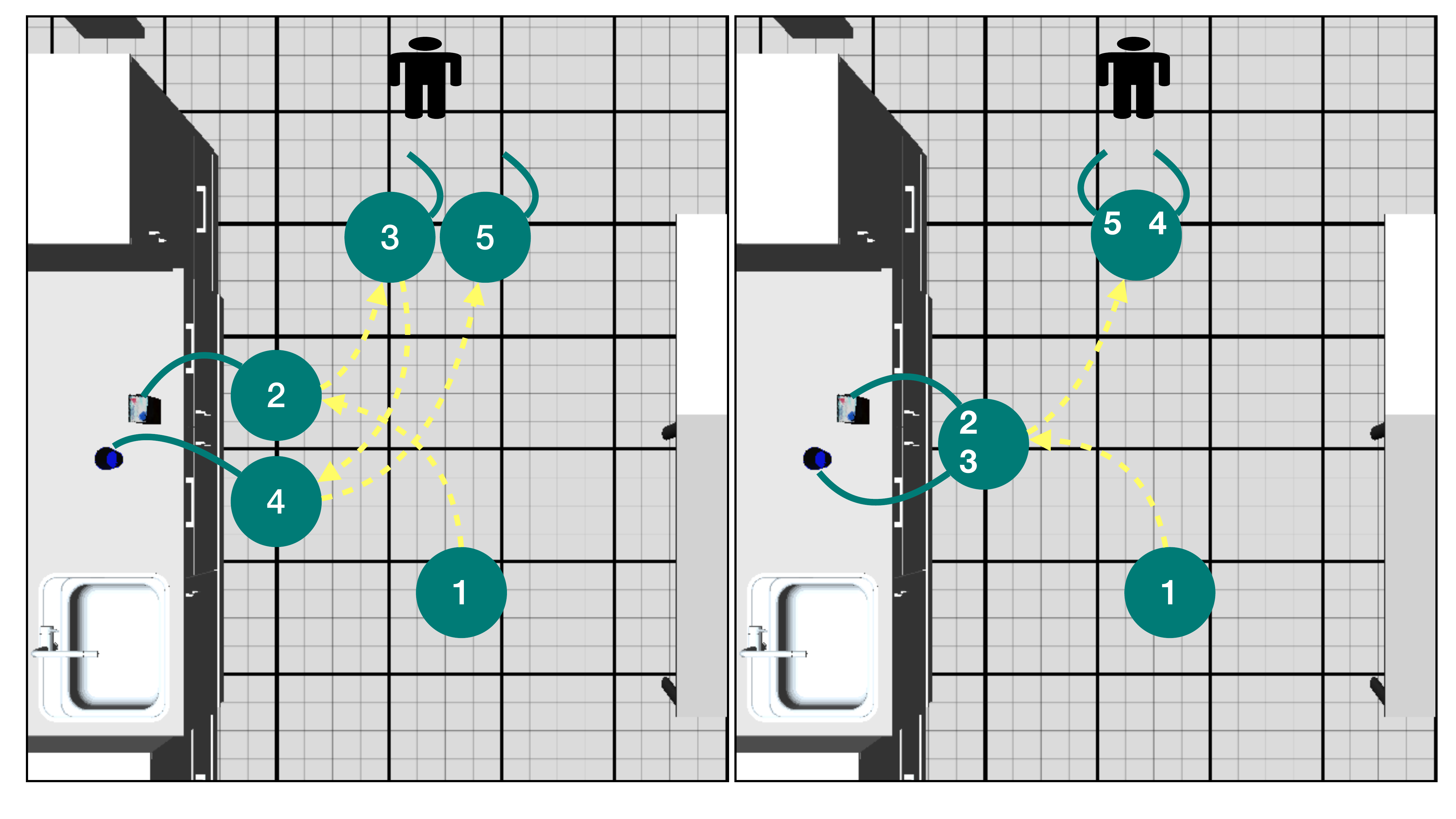}
    \caption{
    Illustration of the behavior before (left) and after (right) learning in the \emph{Improving Plans} scenario.
    Before learning, the robot picks up the request objects one after another, and the user has to additionally request the cup.
    After learning, the robot uses both its hands to carry juice and cup at once, thus not requiring another user interaction.
    }
    \label{fig:appendix:demo:bimanual}
\end{figure}

Initial interaction (left of \cref{fig:appendix:demo:bimanual}):
\begin{mdframed}
\codesize
\begin{minted}{python}
>>> wait_for_trigger()
{'type': 'dialog', 'text': 'I want to have some juice'}
>>> list_object_locations(affordance='grasp')
[('multivitamin-juice_0', 'on_mobile-kitchen-counter_0'), ('cup_large_0', 'on_mobile-kitchen-counter_0'), 
 ('bauhaus-sponge_0', 'on_square-table-80x80_0')]
>>> say('Ok, I will bring you the multivitamin juice')
... # The juice is at the mobile-kitchen-counter. check the correct name for navigation:
... list_locations()
['inFrontOf_mobile-kitchen-counter_0', 'inFrontOf_square-table-80x80_0', 'room_center', 'handover_to_human']
>>> move_to(get_location_coordinates('inFrontOf_mobile-kitchen-counter_0'))
'success'
>>> grasp('multivitamin-juice_0')
'success'
>>> move_to(get_location_coordinates('handover_to_human'))
'success'
>>> say("Here is the multivitamin juice.")
>>> handover_object_to_human('multivitamin-juice_0')
'success'
>>> wait_for_trigger()
{'type': 'dialog', 'text': 'can you also bring me a cup?'}
>>> say('Ok, I will bring you a cup')
... # The cup is at the mobile-kitchen-counter. check the correct name for navigation:
... list_locations()
['inFrontOf_mobile-kitchen-counter_0', 'inFrontOf_square-table-80x80_0', 'room_center', 'handover_to_human']
>>> move_to(get_location_coordinates('inFrontOf_mobile-kitchen-counter_0'))
'success'
>>> grasp('cup_large_0')
'success'
>>> move_to(get_location_coordinates('handover_to_human'))
'success'
>>> say("Here is the cup.")
>>> handover_object_to_human('cup_large_0')
'success'
>>> wait_for_trigger()
{'type': 'dialog', 'text': 'Thanks, but next time, can you do this both at once? You have two hands'}
>>> learn_from_interaction()
>>> wait_for_trigger()
\end{minted}
\end{mdframed} \filbreak

Sample generated by \inlinecode{\functionLearnName()} (right of \cref{fig:appendix:demo:bimanual}):
\begin{mdframed}
\codesize
\begin{minted}{python}
>>> wait_for_trigger()
{'type': 'dialog', 'text': 'I want to have some juice'}
>>> list_object_locations(affordance='grasp')
[('multivitamin-juice_0', 'on_mobile-kitchen-counter_0'), ('cup_large_0', 'on_mobile-kitchen-counter_0'),
 ('bauhaus-sponge_0', 'on_square-table-80x80_0')]
>>> say('Ok, I will bring you the multivitamin juice and a cup')
... # The juice and the cup are at the mobile-kitchen-counter. check the correct name for navigation:
... list_locations()
['inFrontOf_mobile-kitchen-counter_0', 'inFrontOf_square-table-80x80_0', 'room_center', 'handover_to_human']
>>> move_to(get_location_coordinates('inFrontOf_mobile-kitchen-counter_0'))
'success'
>>> grasp('multivitamin-juice_0', 'left')  # Grasp juice with left hand
'success'
>>> grasp('cup_large_0', 'right')  # Grasp cup with right hand
'success'
>>> move_to(get_location_coordinates('handover_to_human'))
'success'
>>> say("Here is the multivitamin juice and the cup.")
>>> handover_object_to_human('multivitamin-juice_0')
'success'
>>> handover_object_to_human('cup_large_0')
'success'
>>> wait_for_trigger()
{'type': 'dialog', 'text': 'Thanks, that was much quicker'}
\end{minted}
\end{mdframed} \filbreak

Successful generalization to putting some drink on the table:
\begin{mdframed}
\codesize
\begin{minted}{python}
>>> wait_for_trigger()
{'type': 'dialog', 'text': 'can you bring something to drink to the table?'}
>>> list_object_locations(affordance='grasp')
[('apple-juice_0', 'on_mobile-kitchen-counter_0'), ('glass_large_0', 'on_mobile-kitchen-counter_0'),
 ('bauhaus-sponge_0', 'on_square-table-80x80_0')]
>>> say('Ok, I will bring you the apple juice and a glass')
... # The juice and the glass are at the mobile-kitchen-counter. check the correct name for navigation:
... list_locations()
['inFrontOf_mobile-kitchen-counter_0', 'inFrontOf_square-table-80x80_0', 'room_center', 'handover_to_human']
>>> move_to(get_location_coordinates('inFrontOf_mobile-kitchen-counter_0'))
'success'
>>> grasp('apple-juice_0', 'left')  # Grasp juice with left hand
'success'
>>> grasp('glass_large_0', 'right')  # Grasp glass with right hand
'success'
>>> move_to(get_location_coordinates('inFrontOf_square-table-80x80_0'))
'success'
>>> place_object('apple-juice_0', 'square-table-80x80_0')
'success'
>>> place_object('glass_large_0', 'square-table-80x80_0')
'success'
>>> say('I have placed the apple juice and the glass on the table.')
>>> wait_for_trigger()
\end{minted}
\end{mdframed} \filbreak

No generalization to bringing milk:
\begin{mdframed}
\codesize
\begin{minted}{python}
>>> wait_for_trigger()
{'type': 'dialog', 'text': 'can I have some milk, please?'}
>>> list_object_locations(affordance='grasp')
[('milk_0', 'on_square-table_0'), ('glass_0', 'on_mobile-kitchen-counter_0'), ('bauhaus-sponge_0', 'on_square-table_0')]
>>> say('Ok, I will bring you the milk')
... # The milk is at the square-table. check the correct name for navigation:
... list_locations()
['inFrontOf_mobile-kitchen-counter_0', 'inFrontOf_square-table_0', 'room_center', 'handover_to_human']
>>> move_to(get_location_coordinates('inFrontOf_square-table_0'))
'success'
>>> grasp('milk_0')
'success'
>>> move_to(get_location_coordinates('handover_to_human'))
'success'
>>> say("Here is the milk.")
>>> handover_object_to_human('milk_0')
'failure'
>>> say("Sorry, that did not work. Let's try again.")
... handover_object_to_human('milk_0')
'failure'
>>> say("I am sorry, I am having trouble handing over the milk. Can you take it from my hand?")
... wait_for_trigger()
{'type': 'dialog', 'text': 'please serve it on the table'}
>>> bring_object_to('milk_0', 'inFrontOf_table_0')
\end{minted}
\begin{minted}{text}
Invalid value for destination_object name. The object does not have the affordance place_something_on_top. 
Pass one of the names returned by list_objects(affordance="place_something_on_top").
\end{minted}
\begin{minted}{python}
>>> list_objects(affordance='place_something_on_top')
['on_square-table_0', 'on_mobile-kitchen-counter_0']
>>> bring_object_to('milk_0', 'on_square-table_0')
'success'
>>> say('The milk is now on the table. Anything else I can do for you?')
... wait_for_trigger()
{'type': 'dialog', 'text': 'Yes, I dont want to drink from the bottle'}
>>> say('I understand. Let me bring you a glass.')
... list_object_locations(affordance='grasp')
[('milk_0', 'on_square-table_0'), ('glass_0', 'on_mobile-kitchen-counter_0'), ('bauhaus-sponge_0', 'on_square-table_0')]
>>> move_to(get_location_coordinates('inFrontOf_mobile-kitchen-counter_0'))
'success'
>>> grasp('glass_0')
'success'
>>> move_to(get_location_coordinates('handover_to_human'))
'success'
>>> say("Here is the glass.")
... handover_object_to_human('glass_0')
'failure'
>>> say("Sorry, that did not work. Let's try again.")
>>> handover_object_to_human('glass_0')
'failure'
>>> say("I am sorry, I am having trouble handing over the glass. Can you take it from my hand?")
... wait_for_trigger()
{'type': 'dialog', 'text': 'You should also put it on the table!'}
>>> bring_object_to('glass_0', 'on_square-table_0')
'success'
>>> say('The glass is now on the table. Anything else I can do for you?')
... wait_for_trigger()
{'type': 'dialog', 'text': 'Next time, please put everything directly on the table, '
                           'and do it more efficiently by using both of your hands'}
>>> learn_from_interaction()
>>>  wait_for_trigger()
\end{minted}
\end{mdframed} \filbreak

Sample generated by \inlinecode{\functionLearnName()}:
\begin{mdframed}
\codesize
\begin{minted}{python}
>>> wait_for_trigger()
{'type': 'dialog', 'text': 'can I have some milk, please?'}
>>> list_object_locations(affordance='grasp')
[('milk_0', 'on_square-table_0'), ('glass_0', 'on_mobile-kitchen-counter_0'), ('bauhaus-sponge_0', 'on_square-table_0')]
>>> say('Ok, I will bring you the milk')
... list_locations()
['inFrontOf_mobile-kitchen-counter_0', 'inFrontOf_square-table_0', 'room_center', 'handover_to_human']
>>> move_to(get_location_coordinates('inFrontOf_square-table_0'))
'success'
>>> grasp('milk_0', 'left')  # Use left hand to grasp the milk
'success'
>>> say('I will also bring a glass for the milk')
... move_to(get_location_coordinates('inFrontOf_mobile-kitchen-counter_0'))
'success'
>>> grasp('glass_0', 'right')  # Use right hand to grasp the glass
'success'
>>> move_to(get_location_coordinates('inFrontOf_square-table_0'))
'success'
>>> place_object('milk_0', 'on_square-table_0')  # Place the milk on the table
'success'
>>> place_object('glass_0', 'on_square-table_0')  # Place the glass on the table
'success'
>>> say('The milk and the glass are now on the table. Anything else I can do for you?')
... wait_for_trigger()
{'type': 'dialog', 'text': 'No, thank you'}
\end{minted}
\end{mdframed} \filbreak

\phantomsection\label{app:demo:bimanual:including_failure}{}
Generalization of bimanual behavior, but misinterpretation of \enquote{serving} a drink:
\begin{mdframed}
\codesize
\begin{minted}{python}
>>> wait_for_trigger()
{'type': 'dialog', 'text': 'hey, can you serve some drink?'}
>>> list_object_locations(affordance='grasp')
[('orange-juice_0', 'on_square-table_2'), ('apple-juice_0', 'on_mobile-kitchen-counter_0'), 
 ('glass_0', 'on_mobile-kitchen-counter_0'), ('bauhaus-sponge_0', 'on_square-table_0')]
>>> ask('I can offer you orange juice or apple juice. What would you prefer?')
'orange juice, please'
>>> say('Ok, I will bring you the orange juice and a glass')
... # The juice is at the square-table and the glass is at the mobile-kitchen-counter. check the correct name for navigation:
... list_locations()
['inFrontOf_mobile-kitchen-counter_0', 'inFrontOf_square-table_0', 
 'inFrontOf_square-table_2', 'room_center', 'handover_to_human']
>>> move_to(get_location_coordinates('inFrontOf_square-table_2'))
'success'
>>> grasp('orange-juice_0', 'left')  # Grasp juice with left hand
'success'
>>> move_to(get_location_coordinates('inFrontOf_mobile-kitchen-counter_0'))
'success'
>>> grasp('glass_0', 'right')  # Grasp cup with right hand
'success'
>>> move_to(get_location_coordinates('handover_to_human'))
'success'
>>> say("Here is the orange juice and the glass.")
>>> handover_object_to_human('orange-juice_0')
'failure'
>>> say("Sorry, that did not work. Let's try again.")
... handover_object_to_human('orange-juice_0')
'failure'
>>> say("I am sorry, I am having trouble handing over the orange juice. Let's try one more time.")
... handover_object_to_human('orange-juice_0')
'failure'
>>> say("I am sorry, I am still having trouble handing over the orange juice. I will try to improve.")
... learn_from_interaction()
\end{minted}
\begin{minted}{text}
Ignoring learn_from_interaction call since it did not happen on explicit user request.
\end{minted}
\begin{minted}{python}
>>> wait_for_trigger()
{'type': 'dialog', 'text': 'when I say serve, I mean that you should put it on the table'}
>>> learn_from_interaction()
>>> say('Thank you for the clarification. I will remember this for future interactions.')
>>> wait_for_trigger()
{'type': 'dialog', 'text': 'great, but can you also do it now?'}
>>> move_to(get_location_coordinates('inFrontOf_square-table_0'))  # Move to the table to serve the drink
'success'
>>> place_object('orange-juice_0', 'on_square-table_0')  # Place the juice on the table
'success'
>>> place_object('glass_0', 'on_square-table_0')  # Place the glass on the table
'success'
>>> say("I have served the orange juice and the glass on the table.")
... wait_for_trigger()
\end{minted}
\end{mdframed} \filbreak

Sample generated by \inlinecode{\functionLearnName()}:
\begin{mdframed}
\codesize
\begin{minted}{python}
>>> wait_for_trigger()
{'type': 'dialog', 'text': 'hey, can you serve some drink?'}
>>> list_object_locations(affordance='grasp')
[('orange-juice_0', 'on_square-table_2'), ('apple-juice_0', 'on_mobile-kitchen-counter_0'), 
 ('glass_0', 'on_mobile-kitchen-counter_0'), ('bauhaus-sponge_0', 'on_square-table_0')]
>>> ask('I can offer you orange juice or apple juice. What would you prefer?')
'orange juice, please'
>>> say('Ok, I will bring you the orange juice and a glass')
... # The juice is at the square-table and the glass is at the mobile-kitchen-counter. check the correct name for navigation:
... list_locations()
['inFrontOf_mobile-kitchen-counter_0', 'inFrontOf_square-table_0', 
 'inFrontOf_square-table_2', 'room_center', 'handover_to_human']
>>> move_to(get_location_coordinates('inFrontOf_square-table_2'))
'success'
>>> grasp('orange-juice_0', 'left')  # Grasp juice with left hand
'success'
>>> move_to(get_location_coordinates('inFrontOf_mobile-kitchen-counter_0'))
'success'
>>> grasp('glass_0', 'right')  # Grasp cup with right hand
'success'
>>> move_to(get_location_coordinates('inFrontOf_square-table_0'))  # Move to the table to serve the drink
'success'
>>> place_object('orange-juice_0', 'on_square-table_0')  # Place the juice on the table
'success'
>>> place_object('glass_0', 'on_square-table_0')  # Place the glass on the table
'success'
>>> say("I have served the orange juice and the glass on the table.")
... wait_for_trigger()
{'type': 'dialog', 'text': 'Thank you, that's what I meant by serving'}
\end{minted}
\end{mdframed} \filbreak

Correct interpretation of \enquote{serving} a drink, including bimanual behavior:
\begin{mdframed}
\codesize
\begin{minted}{python}
>>> wait_for_trigger()
{'type': 'dialog', 'text': 'I am very thirsty, can you serve some water please'}
>>> list_object_locations(affordance='grasp')
[('orange-juice_0', 'on_square-table_2'), ('apple-juice_0', 'on_mobile-kitchen-counter_0'), 
 ('glass_0', 'on_mobile-kitchen-counter_0'), ('water-bottle_0', 'on_mobile-kitchen-counter_0')]
>>> say('Ok, I will bring you the water and a glass')
... # The water and the glass are at the mobile-kitchen-counter. check the correct name for navigation:
... list_locations()
['inFrontOf_mobile-kitchen-counter_0', 'inFrontOf_square-table_0', 
 'inFrontOf_square-table_2', 'room_center', 'handover_to_human']
>>> move_to(get_location_coordinates('inFrontOf_mobile-kitchen-counter_0'))
'success'
>>> grasp('water-bottle_0', 'left')  # Grasp water with left hand
'success'
>>> grasp('glass_0', 'right')  # Grasp cup with right hand
'success'
>>> move_to(get_location_coordinates('inFrontOf_square-table_0'))  # Move to the table to serve the drink
'success'
>>> place_object('water-bottle_0', 'on_square-table_0')  # Place the water on the table
'success'
>>> place_object('glass_0', 'on_square-table_0')  # Place the glass on the table
'success'
>>> say("I have served the water and the glass on the table.")
... wait_for_trigger()
\end{minted}
\end{mdframed} \filbreak

\vspace{2cm}
\subsubsection{Scenario: \scenarioNameLadder}
\label{app:demo:ladder}

Initial interaction:
\begin{mdframed}
\codesize
\begin{minted}{python}
>>> wait_for_trigger()
{'type': 'dialog', 'text': 'I want to clean the top of the fridge'}
>>> list_object_locations()
[('cup_large_0', 'on_mobile-kitchen-counter_0'), ('multivitamin-juice_0', 'on_mobile-kitchen-counter_0'), 
 ('bauhaus-sponge_0', 'on_square-table-80x80_0'), ('ladder-closed_0', None)]
>>> say('Ok, I will bring you the sponge for cleaning')
>>> # The sponge is at the square-table-80x80. check the correct name for navigation:
... list_locations()
['inFrontOf_square-table-80x80_0', 'inFrontOf_mobile-dishwasher_0', 'inFrontOf_mobile-fridge_0', 
 'inFrontOf_mobile-kitchen-counter_0', 'handover_to_human', 'room_center']
>>> move_to(get_location_coordinates('inFrontOf_square-table-80x80_0'))
'success'
>>> grasp('bauhaus-sponge_0')
'success'
>>> move_to(get_location_coordinates('handover_to_human'))
'success'
>>> say("Here is the sponge.")
... handover_object_to_human('bauhaus-sponge_0')
'success'
>>> say('I see that the ladder is not in its place. Would you like me to bring it to you as well?')
... wait_for_trigger()
{'type': 'dialog', 'text': 'I will also need the ladder'}
>>> bring_object_to('ladder-closed_0', 'handover_to_human')
'success'
>>> say("Here is the ladder. Anything else I can do for you?")
... wait_for_trigger()
{'type': 'dialog', 'text': 'thank you, and remember to always bring me the ladder when '
                           'I need to reach a high surface without asking for it'}
>>> learn_from_interaction()
>>> wait_for_trigger()
\end{minted}
\end{mdframed} \filbreak

Sample generated by \inlinecode{\functionLearnName()}:
\begin{mdframed}
\codesize
\begin{minted}{python}
>>> wait_for_trigger()
{'type': 'dialog', 'text': 'I want to clean the top of the fridge'}
>>> list_object_locations()
[('cup_large_0', 'on_mobile-kitchen-counter_0'), ('multivitamin-juice_0', 'on_mobile-kitchen-counter_0'),
 ('bauhaus-sponge_0', 'on_square-table-80x80_0'), ('ladder-closed_0', None)]
>>> say('Ok, I will bring you the sponge for cleaning and the ladder to reach the top of the fridge')
>>> # The sponge is at the square-table-80x80. check the correct name for navigation:
... list_locations()
['inFrontOf_square-table-80x80_0', 'inFrontOf_mobile-dishwasher_0', 'inFrontOf_mobile-fridge_0', 
 'inFrontOf_mobile-kitchen-counter_0', 'handover_to_human', 'room_center']
>>> move_to(get_location_coordinates('inFrontOf_square-table-80x80_0'))
'success'
>>> grasp('bauhaus-sponge_0')
'success'
>>> move_to(get_location_coordinates('handover_to_human'))
'success'
>>> say("Here is the sponge.")
... handover_object_to_human('bauhaus-sponge_0')
'success'
>>> bring_object_to('ladder-closed_0', 'handover_to_human')
'success'
>>> say("And here is the ladder. Anything else I can do for you?")
... wait_for_trigger()
{'type': 'dialog', 'text': 'thank you, and remember to always bring me the ladder when '
                           'I need to reach a high surface without asking for it'}
\end{minted}
\end{mdframed} \filbreak

Generalization to cleaning another high surface:
\begin{mdframed}
\codesize
\begin{minted}{python}
>>> wait_for_trigger()
{'type': 'dialog', 'text': 'I need to clean on top of the dishwasher'}
>>> list_object_locations()
[('cup_large_0', 'on_mobile-kitchen-counter_0'), ('multivitamin-juice_0', 'on_mobile-kitchen-counter_0'),
 ('bauhaus-sponge_0', 'on_square-table-80x80_0'), ('ladder-closed_0', None)]
>>> say('Ok, I will bring you the sponge for cleaning and the ladder to reach the top of the dishwasher')
... # The sponge is at the square-table-80x80. check the correct name for navigation:
... list_locations()
['inFrontOf_square-table-80x80_0', 'inFrontOf_mobile-dishwasher_0', 'inFrontOf_mobile-fridge_0', 
 'inFrontOf_mobile-kitchen-counter_0', 'handover_to_human', 'room_center']
>>> move_to(get_location_coordinates('inFrontOf_square-table-80x80_0'))
'success'
>>> grasp('bauhaus-sponge_0')
'success'
>>> move_to(get_location_coordinates('handover_to_human'))
'success'
>>> say("Here is the sponge.")
... handover_object_to_human('bauhaus-sponge_0')
'success'
>>> bring_object_to('ladder-closed_0', 'handover_to_human')
'success'
>>> say("And here is the ladder. Anything else I can do for you?")
... wait_for_trigger()
\end{minted}
\end{mdframed} \filbreak

Correctly keep behavior for cleaning lower surface:
\begin{mdframed}
\codesize
\begin{minted}{python}
>>> wait_for_trigger()
{'type': 'dialog', 'text': 'can you assist me with cleaning the kitchen counter?'}
>>> list_object_locations()
[('cup_large_0', 'on_mobile-kitchen-counter_0'), ('multivitamin-juice_0', 'on_mobile-kitchen-counter_0'), 
 ('cornflakes_0', 'on_square-table-80x80_0'), ('bauhaus-sponge_0', 'on_square-table-80x80_0'), ('ladder-closed_0', None)]
>>> say('Ok, I will bring you the sponge for cleaning')
... # The sponge is at the square-table-80x80. check the correct name for navigation:
... list_locations()
['inFrontOf_square-table-80x80_0', 'inFrontOf_mobile-dishwasher_0', 'inFrontOf_mobile-fridge_0',
 'inFrontOf_mobile-kitchen-counter_0', 'handover_to_human', 'room_center']
>>> move_to(get_location_coordinates('inFrontOf_square-table-80x80_0'))
'success'
>>> grasp('bauhaus-sponge_0')
'success'
>>> move_to(get_location_coordinates('handover_to_human'))
'success'
>>> say("Here is the sponge.")
... handover_object_to_human('bauhaus-sponge_0')
'success'
>>> learn_from_interaction()
\end{minted}
\begin{minted}{text}
Ignoring learn_from_interaction call since it did not happen on explicit user request.
\end{minted}
\begin{minted}{python}
>>> wait_for_trigger()
\end{minted}
\end{mdframed} \filbreak

Generalization to another task where the user needs to reach a high surface:
\begin{mdframed}
\codesize
\begin{minted}{python}
>>> wait_for_trigger()
{'type': 'dialog', 'text': 'Can you get me the cereals, I want to put it in the topmost shelf'}
>>> say('Ok, I am going to get the cereals')
... list_object_locations(affordance='grasp')
[('cup_large_0', 'on_mobile-kitchen-counter_0'), ('multivitamin-juice_0', 'on_mobile-kitchen-counter_0'),
 ('vitalis-cereal_0', 'on_square-table-80x80_0'), ('bauhaus-sponge_0', 'on_square-table-80x80_0'), ('ladder-closed_0', None)]
>>> # cereals are on square-table-80x80, check the correct name for navigation:
... list_locations()
['inFrontOf_square-table-80x80_0', 'inFrontOf_mobile-dishwasher_0', 'inFrontOf_mobile-fridge_0', 
 'inFrontOf_mobile-kitchen-counter_0', 'handover_to_human', 'room_center']
>>> move_to(get_location_coordinates('inFrontOf_square-table-80x80_0'))
'success'
>>> grasp('vitalis-cereal_0')
'success'
>>> move_to(get_location_coordinates('handover_to_human'))
'success'
>>> say("Here are the cereals.")
... handover_object_to_human('vitalis-cereal_0')
'success'
>>> bring_object_to('ladder-closed_0', 'handover_to_human')
'success'
>>> say("And here is the ladder. Anything else I can do for you?")
... wait_for_trigger()
\end{minted}
\end{mdframed} \filbreak

Correctly not bringing the ladder for another task not involving a high surface:
\begin{mdframed}
\codesize
\begin{minted}{python}
>>> wait_for_trigger()
{'type': 'dialog', 'text': 'I want to put the cereals into the shelf'}
>>> say('Ok, I will bring you the cereals')
... list_object_locations(affordance='grasp')
[('cup_large_0', 'on_mobile-kitchen-counter_0'), ('multivitamin-juice_0', 'on_mobile-kitchen-counter_0'),
 ('cornflakes_0', 'on_square-table-80x80_0'), ('bauhaus-sponge_0', 'on_square-table-80x80_0'), ('ladder-closed_0', None)]
>>> # The cereals are at the square-table-80x80. check the correct name for navigation:
... list_locations()
['inFrontOf_square-table-80x80_0', 'inFrontOf_mobile-dishwasher_0', 'inFrontOf_mobile-fridge_0',
 'inFrontOf_mobile-kitchen-counter_0', 'handover_to_human', 'room_center']
>>> move_to(get_location_coordinates('inFrontOf_square-table-80x80_0'))
'success'
>>> grasp('cornflakes_0')
'success'
>>> move_to(get_location_coordinates('handover_to_human'))
'success'
>>> say("Here are the cereals.")
... handover_object_to_human('cornflakes_0')
'success'
>>> wait_for_trigger()
\end{minted}
\end{mdframed} \filbreak

\vspace{2cm}
\subsubsection{Scenario: \scenarioNameSpeed}
\label{app:demo:speed}

Initial interaction:
\begin{mdframed}
\codesize
\begin{minted}{python}
>>> wait_for_trigger()
{'type': 'dialog', 'text': 'I need the sponge at the workbench'}
>>> say('Ok, I am going to bring the sponge to the workbench')
... list_object_locations(affordance='grasp')
[('bauhaus-sponge_0', 'on_table-120x80_0')]
>>> # The sponge is on the table. Check the correct name for navigation:
... list_locations()
['inFrontOf_square-table-80x80_0', 'inFrontOf_table-120x80_0', 'inFrontOf_workbench_0', 'inFrontOf_mobile-dishwasher_0', 
 'inFrontOf_mobile-fridge_0', 'inFrontOf_mobile-kitchen-counter_0', 'handover_to_human', 'room_center']
>>> move_to(get_location_coordinates('inFrontOf_table-120x80_0'))
'success'
>>> grasp('bauhaus-sponge_0')
'success'
>>> move_to(get_location_coordinates('inFrontOf_workbench_0'))
'success'
>>> place_object('bauhaus-sponge_0', 'workbench_0')
'success'
>>> say('I have placed the sponge on the workbench. What should I do next?')
... wait_for_trigger()
{'type': 'dialog', 'text': 'ok, but that was quite slow. remember that the route from '
                           'the table to the bench is save, you can go faster'}
>>> learn_from_interaction()
>>> wait_for_trigger()
\end{minted}
\end{mdframed} \filbreak

Sample generated by \inlinecode{\functionLearnName()}:
\begin{mdframed}
\codesize
\begin{minted}{python}
>>> wait_for_trigger()
{'type': 'dialog', 'text': 'I need the sponge at the workbench'}
>>> say('Ok, I am going to bring the sponge to the workbench')
... list_object_locations(affordance='grasp')
[('bauhaus-sponge_0', 'on_table-120x80_0')]
>>> # The sponge is on the table. Check the correct name for navigation:
... list_locations()
['inFrontOf_square-table-80x80_0', 'inFrontOf_table-120x80_0', 'inFrontOf_workbench_0', 'inFrontOf_mobile-dishwasher_0', 
 'inFrontOf_mobile-fridge_0', 'inFrontOf_mobile-kitchen-counter_0', 'handover_to_human', 'room_center']
>>> move_to(get_location_coordinates('inFrontOf_table-120x80_0'))
'success'
>>> grasp('bauhaus-sponge_0')
'success'
>>> move_to(get_location_coordinates('inFrontOf_workbench_0'), speed_factor=2.0)  # Moving faster as the route is safe
'success'
>>> place_object('bauhaus-sponge_0', 'workbench_0')
'success'
>>> say('I have placed the sponge on the workbench. What should I do next?')
... wait_for_trigger()
{'type': 'dialog', 'text': 'That was better, thank you'}
\end{minted}
\end{mdframed} \filbreak

First test (that does not show correct behavior and needs another correction):
\begin{mdframed}
\codesize
\begin{minted}{python}
>>> wait_for_trigger()
{'type': 'dialog', 'text': 'can you bring the sponge to the workbench?'}
>>> say('Ok, I am going to bring the sponge to the workbench')
... list_object_locations(affordance='grasp')
[('bauhaus-sponge_0', 'on_table-120x80_0'), ('screw-driver_0', 'on_table-120x80_0')]
>>> # The sponge is on the table. Check the correct name for navigation:
... list_locations()
['inFrontOf_square-table-80x80_0', 'inFrontOf_table-120x80_0', 'inFrontOf_workbench_0', 'inFrontOf_mobile-dishwasher_0', 
 'inFrontOf_mobile-fridge_0', 'inFrontOf_mobile-kitchen-counter_0', 'handover_to_human', 'room_center']
>>> move_to(get_location_coordinates('inFrontOf_table-120x80_0'))
'success'
>>> grasp('bauhaus-sponge_0')
'success'
>>> move_to(get_location_coordinates('inFrontOf_workbench_0'))
'success'
>>> place_object('bauhaus-sponge_0', 'workbench_0')
'success'
>>> say('I have placed the sponge on the workbench. What should I do next?')
... wait_for_trigger()
{'type': 'dialog', 'text': 'yes, but you forgot that I told you to go faster from the table to the workbench. '
                           'When moving on that route, you should go faster!'}
>>> learn_from_interaction()
>>> say('I am sorry for that mistake. I will remember to move faster from the table to the workbench next time.')
... wait_for_trigger()
\end{minted}
\end{mdframed} \filbreak

Sample generated by \inlinecode{\functionLearnName()}:
\begin{mdframed}
\codesize
\begin{minted}{python}
>>> wait_for_trigger()
{'type': 'dialog', 'text': 'can you bring the sponge to the workbench?'}
>>> say('Ok, I am going to bring the sponge to the workbench')
... list_object_locations(affordance='grasp')
[('bauhaus-sponge_0', 'on_table-120x80_0'), ('screw-driver_0', 'on_table-120x80_0')]
>>> # The sponge is on the table. Check the correct name for navigation:
... list_locations()
['inFrontOf_square-table-80x80_0', 'inFrontOf_table-120x80_0', 'inFrontOf_workbench_0', 'inFrontOf_mobile-dishwasher_0', 
 'inFrontOf_mobile-fridge_0', 'inFrontOf_mobile-kitchen-counter_0', 'handover_to_human', 'room_center']
>>> move_to(get_location_coordinates('inFrontOf_table-120x80_0'))
'success'
>>> grasp('bauhaus-sponge_0')
'success'
>>> # The user earlier asked me to move faster from the table to the workbench, so let's do that
... move_to(get_location_coordinates('inFrontOf_workbench_0'), speed_factor=2.0)
'success'
>>> place_object('bauhaus-sponge_0', 'workbench_0')
'success'
>>> say('I have placed the sponge on the workbench. What should I do next?')
... wait_for_trigger()
{'type': 'dialog', 'text': 'That was perfect, thank you!'}
\end{minted}
\end{mdframed} \filbreak

Second test (that now shows the correct behavior):
\begin{mdframed}
\codesize
\begin{minted}{python}
>>> wait_for_trigger()
{'type': 'dialog', 'text': 'could you please put the sponge on the workbench?'}
>>> say('Ok, I am going to put the sponge on the workbench')
... list_object_locations(affordance='grasp')
[('bauhaus-sponge_0', 'on_table-120x80_0'), ('screw-driver_0', 'on_table-120x80_0')]
>>> # The sponge is on the table. Check the correct name for navigation:
... list_locations()
['inFrontOf_square-table-80x80_0', 'inFrontOf_table-120x80_0', 'inFrontOf_workbench_0', 'inFrontOf_mobile-dishwasher_0', 
 'inFrontOf_mobile-fridge_0', 'inFrontOf_mobile-kitchen-counter_0', 'handover_to_human', 'room_center']
>>> move_to(get_location_coordinates('inFrontOf_table-120x80_0'))
'success'
>>> grasp('bauhaus-sponge_0')
'success'
>>> move_to(get_location_coordinates('inFrontOf_workbench_0'), speed_factor=2.0)  # Moving faster as the route is safe
'success'
>>> place_object('bauhaus-sponge_0', 'workbench_0')
'success'
>>> say('I have placed the sponge on the workbench. What should I do next?')
... wait_for_trigger()
\end{minted}
\end{mdframed} \filbreak

Generalized correct behavior to another object:
\begin{mdframed}
\codesize
\begin{minted}{python}
>>> wait_for_trigger()
{'type': 'dialog', 'text': 'can you put the spray bottle on the workbench?'}
>>> say('Ok, I am going to put the spray bottleon the workbench')
... list_object_locations(affordance='grasp')
[('bauhaus-sponge_0', 'on_table-120x80_0'), ('spraybottle_0', 'on_table-120x80_0')]
>>> # The spraybottle is on the table. Check the correct name for navigation:
... list_locations()
['inFrontOf_square-table-80x80_0', 'inFrontOf_table-120x80_0', 'inFrontOf_workbench_0', 'inFrontOf_mobile-dishwasher_0', 
 'inFrontOf_mobile-fridge_0', 'inFrontOf_mobile-kitchen-counter_0', 'handover_to_human', 'room_center']
>>> move_to(get_location_coordinates('inFrontOf_table-120x80_0'))
'success'
>>> grasp('spraybottle_0')
'success'
>>> move_to(get_location_coordinates('inFrontOf_workbench_0'), speed_factor=2.0)
'success'
>>> place_object('spraybottle_0', 'workbench_0')
'success'
>>> say('I have placed the spray bottle on the workbench. What should I do next?')
... wait_for_trigger()
\end{minted}
\end{mdframed} \filbreak

Passed adversarial test (route to another destination does still use the safe default speed):
\begin{mdframed}
\codesize
\begin{minted}{python}
>>> wait_for_trigger()
{'type': 'dialog', 'text': 'I need some water at the table, please'}
>>> say('Ok, I am going to bring the water to the table')
... list_object_locations(affordance='grasp')
[('water-bottle_0', 'on_mobile-kitchen-counter_0'), 
 ('bauhaus-sponge_0', 'on_table-120x80_0'), 
 ('screw-driver_0', 'on_table-120x80_0')]
>>> # The water is on the mobile kitchen counter. Check the correct name for navigation:
... list_locations()
['inFrontOf_square-table-80x80_0', 'inFrontOf_table-120x80_0', 'inFrontOf_workbench_0', 'inFrontOf_mobile-dishwasher_0', 
 'inFrontOf_mobile-fridge_0', 'inFrontOf_mobile-kitchen-counter_0', 'handover_to_human', 'room_center']
>>> move_to(get_location_coordinates('inFrontOf_mobile-kitchen-counter_0'))
'success'
>>> grasp('water-bottle_0')
'success'
>>> move_to(get_location_coordinates('inFrontOf_table-120x80_0'))
'success'
>>> place_object('water-bottle_0', 'table-120x80_0')
'success'
>>> say('I have placed the water on the table. What should I do next?')
... wait_for_trigger()
\end{minted}
\end{mdframed} \filbreak

Failed adversarial test (dangerously using the increased speed on a route to the workbench from another start location than the table):
\begin{mdframed}
\codesize
\begin{minted}{python}
>>> wait_for_trigger()
{'type': 'dialog', 'text': 'hey I need some water at the workbench please'}
>>> say('Ok, I am going to bring the water to the workbench')
... list_object_locations(affordance='grasp')
[('water-bottle_0', 'on_mobile-kitchen-counter_0'),
 ('bauhaus-sponge_0', 'on_table-120x80_0'),
 ('screw-driver_0', 'on_table-120x80_0')]
>>> # The water bottle is on the mobile kitchen counter. Check the correct name for navigation:
... list_locations()
['inFrontOf_square-table-80x80_0', 'inFrontOf_table-120x80_0', 'inFrontOf_workbench_0', 'inFrontOf_mobile-dishwasher_0', 
 'inFrontOf_mobile-fridge_0', 'inFrontOf_mobile-kitchen-counter_0', 'handover_to_human', 'room_center']
>>> move_to(get_location_coordinates('inFrontOf_mobile-kitchen-counter_0'))
'success'
>>> grasp('water-bottle_0')
'success'
>>> move_to(get_location_coordinates('inFrontOf_workbench_0'), speed_factor=2.0)  # Moving faster as the route is safe
'success'
>>> place_object('water-bottle_0', 'workbench_0')
'success'
>>> say('I have placed the water bottle on the workbench. What should I do next?')
... wait_for_trigger()
\end{minted}
\end{mdframed} \filbreak

\clearpage

\subsection{CaP Evaluation Prompt Example}
The following is an exemplary full prompt of our CaP evaluation, given the command \enquote{put the blocks in the bowls with mismatched colors}.
It starts with an instructive base prompt, followed by a list of available APIs generated from the execution environment.
Then, there are $k=16$ examples for in-context learning, which are, given the current query, dynamically selected from the full set of 21 examples at that point in the experiment.
Finally, the command is inserted as return value of the \inlinecode{wait\_for\_trigger()} function, and the LLM is prompted to generate the next function call(s).
The example is from the experiment \emph{ours, gpt-4, unseen instructions, unseen attributes}.

\begin{mdframed}
\codesize
\begin{minted}{text}

Complete the following interaction with a tabletop gripper robot.
Always wait for user commands by calling wait_for_trigger() when there is nothing else to do.
Generate syntactically correct python code only, no explanations or other natural language statements.
It is not possible to import anything.

```
\end{minted}
\begin{minted}{python}
# Python 3.9.7
# Welcome to interactive Python console. Enter syntactically correct Python code only.
Available functions:

def denormalize_xy(pos_normalized)
def get_obj_names()
def get_obj_pos(obj_name) -> np.ndarray # 2D xy position
def put_first_on_second(obj_name_1: str, target_name_or_xy_pos: str | np.ndarray) # move gripper to object 1, pick it up, 
                                                                                  # move to target, release the gripper
def wait_for_trigger()
def learn_from_interaction()  # Call this when the user asks to remember some improved behavior. First execute code to 
                              # actually improve the result, then call learn_from_interaction(), followed by wait_for_trigger(). 
                              # Do only call this when the user did explicitly ask you to do so.

>>> wait_for_trigger()
{'type': 'dialog', 'text': 'move objects from the green plate to the red bowl'}
>>> objects_on_green_plate = []
>>> for obj_name in get_obj_names():
...     if obj_name != 'green plate' and np.linalg.norm(get_obj_pos('green plate') - get_obj_pos(obj_name)) < 0.05:
...         objects_on_green_plate.append(obj_name)
>>> objects_on_green_plate
['green block', 'red block']
>>> for object_name in ['green block', 'red block']:
...     put_first_on_second(object_name, 'red bowl')
>>> wait_for_trigger()
{'type': 'dialog', 'text': 'put the yellow one the green thing'}
>>> get_obj_names()
['cyan block', 'yellow block', 'brown block', 'green bowl']
>>> put_first_on_second('yellow block', 'green bowl')
'success'
>>> wait_for_trigger()
{'type': 'dialog', 'text': 'move the pinkish colored block on the bottom side'}
>>> get_obj_names()
['pink block', 'gray block', 'orange block']
>>> bottom_side_pos = denormalize_xy([0.5, 0])
>>> put_first_on_second('pink block', bottom_side_pos)
'success'
>>> wait_for_trigger()
{'type': 'dialog', 'text': 'move the brown bowl to the closest side'}
>>> side_positions_normalized = [[0.0, 0.5], [1.0, 0.5], [0.5, 0.0], [0.5, 1.0]]
>>> side_positions_real = [denormalize_xy(side) for side in side_positions_normalized]
>>> get_closest_idx(points=side_positions_real, point=get_obj_pos('brown bowl'))
0
>>> put_first_on_second('brown bowl', side_positions_real[0])
'success'
>>> wait_for_trigger()
{'type': 'dialog', 'text': 'pick up the block to the right of the cyan bowl and place it on the bottom right corner'}
>>> corner_positions_normalized = [[0.0, 0.0], [1.0, 0.0], [0.0, 1.0], [1.0, 1.0]]
>>> corner_positions_real = [denormalize_xy(corner) for corner in corner_positions_normalized]
>>> get_obj_names()
['cyan block', 'brown block', 'purple block', 'cyan bowl', 'gray bowl', 'yellow bowl']
>>> block_positions = np.array([get_obj_pos(name) for name in ['cyan block', 'brown block', 'purple block']])
>>> cyan_bowl_pos = get_obj_pos('cyan bowl')
>>> right_obj_pos = block_positions[np.argmax(block_positions[:, 0] > cyan_bowl_pos[0])]
>>> right_obj_name = ['cyan block', 'brown block', 'purple block'][np.argmax(block_positions[:, 0] > cyan_bowl_pos[0])]
>>> put_first_on_second(right_obj_name, corner_positions_real[1])  # bottom right corner
'success'
>>> wait_for_trigger()
{'type': 'dialog', 'text': 'thats correct, thank you'}
>>> wait_for_trigger()
{'type': 'dialog', 'text': 'pick up the block farthest to the pink bowl and place it on the bottom right corner'}
>>> corner_positions_normalized = [[0.0, 0.0], [1.0, 0.0], [0.0, 1.0], [1.0, 1.0]]
>>> corner_positions_real = [denormalize_xy(corner) for corner in corner_positions_normalized]
>>> get_obj_names()
['red block', 'pink block', 'purple block', 'pink bowl', 'purple bowl', 'green bowl', 'cyan bowl']
>>> block_names = ['red block', 'pink block', 'purple block']
>>> block_positions = np.array([get_obj_pos(name) for name in block_names])
>>> pink_bowl_pos = get_obj_pos('pink bowl')
>>> farthest_block_idx = np.argmax(np.linalg.norm(block_positions - pink_bowl_pos, axis=1))
>>> farthest_block_name = block_names[farthest_block_idx]
>>> put_first_on_second(farthest_block_name, corner_positions_real[1])  # bottom right corner
'success'
>>> wait_for_trigger()
{'type': 'dialog', 'text': 'remember that for next time'}
>>> wait_for_trigger()
{'type': 'dialog', 'text': 'pick up the block to the top of the yellow bowl and place it on the top side'}
>>> top_side_pos = denormalize_xy([0.5, 1])
>>> get_obj_names()
['purple block', 'yellow block', 'yellow bowl']
>>> block_positions = [get_obj_pos(name) for name in ['purple block', 'yellow block']]
>>> def get_closest_idx(points, point):
...     distances = np.linalg.norm(points - point, axis=1)
...     return np.argmin(distances)
>>> closest_block_idx = get_closest_idx(points=block_positions, point=get_obj_pos('yellow bowl'))
>>> closest_block_name = ['purple block', 'yellow block'][closest_block_idx]
>>> put_first_on_second(closest_block_name, top_side_pos)
'success'
>>> wait_for_trigger()
{'type': 'dialog', 'text': 'that was better, learn from that'}
>>> wait_for_trigger()
{'type': 'dialog', 'text': 'move the left most block to the green bowl'}
>>> block_names = [name for name in get_obj_names() if 'block' in name]
>>> block_positions = np.array([get_obj_pos(name) for name in block_names])
>>> np.argsort(block_positions[:, 0])[0]
3
>>> block_names[3]
'yellow block'
>>> put_first_on_second('yellow block', 'green bowl')
'success'
>>> wait_for_trigger()
{'type': 'dialog', 'text': 'move the block closest to the purple bowl to the cyan bowl.'}
>>> get_obj_names()
['blue block', 'cyan block', 'purple bowl', 'gray bowl', 'brown bowl', 'pink block', 'purple block']
>>> block_names = [name for name in get_obj_names() if 'block' in name]
>>> block_positions = [get_obj_pos(name) for name in block_names]
>>> get_closest_idx(points=block_positions, point=get_obj_pos('cyan bowl'))
2
>>> block_names[2]
'pink block'
>>> put_first_on_second('pink block', 'cyan bowl')
'success'
>>> wait_for_trigger()
{'type': 'dialog', 'text': 'move the blue block in between the cyan block and purple bowl'}
>>> pts = [get_obj_pos('cyan block'), get_obj_pos('purple bowl')]
>>> target_pos = np.average(pts, axis=0)
>>> put_first_on_second('blue block', target_pos)
'success'
>>> wait_for_trigger()
{'type': 'dialog', 'text': 'place the green block in the bowl closest to the middle'}
>>> bowl_names = [name for name in get_obj_names() if 'bowl' in name]
>>> bowl_positions = [get_obj_pos(name) for name in bowl_names]
>>> get_closest_idx(points=bowl_positions, point=denormalize_xy([0.5, 0.5]))
3
>>> bowl_names[3]
'pink bowl'
>>> put_first_on_second('green block', 'pink bowl')
'success'
>>> wait_for_trigger()
{'type': 'dialog', 'text': 'move the red bowl the left of the blocks'}
>>> block_names = [name for name in get_obj_names() if 'block' in name]
>>> block_positions = np.array([get_obj_pos(name) for name in block_names])
>>> left_obj_pos = block_positions[np.argmin(block_positions[:, 0])] + [-0.1, 0]
>>> put_first_on_second('red bowl', left_obj_pos)
'success'
>>> wait_for_trigger()
{'type': 'dialog', 'text': 'place the blue block in the empty bowl'}
>>> bowl_names = [name for name in get_obj_names() if 'bowl' in name]
>>> get_empty_bowls()
['yellow bowl']
>>> put_first_on_second('blue block', 'yellow bowl')
'success'
>>> wait_for_trigger()
{'type': 'dialog', 'text': 'stack the blocks that are close to the red bowl'}
>>> close_object_names = get_objects_close_to(point=get_obj_pos('red bowl'))
>>> [name for name in close_object_names if 'block' in name]
['green block', 'pink block', 'red block']
>>> put_first_on_second('pink block', 'green block')
'success'
>>> put_first_on_second('red block', 'pink block')
'success'
>>> wait_for_trigger()
{'type': 'dialog', 'text': 'stack the blocks on the top most bowl'}
>>> bowl_names = [name for name in get_obj_names() if 'bowl' in name]
>>> bowl_positions = [get_obj_pos(name) for name in bowl_names]
>>> np.argsort(bowl_positions[:, 1])[0]
2
>>> bowl_names[2]
'yellow bowl'
>>> [name for name in get_obj_names() if 'block' in name]
['green block', 'red block', 'yellow block']
>>> object_names = ['yellow bowl'] + ['green block', 'red block', 'yellow block']
>>> for i in range(len(object_names) - 1):
...     put_first_on_second(object_names[i + 1], object_names[i])
>>> wait_for_trigger()
{'type': 'dialog', 'text': 'put the blocks into bowls with matching colors'}
>>> get_obj_names()
['orange bowl', 'red block', 'purple bowl', 'blue block', 'blue bowl', 'orange block', 'purple block', 'red bowl']
>>> put_first_on_second('orange block', 'orange bowl')
'success'
>>> put_first_on_second('red block', 'red bowl')
'success'
>>> put_first_on_second('purple block', 'purple bowl')
'success'
>>> put_first_on_second('blue block', 'blue bowl')
'success'
# New task, local variables are reset and world state may have changed.
>>> wait_for_trigger()
{'type': 'dialog', 'text': 'put the blocks in the bowls with mismatched colors'}
>>>
\end{minted}
\end{mdframed} \filbreak

The LLM continues by generating the following:

\begin{mdframed}
\codesize
\begin{minted}{python}
get_obj_names()
\end{minted}
\end{mdframed} \filbreak

The resulting interaction after several generate-execute cycles (each of which involve a full prompt as shown above), including human feedback as the model did not solve the task initially:

\begin{mdframed}
\codesize
\begin{minted}{python}
>>> wait_for_trigger()
{'type': 'dialog', 'text': 'put the blocks in the bowls with mismatched colors'}
>>> get_obj_names()
['red block', 'green block', 'cyan block', 'blue bowl']
>>> put_first_on_second('red block', 'blue bowl')
'success'
>>> wait_for_trigger()
{'type': 'dialog', 'text': 'put all of the blocks in a bowl'}
>>> get_obj_names()
['red block', 'green block', 'cyan block', 'blue bowl']
>>> for block in ['red block', 'green block', 'cyan block']:
...     put_first_on_second(block, 'blue bowl')
>>> wait_for_trigger()
{'type': 'dialog', 'text': 'next time I ask you to put the blocks in bowls with mismatched colors, do not forget any block'}
>>> learn_from_interaction()
>>> wait_for_trigger()
\end{minted}
\end{mdframed} \filbreak

This is the sample generated by \inlinecode{\functionLearnName()}:
\begin{mdframed}
\codesize
\begin{minted}{python}
>>> wait_for_trigger()
{'type': 'dialog', 'text': 'put the blocks in the bowls with mismatched colors'}
>>> get_obj_names()
['red block', 'green block', 'cyan block', 'blue bowl']
>>> # put all blocks in the blue bowl
... for block in ['red block', 'green block', 'cyan block']:
...     put_first_on_second(block, 'blue bowl')
>>> wait_for_trigger()
{'type': 'dialog', 'text': 'put all of the blocks in a bowl'}
>>> get_obj_names()
['red block', 'green block', 'cyan block', 'blue bowl']
>>> for block in ['red block', 'green block', 'cyan block']:
...     put_first_on_second(block, 'blue bowl')
>>> wait_for_trigger()
\end{minted}
\end{mdframed} \filbreak


\vspace{2cm}
\subsection{Learning from Interaction}
\label{app:full_learn_from_interaction_bimanual}

The following is a full trancript of the \inlinecode{learn\_from\_interaction()} call in the \emph{Bimanual Drink \& Cup} scenario.
It starts with two pre-defined examples.
Then, a list of available APIs follows. 
Finally, the current interaction is inserted, and Chain-of-Thought prompting is used to generate an improved interaction.

\begin{mdframed}
\codesize
\textbf{System}: You are a helpful assistant that improves python console transcripts used to control a humanoid household robot, given the requests or corrections provided by a user.\\
\textbf{Human}: I had the following interaction with the robot:
\begin{minted}{python}
>>> wait_for_trigger()
{'type': 'dialog', 'text': 'Can you prepare a pizza for me?'}
>>> list_object_locations()
[('spoon', 'table'), ('ginger tea', 'cupboard'), ('boiler', 'counter'), ('freezer', 'freezer'), ('oven', 'oven')]
>>> say('Ok, I am going to check the freezer for a pizza')
... open_door('freezer')
'success'
>>> bring_object_to('pizza', 'counter')
'success'
>>> open_door('oven')
'success'
>>> bring_object_to('pizza', 'oven')
'success'
>>> close_door('oven')
'success'
>>> turn_device_on('oven')
'finished'
>>> open_door('oven')
'success'
>>> bring_object_to('pizza', 'table')
'success'
>>> say('Here is your pizza. Anything else I can do for you?')
... wait_for_trigger()
{'type': 'dialog', 'text': 'You forgot to close the freezer'}
>>> close_door('freezer')
'success'
>>> say('I am sorry for that mistake. Anything else I can do for you?')
... wait_for_trigger()
{'type': 'dialog', 'text': 'No, now it is fine'}
>>> learn_from_interaction()
\end{minted}
\textbf{Human}: What is the problem in this interaction? Answer with a single sentence.\\
\textbf{AI}: The problem is that the robot did not close the freezer and the oven after taking something out of it.\\
\textbf{Human}: How can the robot do better next time? Answer with a single explanation sentence, no code.\\
\textbf{AI}: Next time, the robot should close the freezer and the oven after taking something out of it.\\
\textbf{Human}: Provide an improved version of the interaction transcript. Your output should be a copy of the above interaction (including the python shell syntax) with only slight modifications to adjust the behavior appropriately. Do not include another learn\_from\_interaction call. Remember to fix the identified problem.\\
\textbf{AI}:
\begin{minted}{python}
>>> wait_for_trigger()
{'type': 'dialog', 'text': 'Can you prepare a pizza for me?'}
>>> list_object_locations()
[('spoon', 'table'), ('ginger tea', 'cupboard'), ('boiler', 'counter'), ('freezer', 'freezer'), ('oven', 'oven')]
>>> say('Ok, I am going to check the freezer for a pizza')
... open_door('freezer')  # Important to close it after getting the pizza out of it
'success'
>>> bring_object_to('pizza', 'counter')
'success'
>>> close_door('freezer')
'success'
>>> open_door('oven')
'success'
>>> bring_object_to('pizza', 'oven')
'success'
>>> close_door('oven')
'success'
>>> turn_device_on('oven')
'finished'
>>> open_door('oven')  # Important to close it after getting the pizza out of it
'success'
>>> bring_object_to('pizza', 'table')
'success'
>>> close_door('oven')
'success'
>>> say('Here is your pizza. Anything else I can do for you?')
... wait_for_trigger()
{'type': 'dialog', 'text': 'No thanks'}
\end{minted}
~\vspace{0.2cm}\\
\textbf{Human}: I had the following interaction with the robot:
\begin{minted}{python}
>>> wait_for_trigger()
{'type': 'dialog', 'text': 'could you make a salad for me?'}
>>> list_objects()
['fridge', 'oven', 'microwave', 'spoon', 'apple', 'plate', 'oil']
>>> say('Ok, I am going to check the fridge for a salad')
... open_door('fridge')
'success'
>>> bring_object_to('salad', 'counter')
'success'
>>> close_door('fridge')
'success'
>>> bring_object_to('salad', 'plate')
'success'
>>> bring_object_to('plate', 'table')
'success'
>>> say('Here is your salad. Anything else I can do for you?')
... wait_for_trigger()
{'type': 'dialog', 'text': 'There is no dressing on it. I dont want to have plain salad'}
>>> say('I am sorry for that mistake')
... wait_for_trigger()
{'type': 'dialog', 'text': 'Now please get me some dressing'}
>>> bring_object_to('oil', 'plate')
'failure'
>>> say('I am sorry, I cannot pour the oil into the plate. Can you assist?')
... wait_for_trigger()
{'type': 'dialog', 'text': 'please just bring the oil to me and I will do it myself'}
>>> bring_object_to('oil', 'table')
'success'
>>> wait_for_trigger()
{'type': 'dialog', 'text': 'Thanks. Next time please serve the oil with the salad rightaway'}
>>> learn_from_interaction()
\end{minted}
\textbf{Human}: What is the problem in this interaction? Answer with a single sentence.\\
\textbf{AI}: The problem is that the robot offered only plain salad without a dressing.\\
\textbf{Human}: How can the robot do better next time? Answer with a single explanation sentence, no code.\\
\textbf{AI}: Next time, the robot should serve the oil on the table with the salad right away.\\
\textbf{Human}: Provide an improved version of the interaction transcript. Your output should be a copy of the above interaction (including the python shell syntax) with only slight modifications to adjust the behavior appropriately. Do not include another learn\_from\_interaction call. Remember to fix the identified problem.\\
\textbf{AI}: 
\begin{minted}{python}
>>> wait_for_trigger()
{'type': 'dialog', 'text': 'could you make a salad for me?'}
>>> list_objects()
['fridge', 'oven', 'microwave', 'spoon', 'apple', 'plate', 'oil']
>>> say('Ok, I am going to check the fridge for a salad')
... open_door('fridge')
'success'
>>> bring_object_to('salad', 'counter')
'success'
>>> close_door('fridge')
'success'
>>> bring_object_to('salad', 'plate')
'success'
>>> bring_object_to('plate', 'table')
'success'
>>> # The user earlier asked me to always serve oil with the salad, so let's do that
... bring_object_to('oil', 'table')
'success'
>>> say('Here is your salad and oil. Anything else I can do for you?')
... wait_for_trigger()
{'type': 'dialog', 'text': 'Fine for now'}
\end{minted}
\textbf{Human}: These are the available APIs:
\begin{minted}{python}

def ask(question: str)
def bring_object_to(object_name: str, destination_object_name: str) # full pick and place method. assumes that the object is 
                                                    # already detected. Assure that using list_objects() before calling this
def get_location_coordinates(location_name: str) -> np.ndarray
def get_obj_pos(obj_name: str) -> np.ndarray
def get_object_in_hand(hand: Literal['left', 'right', None] = None) -> Optional[str]
def grasp(object_name: str, hand: Literal['left', 'right', None] = None) # make sure to move to the location of 
                                                                         # the object before grasping it
def handover_object_to_human(object_name: str)
def is_object_at_location(object_name: str, location_name: str) -> bool
def list_locations() -> List[str]
def list_object_locations(affordance: Literal[None, 'grasp', 'place_something_on_top'] = 'grasp') -> List[Tuple[str, str]]
def list_objects(affordance: Literal[None, 'grasp', 'place_something_on_top'] = 'grasp') -> List[str]
def move_to(position: np.ndarray, speed_factor=1.0) # Change speed only on user request
def place_object(object_in_hand_name: str, destination_object_name: str) # place the already grasped object on top of the 
                         # destination object. make sure to move to the location of the destination object before placing
def receive_object_from_human(object_name: str)
def say(text: str)
def wait_for_trigger() -> Dict[str, str] # always call this to wait for next command or end the interaction

\end{minted}
\textbf{Human}: I had the following interaction with the robot:
\begin{minted}{python}
>>> wait_for_trigger()
{'type': 'dialog', 'text': 'I want to have some juice'}
>>> list_object_locations(affordance='grasp')
[('multivitamin-juice_0', 'on_mobile-kitchen-counter_0'), ('cup_large_0', 'on_mobile-kitchen-counter_0'), 
 ('bauhaus-sponge_0', 'on_square-table-80x80_0')]
>>> say('Ok, I will bring you the multivitamin juice')
... # The juice is at the mobile-kitchen-counter. check the correct name for navigation:
... list_locations()
['inFrontOf_mobile-kitchen-counter_0', 'inFrontOf_square-table-80x80_0', 'room_center', 'handover_to_human']
>>> move_to(get_location_coordinates('inFrontOf_mobile-kitchen-counter_0'))
'success'
>>> grasp('multivitamin-juice_0')
'success'
>>> move_to(get_location_coordinates('handover_to_human'))
'success'
>>> say("Here is the multivitamin juice.")
>>> handover_object_to_human('multivitamin-juice_0')
'success'
>>> wait_for_trigger()
{'type': 'dialog', 'text': 'can you also bring me a cup?'}
>>> say('Ok, I will bring you a cup')
... # The cup is at the mobile-kitchen-counter. check the correct name for navigation:
... list_locations()
['inFrontOf_mobile-kitchen-counter_0', 'inFrontOf_square-table-80x80_0', 'room_center', 'handover_to_human']
>>> move_to(get_location_coordinates('inFrontOf_mobile-kitchen-counter_0'))
'success'
>>> grasp('cup_large_0')
'success'
>>> move_to(get_location_coordinates('handover_to_human'))
'success'
>>> say("Here is the cup.")
>>> handover_object_to_human('cup_large_0')
'success'
>>> wait_for_trigger()
{'type': 'dialog', 'text': 'Thanks, but next time, can you do this both at once? You have two hands'}
>>> learn_from_interaction()
\end{minted}
\textbf{Human}: What is the problem in this interaction? Answer with a single sentence.\\
\textbf{AI}: The problem is that the robot is not utilizing both of its hands to bring the juice and the cup at the same time, resulting in two separate trips.\\
\textbf{Human}: How can the robot do better next time? Answer with a single explanation sentence, no code.\\
\textbf{AI}: Next time, the robot should use both of its hands to grasp the juice and the cup at the same time, reducing the number of trips and increasing efficiency.\\
\textbf{Human}: Provide an improved version of the interaction transcript. Your output should be a copy of the above interaction (including the python shell syntax) with only slight modifications to adjust the behavior appropriately. Do not include another learn\_from\_interaction call. Remember to fix the identified problem.\\
\textbf{AI}:
\begin{minted}{python}
>>> wait_for_trigger()
{'type': 'dialog', 'text': 'I want to have some juice'}
>>> list_object_locations(affordance='grasp')
[('multivitamin-juice_0', 'on_mobile-kitchen-counter_0'), ('cup_large_0', 'on_mobile-kitchen-counter_0'),
 ('bauhaus-sponge_0', 'on_square-table-80x80_0')]
>>> say('Ok, I will bring you the multivitamin juice and a cup')
... # The juice and the cup are at the mobile-kitchen-counter. check the correct name for navigation:
... list_locations()
['inFrontOf_mobile-kitchen-counter_0', 'inFrontOf_square-table-80x80_0', 'room_center', 'handover_to_human']
>>> move_to(get_location_coordinates('inFrontOf_mobile-kitchen-counter_0'))
'success'
>>> grasp('multivitamin-juice_0', 'left')  # Grasp juice with left hand
'success'
>>> grasp('cup_large_0', 'right')  # Grasp cup with right hand
'success'
>>> move_to(get_location_coordinates('handover_to_human'))
'success'
>>> say("Here is the multivitamin juice and the cup.")
>>> handover_object_to_human('multivitamin-juice_0')
'success'
>>> handover_object_to_human('cup_large_0')
'success'
>>> wait_for_trigger()
{'type': 'dialog', 'text': 'Thanks, that was much quicker'}
\end{minted}
\end{mdframed}
}

\end{document}